\documentclass[
paper=A4,               
pagesize=auto,          
fontsize=12pt,          
DIV=16,                 
twoside=true,           
BCOR=20mm,              
parskip=false,          
chapterprefix=true,     
appendixprefix=true,    
listof=totoc,           
bibliography=totoc,     
headinclude=true,       
footinclude=false,      
headsepline=true,       
footsepline=false,      
headings=small,         
numbers=noenddot        
] {scrbook}

\usepackage{lmodern}
\usepackage[T1]{fontenc}
\usepackage[utf8]{inputenc}
\usepackage[onehalfspacing]{setspace}
\usepackage{amsmath,amssymb}
\usepackage{graphicx}
\usepackage{wrapfig}
\usepackage{booktabs}
\usepackage[printonlyused]{acronym}
\usepackage[acronym]{glossaries}
\usepackage{pdfpages}
\usepackage{svg}
\usepackage{blindtext} 
\graphicspath{{figures/}}
\usepackage[export]{adjustbox}
\usepackage{graphicx}
\usepackage{marvosym}
\usepackage{float}
\usepackage{svg}
\usepackage{caption}
\usepackage{subcaption}
\usepackage{amsmath}
\usepackage{pgfplots}
\usepackage{datetime}
\usepackage{tikz}
\usetikzlibrary{3d}
\usepackage{pgfplots}
\usepackage{tikz-3dplot}
\usepackage{tikz-3dplot}
\usepackage{tabularx}
\usepackage{tikz}
\usetikzlibrary{calc}
\usetikzlibrary{matrix}
\usetikzlibrary{positioning}
\usetikzlibrary{shapes.geometric}
\recalctypearea
\title{Projection-Aware End-to-End Learned Video Compression for 360-Degree Video}
\author{Niloofar Maani}
\date{April 2023}
\begin{document}
    \pagenumbering{Alph} 
    \maketitle
    \cleardoublepage
    \includepdf[pages={1},scale={0.95}]{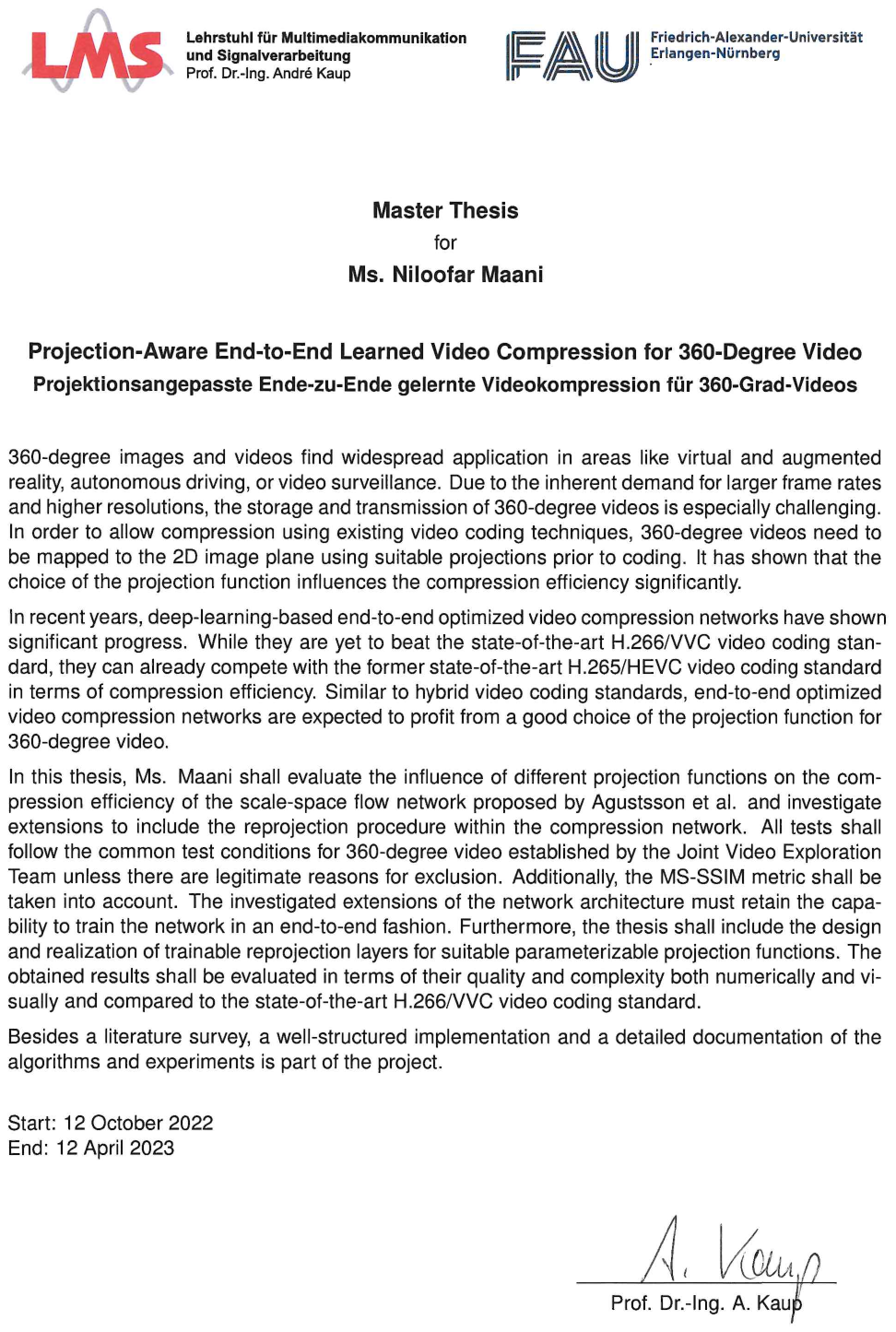}
    \cleardoublepage
    
    \chapter*{Declaration}
\thispagestyle{empty}

\noindent
I confirm that I have written this thesis unaided and without using sources other than those listed and that this thesis has never been submitted to another examination authority and accepted as part of an examination achievement, neither in this form nor in a similar form.
All content that was taken from a third party either verbatim or in substance has been acknowledged as such.

\vspace{3cm}

\begin{minipage}[t]{0.45\textwidth}
    \rule{\textwidth}{0.5pt}\\
	Place, Date
\end{minipage}
\hfill
\begin{minipage}[t]{0.45\textwidth}
	\rule{\textwidth}{0.5pt}\\
	Niloofar Maani\\

\end{minipage}

    \frontmatter
    \pagenumbering{Roman}
    \tableofcontents
    \chapter{Kurzfassung}

Bilder und Videos mit einer 360-Grad-Perspektive werden in verschiedenen Anwendungsbereichen eingesetzt, wie zum Beispiel in der virtuellen Realität (VR) \cite{sreedhar2016viewport}, selbstfahrenden Fahrzeugen \cite{chang2019argoverse} und Bildungszwecken \cite{feurstein2018towards}. 360-Grad-Videos umfassen eine umfassende horizontale Abdeckung von 360 Grad und einen vertikalen Bereich von 180 Grad. Bei dieser Art von Video ist die Perspektive des Betrachters nicht festgelegt. Diese Technologie ermöglicht es den Zuschauern, die Szene zu erkunden und zu steuern, indem sie ihren bevorzugten Betrachtungswinkel auswählen. Daher haben 360-Grad-Bilder und -Videos eine höhere Auflösung \cite{sreedhar2016viewport} und ihre Speicherung und Übertragung \cite{huang2019utility} kann herausfordernd sein. Da 360-Grad-Videos in Kugelform vorliegen, können sie nicht mit Standard-Videocodierungstechniken verarbeitet werden. Daher müssen sie in 2D-Bilder projiziert werden. Verschiedene Projektionstechniken können auf 360-Grad-Inhalte angewendet werden. In dieser Arbeit werden die Projektionen untersucht, die von der JVET 360-Degree Projection Software \cite{he2017360lib} angeboten werden, und miteinander verglichen, um herauszufinden, welche für die Kompression am effizientesten ist.

In dieser Arbeit wird ein auf Deep Learning basierendes, end-to-end optimiertes Videokompressionsnetzwerk verwendet. Diese Modelle übertreffen die Videocodierungsstandards H.265/HEVC \cite{wiegand2003overview} und H.264/HEVC \cite{sullivan2012overview} in Bezug auf die Kompressionseffizienz \cite{lu2019dvc},\cite{pessoa2020end}. Das Ziel ist es, die Leistung von end-to-end optimierten Videokompressionsnetzwerken in Gegenwart von Projektionsfunktionen für 360-Grad-Videos zu bewerten. Um dieses Ziel zu erreichen, wird das Scale-Space-Flow-Netzwerk \cite{agustsson2020scale} verwendet. Die Bewertung basiert auf den gemeinsamen Testbedingungen und Bewertungsverfahren von JVET für 360-Grad-Videos \cite{boyce2017jvet}. Für diese Bewertung werden 360-Grad-Testsequenzen verwendet, die von JVET bereitgestellt und mit einer anfänglichen Quellprojektion versehen sind. Diese Sequenzen werden projiziert, um das Codierprojektionsformat zu erzeugen, dann wird die Sequenz mit einem auf Deep Learning basierenden, end-to-end optimierten Videokompressionsnetzwerk komprimiert. Anschließend wird das rekonstruierte Signal in das Quellprojektionsformat konvertiert. Dieses Verfahren dient als Hauptmethode zur Bewertung verschiedener Projektionen unter Verwendung der objektiven Metriken, die von der 360Lib-Software berechnet werden.

    \chapter{Abstract}

Images and videos with a 360-degree perspective are utilized across a range of applications, encompassing virtual reality (VR) \cite{sreedhar2016viewport}, self-driving vehicles \cite{chang2019argoverse}, and educational purposes \cite{feurstein2018towards}. 360-degree videos encompass a comprehensive horizontal coverage of 360 degrees and a vertical range of 180 degrees. In this type of video, the viewer's perspective is not fixed. This technology allows viewers to explore and control the scene by selecting their preferred viewing angle. Therefore, 360-degree images and videos have higher resolution \cite{sreedhar2016viewport}, and storing and transmitting \cite{huang2019utility} them can be challenging. Since 360-degree videos are in a sphere form, standard video coding techniques cannot process them. As a result, they need to be projected into 2D images. Different types of projection techniques can be applied to 360-degree content. This thesis explores the projections offered by the JVET 360-Degree projection Software \cite{he2017360lib} and compares them to find which one is more efficient for compression.

In this thesis, a deep-learning-based end-to-end optimized video compression network is used. These models outperform the H.265/HEVC \cite{wiegand2003overview} and H.264/HEVC \cite{sullivan2012overview} video coding standards in terms of compression efficiency \cite{lu2019dvc},\cite{pessoa2020end}. The aim is to evaluate the performance of end-to-end optimized video compression networks in the presence of projection functions for 360-degree videos. To achieve this goal, the scale-space flow network \cite{agustsson2020scale} is used. The evaluation is based on JVET common test conditions and evaluation procedures for 360-degree video \cite{boyce2017jvet}. For this evaluation, 360-degree testing sequences provided by JVET, which have an initial source projection, are projected to generate the coding projection format, then the sequence is compressed using a deep-learning-based end-to-end optimized video compression network. Afterward, the reconstructed signal is converted to the source projection format. This pipeline serves as the main method for evaluating different projections using the objective metrics calculated by the 360Lib software.

    \chapter{Symbols and Notations}
\begin{tabular}{ll}
$v_t$	&	optical flow\\

$m_t$	&	quantized signal\\

$\hat{v_t}$	& reconstructed motion vector\\

$x_t$	& current frame\\

$\hat{x_t}$	& reconstructed frame\\

$\Bar{x_t}$	& predicted frame\\

$\hat{y_t}$	& residual data\\

$h$	& convolution parameter\\

$c_k$	& convolution bias\\

$s_k$	& down sampling factor\\

$\beta_k$	& normalization bias\\

$\gamma$	& normalization scale\\

$L(.)$	& loss function\\

$||.||$	& $l_2$ norm\\

$q$	& posterior distribution\\

$\phi, \theta$	& VAE parameter\\

$D_{kl}$	&  Kullback liebler divergence\\

$I(x,y,t)$	&  pixel intensity function\\

$(u,v)$	&  optical flow vector\\

$p(x)$	&  charbonnier loss\\

$(f_x,f_y)$	&  2D displacement field\\

$(g_x,g_y,g_z)$	&  3D displacement field\\

$[z_0]$	&  latent space\\

$[w_i]$	&  wrapped latent\\

$(X,Y,Z)$	&  3D cordinate system\\

$(u,v)$	&  2D plane\\

$\phi$	&  longitude\\

\end{tabular}

\begin{tabular}{ll}

$\theta$	&  latitude\\

$(W,H)$	&  frame Width and Height\\

$\tan$	&  tangent function\\

$\arctan$	& inverse tangent function\\

$\mathrm{sgn}$	& sign function\\

$f_r$	& reference frame\\

$f_c$	& coding face\\

$\beta$	& yaw rotation angle\\

$\alpha$	& pitch rotation angle\\

$\gamma$	& roll rotation angle\\

$d(.,.)$	& distortion metric\\

$\lambda$	& Lagrange multiplier\\
$(M,N)$	& frame size\\
$s(i,j)$	& spherical factor\\
$r$	& radius\\
\end{tabular}

    \chapter{Abbreviations and Acronyms}

\begin{tabular}{ll}
\textbf{ACP} & Adjusted Cubemap Projection \\
\textbf{BD Rate} & Bjontegaard Delta rate \\
\textbf{BDCT} & Block-based Discrete Cosine Transform \\
\textbf{CNN} & Convolutional Neural Network \\
\textbf{CMP} & CubeMap Projection \\
\textbf{DCT} & Discrete Cosine Transform \\
\textbf{DNN} & Deep Neural Network \\
\textbf{DVC} & Deep Video Compression \\
\textbf{EAC} & Equi Angular Cubemap Projection \\
\textbf{EBCOT} & Embedded Block Coding with Optimized Truncation \\
\textbf{ERP} & Equi-Rectangular Projection \\
\textbf{FPS} & Frames Per Second \\
\textbf{FlowNet} & Flow Network \\
\textbf{GDN} & Generalized Division Normalization \\
\textbf{HEC} & Hybrid Equi Angular Cubemap Projection \\
\textbf{HEVC} & High Efficiency Video Coding \\
\textbf{JPEG} & Joint Photographic Experts Group \\
\textbf{JVET} & Joint Video Exploration Team \\
\textbf{KL} & Kullback Leibler \\
\textbf{MV} & Motion Vector \\
\textbf{PERP} & Padded Equi-Rectangular Projection \\
\textbf{PRELU} & Parametric Rectified Linear Unit \\
\textbf{PSNR} & Peak Signal-to-Noise Ratio \\
\textbf{RSP} & Rotated Sphere Projection \\
\textbf{S-PSNR} & Spherical PSNR \\
\textbf{VAE} & Variational Auto Encoder \\
\textbf{VR} & Virtual Reality \\
\textbf{WS-PSNR} & Weighted to Spherically uniform PSNR \\
\end{tabular}

    \mainmatter
    \chapter{Introduction}
\label{chap:Introduction}

\begin{figure}[h]
\centering
\includegraphics[width=90mm]{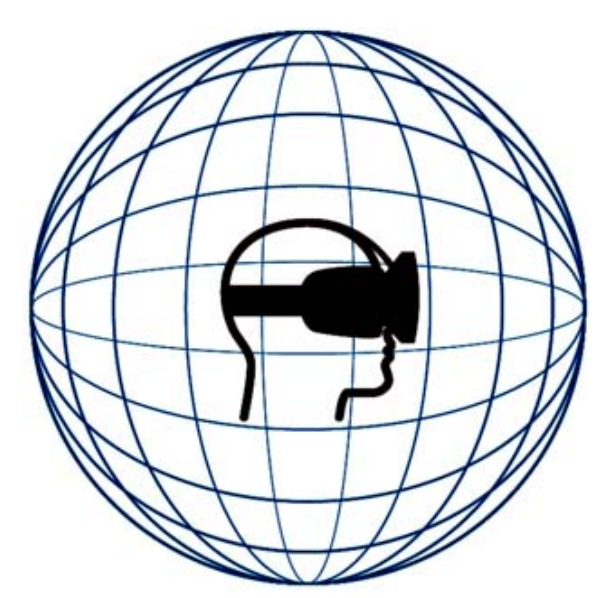}
\caption{360-degree video field of view from \cite{hosseini2016adaptive}.}
\label{fig:360-degree}
\end{figure}

A 360-degree video is a special type of video that allows users to look around in all directions instead of just seeing what the camera points at \cite{hosseini2016adaptive}. It is made by combining videos from cameras placed in a circle, providing a wide view and making the experience more engaging, as depicted in Figure \ref{fig:360-degree}. 360-degree video applications cover a wide variety of areas and are becoming increasingly popular in sectors like entertainment, shopping, and healthcare \cite{jamali2019comparison}, \cite{argyriou2016engaging}. As a key format in virtual reality (VR) \cite{herglotz2019efficient}, these videos open up countless opportunities for activities like gaming.

Different from traditional videos, VR surrounds users with a full view, containing 360-degree horizontal and 180-degree vertical perspectives for an enhanced experience. Despite the rising demand and interest in 360-degree content, there are still many challenges in processing these images. 360-degree videos require high resolution and frame rates for an immersive experience, resulting in large file sizes, which makes it challenging to store and transmit these videos quickly \cite{huang2019utility}. Improving compression methods is important to help solve this issue.

Processing 360-degree videos is difficult because they are spherical, while current video standards treat videos as flat rectangles \cite{jamali2019comparison}. This difference creates unique challenges that need to be addressed. To address this issue, the 3D image is projected onto a 2D plane for encoding, allowing the use of existing video coding frameworks \cite{jamali2019comparison}.

Two widely used projection formats are equirectangular projection (ERP) \cite{ye2017jvet} and cubemap projection (CMP) \cite{ye2017jvet}. The ERP method involves projecting spherical content onto a cylinder, which is then unraveled into a rectangular shape. However, a significant disadvantage of ERP is that it leads to oversampling in the polar regions, causing distortions in the content on the projected plane. The mapping process introduces distortion, making the selection of an appropriate projection essential for effective compression.

Addressing the challenges posed by 360-degree videos is crucial for their successful transmission. A pipeline is being established to evaluate the efficiency of different projections in terms of compressing these videos. Studies have explored coding performance under different projection methods, with a focus on codecs such as HEVC. One study evaluated the impact of projections on the coding efficiency of monoscopic videos and considered the quality of various viewports. Another provided an overview of recent advancements in omnidirectional video. Yet another study compared the effectiveness of various projections in multi-view 360-degree videos.

In this study, a deep-learning-based end-to-end optimized video compression network is employed for encoding and decoding projected 360-degree video frames. Deep neural networks have demonstrated significant advancements in end-to-end video compression, surpassing traditional methods like H.265/HEVC and H.264/AVC due to their data-driven approach and adaptability \cite{lu2019dvc}. Unlike traditional codecs, which rely on separately optimized handcrafted techniques, machine learning models such as deep neural network-based autoencoders excel in performance. These learning-based methods utilize extensive end-to-end training and highly non-linear transformations, offering efficient solutions through a latent representation of input videos acquired by training on similar content \cite{lu2019dvc}.

The goal of this research is to assess the performance of end-to-end optimized video compression networks in the context of projection functions for 360-degree videos. To achieve this, the scale-space flow network \cite{agustsson2020scale} is used, and the JVET common test conditions and evaluation procedures for 360-degree video are followed \cite{boyce2017jvet}. The study leverages 360-degree testing sequences provided by JVET, which include an initial source projection. These sequences are projected to generate the coding projection format, compressed with the deep-learning-based end-to-end optimized video compression network, and the reconstructed signal is then converted back to the source projection format. This pipeline forms the primary method for evaluating different projections, with objective metrics calculated by the 360Lib software.

To further enhance the approach in handling 360-degree video compression, trainable reprojection layers in Python have been developed alongside the existing 360Lib software. This novel method employs a three-layer model, where the first layer transforms the reference projection into the coding projection, the middle layer compresses the coding projection, and the final layer converts the coding projection back to the reference projection. It is important to highlight that these projection conversion layers do not contain learning parameters; they are solely responsible for pre- and post-processing of the input and output of the compression layer. This innovative approach is introduced to address the unique challenges of 360-degree video compression effectively.

Chapter 2 delves into the background of image and video compression, discussing the current state of learning-based video compression methods and their relevance to the project. In Chapter 3, various projections utilized in this project are introduced, explaining their formulation and implementation within the 360Lib software. Chapter 4 provides a comprehensive description of the pipeline implemented to evaluate the compression model in the presence of projections. It details the methodology, design choices, and rationale behind the approach. In Chapter 5, the results obtained from the pipeline are presented, and a comparative analysis of the two implemented methods is conducted. The implications of these findings and their relevance to the broader context of 360-degree video compression are discussed. Finally, Chapter 6 summarizes the methods and results, drawing conclusions based on the findings and highlighting possible future directions for research in this area.

    \chapter{Learning-based Image and Video compression}
\label{chap: Learning-based Image and Video compression}

 This chapter provides a comprehensive overview of the critical principles of video compression using deep learning. It highlights the various projection functions evaluated as part of this project. Additionally, the chapter delves into the significance of using deep learning for video compression and explains how it has the potential to revolutionize the field by delivering high-quality results with increased efficiency.

\section{Image compression}

\begin{figure}[h]
\centering
\includegraphics[width=100mm]{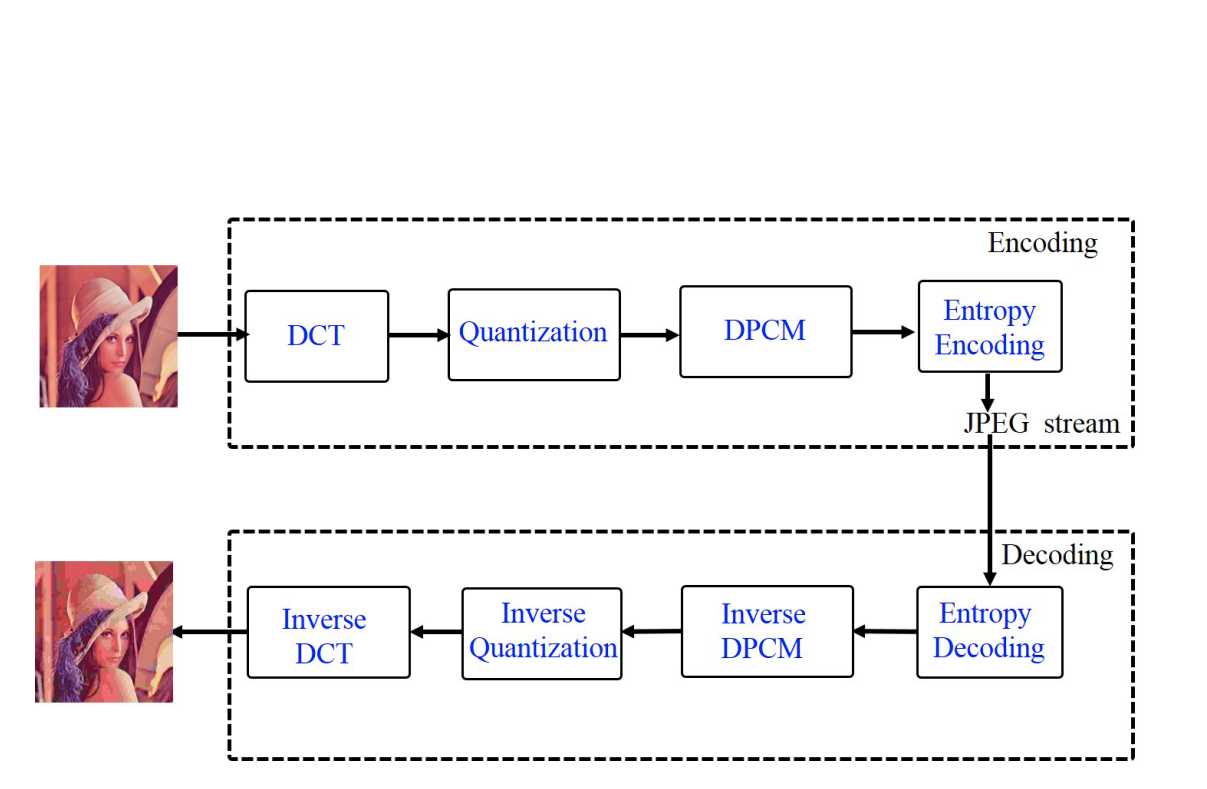}
\caption{JPEG based image codec from \cite{ma2019image}.}
\label{fig:jpeg}
\end{figure}

The objective of image and video compression methods is to decrease the bit count needed for representing images or videos while ensuring minimal compromise in quality. Compression removes redundancies and irrelevant data, which leads to  reduced storage space and bandwidth required for transmission \cite{sadeeq2021image}. Within the domain of image and video compression, key methodologies include transformation and prediction techniques. The well-established image compression standard, JPEG \cite{wallace1991jpeg}, integrates essential transform/prediction components as illustrated in Figure \ref{fig:jpeg}. Within the JPEG procedure, input images are partitioned into distinct 8x8 blocks with no overlap. After processing the blocks, they are converted into a different representation through the use of block-DCT (BDCT). As an alternative to directly compressing the DC value, the differential pulse code modulation (DPCM) is utilized for each block, targeting the DC components. This method compresses the prediction residuals between neighboring DCT blocks' DC components \cite{mun2012dpcm}. Afterward, quantization and entropy coding procedures are employed to compress the coefficients of the representation into a binary sequence \cite{ma2019image}. JPEG 2000 \cite{christopoulos2000jpeg2000} is also a popular image compression standard that employs 2D wavelet transform to represent still images efficiently. The utilization of EBCOT \cite{taubman2000high}, an efficient arithmetic coding technique, is employed to minimize the statistical redundancy within wavelet coefficients. Additionally, JPEG 2000 employs multi-scale orthogonal wavelet decomposition to further enhance compression. 

\begin{figure}[h]
\centering
\includegraphics[width=100mm]{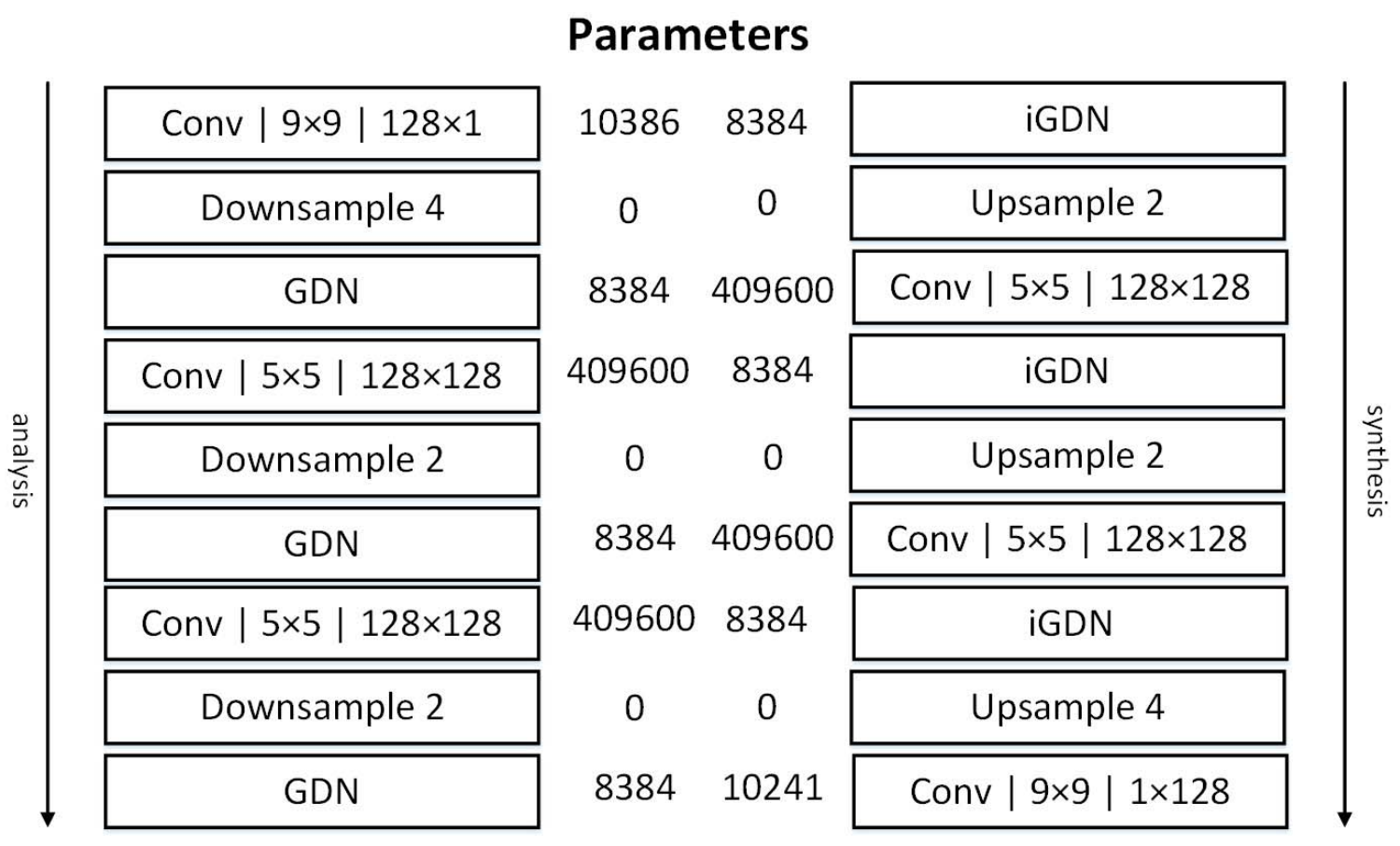}
\caption{Architecture of CNN based compression model \cite{balle2016end} From \cite{ma2019image}.}
\label{fig:bale}
\end{figure}
While typically, transform coding methods use separate optimization of the three components - transform, quantizer, and entropy code - through manual parameter adjustments, JPEG 2000 integrates these components for better compression performance.

Image compression techniques using neural networks have emerged and shown remarkable performance \cite{toderici2015variable,theis2017lossy,rippel2017real}. Convolutional Neural Networks (CNNs) \cite{o2015introduction} have particularly stood out, showing exceptional results in processes including image classification and object identification, outperforming traditional algorithms \cite{lecun2015deep}. This is due to the convolution operation in CNNs, which effectively captures the relationship between neighboring pixels and mirrors the statistical properties of natural images  \cite{ma2019image}.
\clearpage
In 2016, Ballé et al. presented a CNN-based model for end-to-end image compression \cite{balle2016end}. The framework, as depicted in Figure \ref{fig:bale}, is made up of two principal parts, namely the analysis transform functioning as the encoder and the synthesis transform serving as the decoder. As mentioned in \cite{ma2019image} the analysis process involves three components: the convolution, subsampling, and the divisive normalization stages. The initial step of the convolution stage involves an affine convolution, as 
\begin{equation}
v^{(k)}_{i}(m, n) = (h_{k,ij} * u^{(k)}_j)(m, n) + c_{k,i}.
\label{equ:convolution}
\end{equation}

 $u^{(k)}_j$ is defined as the $j^{th}$ channel at the $(m, n)^{th}$ position of the $k^{th}$ stage. The 2D convolution operation is symbolized by $*$, while the convolution parameter is indicated as $h_{k,ij}$. The bias for the convolution neural network is denoted by $c_{k,i}$.

The results of the convolution step are then reduced in resolution

\begin{equation}
w^{(k)}_i(m, n) = v^{(k)}_i(s_km,s_kn).
\label{equ:downsampling}
\end{equation}

The downsampling factor is represented by $s_k$.
Generalized Divisive Normalization (GDN) is applied to the signals following the downsampling step for further processing, as  

\begin{equation}
u^{(k+1)}i(m, n) = \frac{w^{(k)}i(m, n)}{(\beta_{k,i} + \sum_j \gamma_{k,ij}(w^{(k)}_j(m, n))^2)^{\frac{1}{2}}}, 
\label{equ:gdn}
\end{equation}

in this equation, the normalization operation's bias and scale parameters are denoted by $\beta_{k,i}$ and $\gamma_{k,ij}$, respectively. GDN is a collaborative nonlinearity which has proven to be extremely efficient in transforming the local joint statistics of natural images to resemble Gaussian distributions. GDN demonstrates superior efficiency in transforming image densities to Gaussian-like forms compared to the sequential application of linear transformations and pointwise nonlinearities \cite{balle2016end}. 

In the synthesis transform, all parameters, including the convolution parameters, biases, normalization biases, and normalization scales, are optimized end-to-end based on a rate-distortion objective function. Ballé et al. resolved the issue of zero derivatives in quantization by integrating an i.i.d uniform noise into the CNN model, effectively transforming the quantizer. This allowed them to address the optimization problem using a stochastic gradient descent approach \cite{ma2019image}. This approach outperformed JPEG2000 on the PSNR and MS-SSIM metrics.

\section{Video compression}
Traditional video compression strategies have relied on meticulously crafted components \cite{sullivan2012overview,wiegand2003overview}, including block-based motion estimation \cite{zhu2000new} and Discrete Cosine Transform (DCT) \cite{watson1994image}. Block-based motion estimation seeks to minimize temporal redundancies by detecting motion between adjacent frames, dividing them into blocks, and then estimating the motion vectors for each block. Discrete Cosine Transform (DCT) is another essential component in video compression, responsible for reducing spatial redundancies. DCT transforms video frames into a frequency domain representation, focusing on retaining more visually significant low-frequency components while discarding less important high-frequency components. By doing so, DCT achieves a higher compression rate while maintaining acceptable visual quality. This method has been a cornerstone of video compression codecs, including the widely used JPEG image compression and various video compression standards. Despite the efficiency of these handcrafted components, these components lack end-to-end optimization, which could potentially unlock further improvements in video compression performance \cite{lu2019dvc}.  Lately, methods based on deep neural networks (DNNs) have been developed to enhance particular components within conventional video compression algorithms such as \cite{chen2017deepcoder,liu2016cu,song2017neural,lu2018deep}. However, these techniques do not fully address end-to-end optimization or include aspects like motion estimation and compression. DVC, introduced by Guo Lu et al. \cite{lu2019dvc}, distinguishes itself among various compression models due to its end-to-end implementation, which will be discussed in more detail here. But first, the main parts of a traditional video codec as mentioned in \cite{lu2019dvc} are introduced.

\begin{figure}[h]
\centering
\includegraphics[width=130mm]{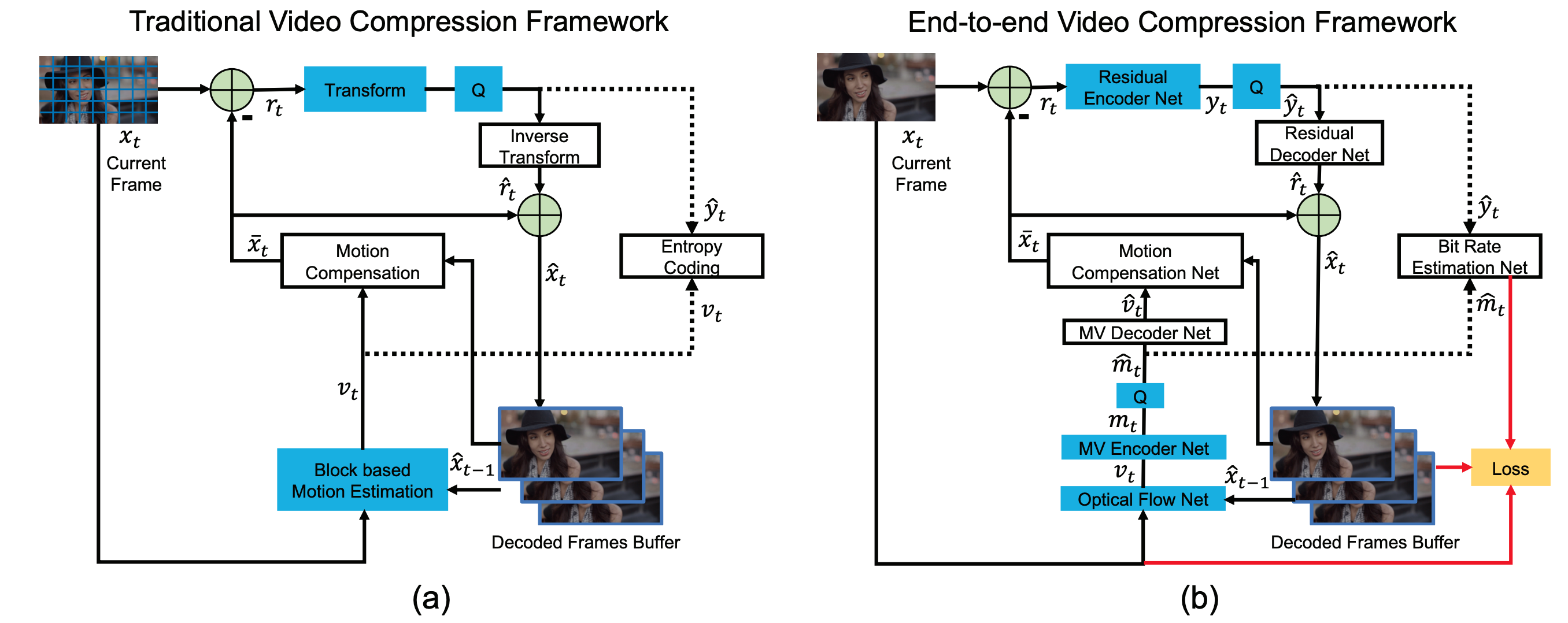}
\caption[Hybrid video codec and compression model architecture]{(a): An illustration of the conventional video codecs. (b): The
Deep compression model from \cite{lu2019dvc}.}
\label{fig:videocompression}
\end{figure}

Figure \ref{fig:videocompression}(a) presents an overview of the video compression process \cite{sullivan2012overview}, \cite{wiegand2003overview}. First, input frames are classified as I-frames, P-frames, or B-frames based on their content and relationship with other frames. Next, motion estimation calculates movement between frames using block-matching algorithms. With these calculations, motion compensation creates a predicted frame. The prediction is either intra prediction, which uses data from the current frame, or inter prediction, which uses data from previous or future frames. This distinction leads to the classification of frames: I-frames for intra prediction and P-frames or B-frames for inter prediction. P-frames depend on previous frames, while B-frames utilize both previous and future frames, made possible by encoding and decoding frames in an order different from their display sequence. The difference between the original and predicted frames, known as the residual frame, is then transformed by employing methods like DCT. 
\clearpage

This transformation allows for efficient processing in the frequency domain and enables smoother quantization with less noticeable artifacts. After transformation, quantization is applied to the coefficients to reduce the bitrate and improve compression. Both motion vectors and quantized coefficients are encoded by entropy coding techniques, like Huffman or Arithmetic coding, before being transmitted or stored. During decoding, the compressed data is first entropy-decoded and dequantized. It then undergoes an inverse transform to reconstruct the residual frame. Simultaneously, the predicted frame is generated using the reference frame and decoded motion vectors through motion-compensated prediction. Afterward, the reconstructed residual frame is combined with the predicted frame to produce the final decoded frame. This frame is displayed on the screen, and the process repeats for subsequent frames, creating a seamless connection between each step in the workflow.

\begin{figure}[h]
\centering
\includegraphics[width=130mm]{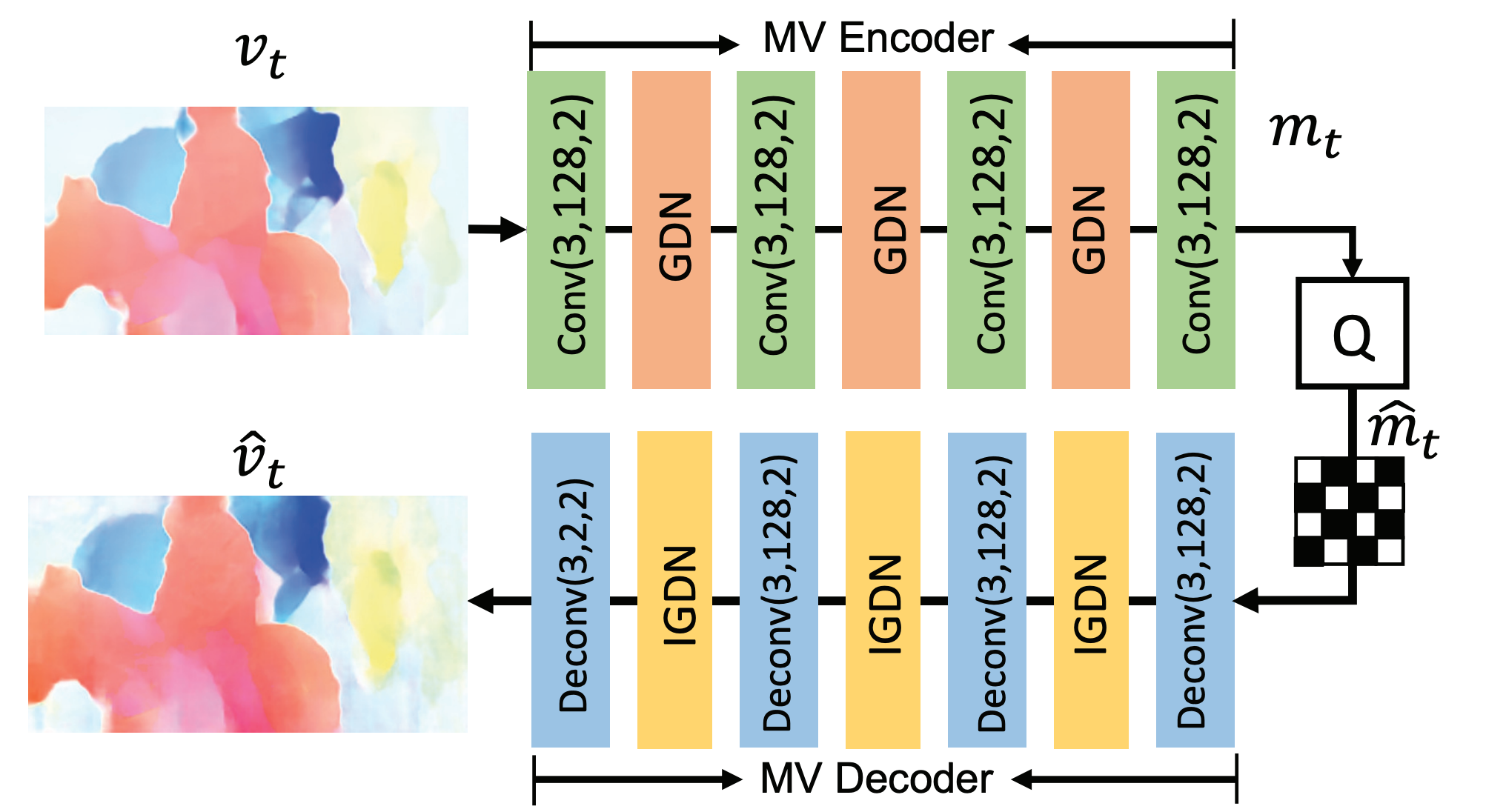}
\caption{DVC MV Encoder-decoder from \cite{lu2019dvc}.}
\label{fig:videocompression3}
\end{figure}

\begin{figure}[h]
\centering
\includegraphics[width=130mm]{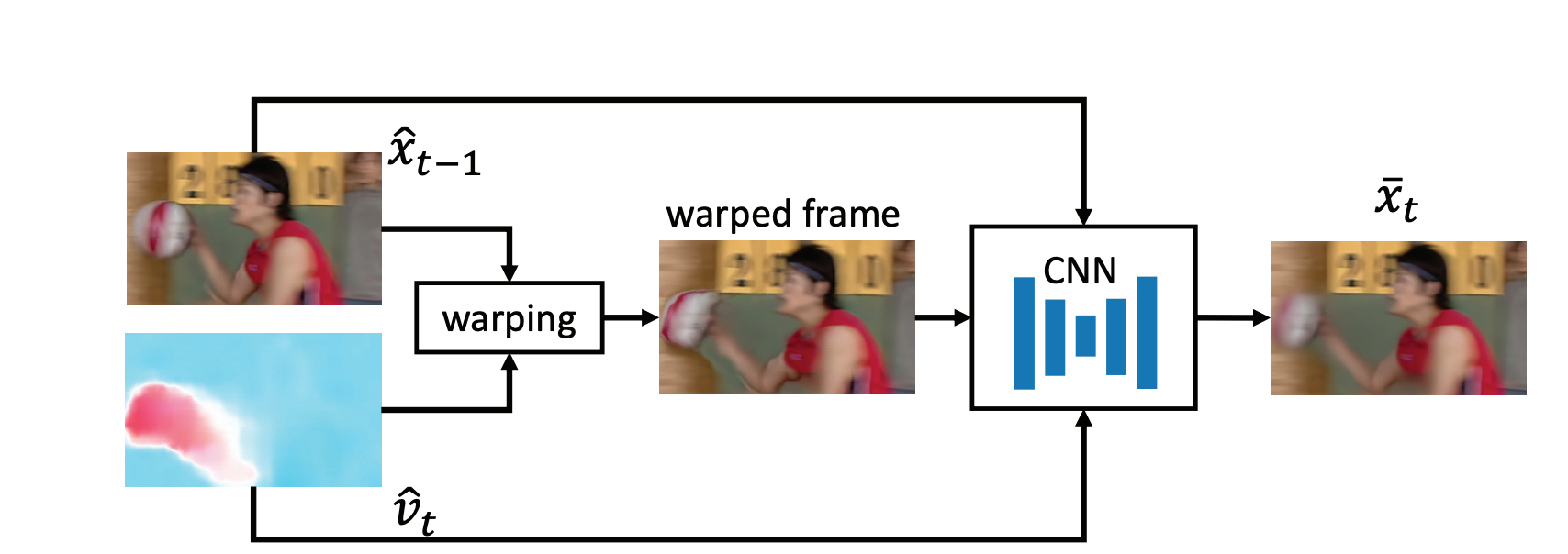}
\caption{DVC model compensation model from \cite{lu2019dvc}.}
\label{fig:videocompression2}
\end{figure}

Figure \ref{fig:videocompression}(b) represents the DVC framework suggested by \cite{lu2019dvc}. The main distinctions and connections between these two frameworks are discussed here.

Initially, a CNN model is utilized for motion estimation and compression by estimating optical flow $v_t$. Subsequently, an MV encoder-decoder model is employed to compress and decode the optical flow $v_t$. This leads to a discretized representation of motion, denoted as $\hat{m}_t$, and the reassembled motion data, symbolized by $\hat{v}_t$, can be decoded through the Motion Vector (MV) decoder network. MV is a CNN, which converts the optical flow into corresponding representations that facilitate more effective encoding. As illustrated in Figure \ref{fig:videocompression3},  $v_t$ undergoes a sequence of convolution operations and nonlinear functions. The motion representation is subsequently quantized to $\hat{m}_t$. The quantized illustration is processed by the MV decoder to reconstruct motion data, resulting in $\hat{v}_t$.

In the following step, $\bar{x}_t$ is generated using a motion compensation model, which is designed to leverage the optical flow obtained previously by the MV. Utilizing the previously reconstructed frame and motion vector, $\hat{x}_{t-1}$ and $\hat{v}_t$ respectively, the motion compensation model strives to create a predicted frame $\bar{x}_t$ that closely mirrors the current frame $x_t$, as illustrated in Figure \ref{fig:videocompression2}. At the outset, utilizing the motion information $\hat{v}_t$, the former frame $\hat{x}_{t-1}$ transforms to align it with the current frame. Although motion information $\hat{v}_t$ is used to align the former frame $\hat{x}_{t-1}$ with the current frame, the resulting transformed frame still displays artifacts. To tackle these drawbacks, the transformed frame $w(\hat{x}_{t-1}, \hat{v}_t)$, $\hat{x}_{t-1}$, and $\hat{v}_t$ are merged and passed through an additional CNN, subsequently producing the refined predicted frame $\bar{x}_t$. By adopting a pixel-wise motion compensation method, this technique provides enhanced temporal accuracy and avoids the blockiness artifacts typically associated with conventional block-based motion compensation methods.

The next stages involve transform, quantization, and inverse transform. A residual encoder-decoder model supersedes the linear transform. The remaining $r_t$ undergoes a non-linear transformation, resulting in the output $y_t$, which is then quantized to $\hat{y}_t$. A quantization method is added to ensure this framework is close to an end-to-end framework. To drive the reconstructed residual $\hat{r}_t$, quantized $\hat{y}_t$ is used as the input into the residual decoder model.

In the entropy encoding stage, the compressed motion data $\hat{m}_t$ and the remaining data $\hat{y}_t$ are converted into a binary format and sent to the decoding system for evaluation. To approximate the bit expense of the suggested technique while training, convolutional neural networks are utilized to identify the likelihood distribution for every symbol present in $\hat{m}_t$ and $\hat{y}_t$.

Lastly, the process concludes with frame reconstruction.

 \section{Compression models}\label{chap:Compression models}
 This section comprehensively reviews current deep-learning models developed for video compression. However, it is essential to note that deep learning-based video compression is still in its nascent stage. As such, there is no definitive approach for selecting the optimal solution.
 
\subsection{Autoencoders}
\begin{figure}[h]
\centering
\includegraphics[width=100mm]{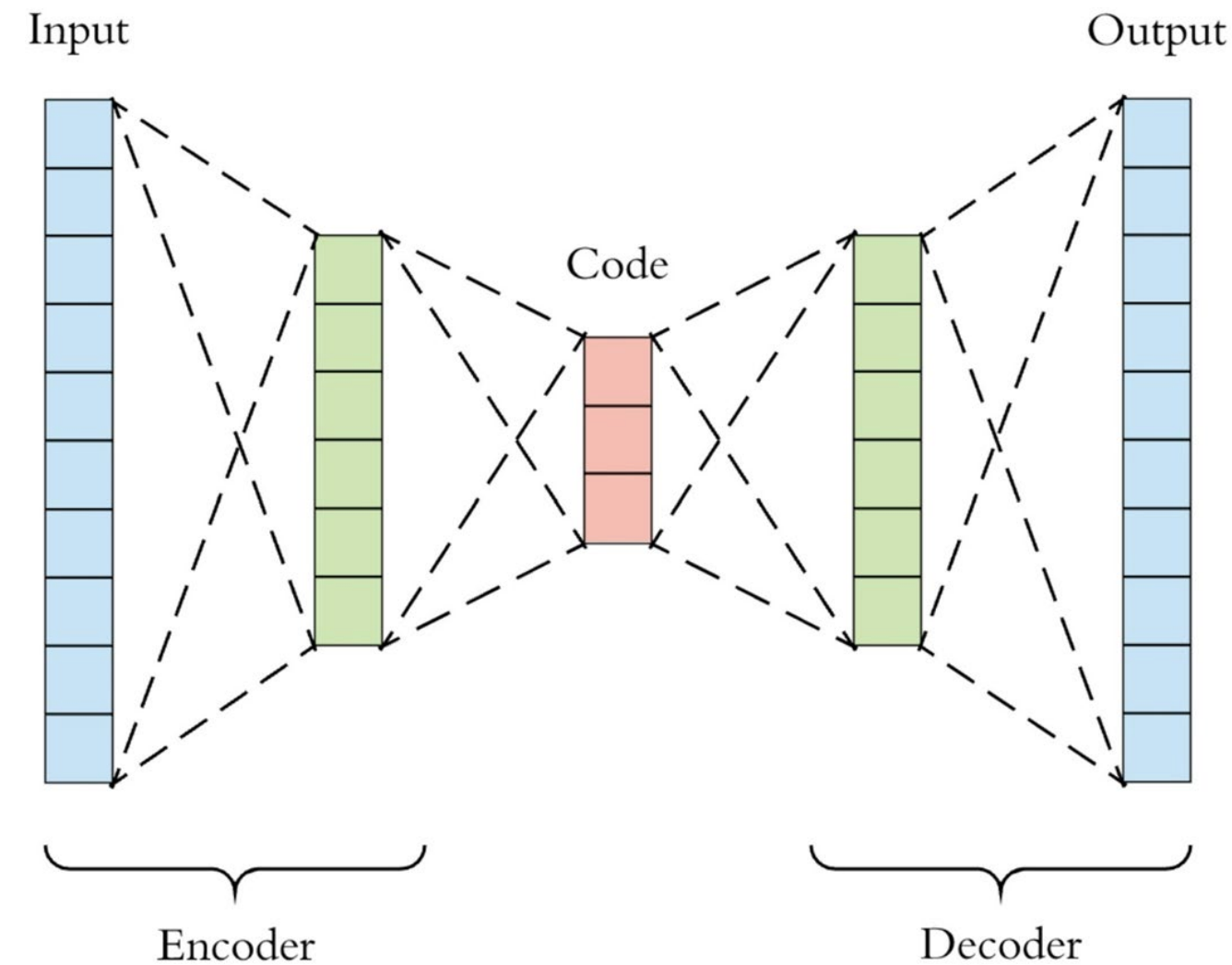}
\caption{Autoencoder architecture scheme from \cite{alkhayrat2020comparative}.}
\label{fig:autoencoder}
\end{figure}

Autoencoders have emerged as a powerful technique for unsupervised learning that allows the discovery of useful representations from raw data \cite{theis2017lossy}. This method aims to acquire a hidden representation of the data that can identify the most significant changes within the initial data. This is achieved without requiring direct supervision. The latent space has lower dimensions than the input data \cite{hinton2006reducing} to prevent the autoencoder from learning the identity transformation. The two central elements of an autoencoder are the encoder and the decoder \cite{goodfellow2016deep}. The encoder generates a lower-dimensional latent space $\mathbf{z}=f(\mathbf{x})$ from the input  $\mathbf{x}$, using one or more layers of non-linearity, as depicted in Figure \ref{fig:autoencoder}. 
\clearpage

The decoder then utilizes the latent space to reconstruct the original data by mapping the hidden representation $\mathbf{z}$ to the output 
\begin{equation}
\hat{\mathbf{x}}=g(\mathbf{z}) 
\end{equation}
 using one or more layers of non-linearity, with parameters optimized to minimize the cost function using the mean squared error 
\begin{equation}
L(\mathbf{x}, g(f(\mathbf{x}))) = \frac{1}{2} ||\hat{\mathbf{x}} - \mathbf{x}||^2. 
\end{equation}

The reconstruction accuracy relies on the latent space learning the most critical variations in the original data. This process minimizes the reconstruction loss by generating an output closer to the initial input \cite{dumas2018autoencoder}. By limiting the dimensionality of the coding layer, the autoencoder is compelled to learn a meaningful representation of the underlying data distribution in order to reconstruct the input accurately. As a result, the autoencoder is prevented from simply copying the input to the output layer, which is also known as the identity function \cite{goodfellow2016deep}. Autoencoders are neural networks and can achieve comparable or even superior performance in terms of compression efficiency and image quality \cite{agustsson2019generative} \cite{balle2016end}.
 The model generation technique in image compression technology is an innovative approach that allows the generation of image data that closely resembles the original image by inputting a small amount of information \cite{li2020anomaly}. One popular generative model used for this purpose is the Variational Autoencoder (VAE) \cite{balle2018variational}. Unlike traditional autoencoders that struggle to generate data autonomously due to unknown output vectors in the hidden layer, VAEs add a hidden variable to the hidden layer to allow for automatic data generation.

 VAEs combine deep learning and statistical learning to effectively reduce the dimensionality of image data through the encoder network and generate similar images using the decoder network. Overall, VAEs are a powerful generative model that leverages the strengths of deep models for nonlinear fitting, making them a valuable tool in image compression technology  \cite{sun2021image}. VAEs employ a probabilistic encoder network to represent the unknown posterior distribution $q(z|x)$, exhibiting a normal distribution. Then the decoder reconstructs the image by modeling  $p(x|z)$. The encoder and decoder networks in a VAE are trained together to optimize a variational lower bound on the log likelihood of the input data \cite{kingma2013auto} \cite{rezende2014stochastic} using parameters $\theta$ and $\phi$ \begin{equation}
\label{equ:vae}
\mathcal{L}(\theta, \phi; x) = \mathbb{E}{q_{\phi}(z|x)}[\log p_{\theta}(x|z)] - D_{KL}[q_{\phi}(z|x)||p(z)], 
\end{equation}
where $D_{KL}$ is the Kullback-Leibler divergence. This divergence factor quantifies the dissimilarity of a pair of probability distributions \cite{hershey2007approximating}. Within the VAE loss function, KL serves to maintain the learned latent distribution's proximity to a predetermined prior distribution, ultimately resulting in a richer and more transferable latent space depiction.

\subsection{Frame interpolation}

\begin{figure}[h]
\centering
\includegraphics[width=100mm]{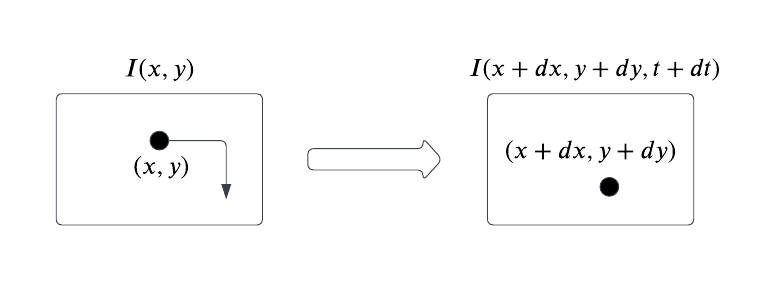}
\caption{Optical flow for a pixel located at $(x, y)$. }
\label{fig:opticalflow}
\end{figure}

Frame interpolation is a technique commonly used in video processing to create new frames from temporal neighbors to fill in the gaps between existing frames \cite{wu2018video} \cite{niklaus2017video}. Motion estimation using optical flow and pixel synthesis are the main steps for frame interpolation \cite{baker2011database} \cite{werlberger2011optical}. Optical flow refers to the movement pattern present between two consecutive frames within an image series. Analyzing the motion of each pixel at a granular level is crucial for better understanding this motion. As depicted in Figure \ref{fig:opticalflow}, a pixel situated at position $(x,y)$ during time t transitions to the location $(x + dx, y + dy)$ within the subsequent frame; in this situation, optical flow is represented by the vector $f(x) = (u, v)$, where $u$ corresponds to the movement along the x-axis and $v$ represents motion along the y-axis. The objective of motion estimation by optical flow is to find these motion vectors for each pixel. To address this problem, as mentioned in \cite{beauchemin1995computation}, the process starts by defining the pixel intensity function $I(x,y,t)$. Knowing the fact that pixel intensity does not change with motion, the problem can be solved using 

\begin{equation} \label{equ:opticalflow1} I(x,y,t) = I(x+dx,y+dy,t+dt). \end{equation}
Calculating the Taylor series of \ref{equ:opticalflow1} and dividing it by $dt$ results in
\begin{equation}\label{equ:opticalflow2} f_x u + f_y v + f_t = 0, \end{equation}

where $f_x = \frac{\partial f}{\partial x}$, $f_y = \frac{\partial f}{\partial y}$ represent the partial derivatives of the optical flow function $f$ with respect to the spatial coordinates $x$ and $y$, and $f_t$ represents the partial derivative with respect to time $t$.
The optical flow vector $(u,v)$ can be expressed 
\begin{equation} \label{equ:opticalflow4} u = \frac{dx}{dt} , v = \frac{dy}{dt}. \end{equation}
To find $(u,v)$, numerous approaches have been suggested, including  "Horn and Schunck" \cite{horn1981determining} and "Lucas and Kanade" \cite{bruhn2005lucas}, \cite{barron1994performance}. Once the optical flow is calculated, the original input frames are then warped or deformed based on the estimated flow and blended together to generate an interpolated frame \cite{liu2017video}. In contrast to methods that use motion estimation, phase-based techniques are emerging as promising alternatives \cite{niklaus2017video}. These methods show great potential for use in video processing. Phase-based techniques encode motion by taking advantage of the phase difference between frames \cite{meyer2015phase}. They accomplish this by representing the motion through the phase shift of individual pixels. The phase shift denotes the spatial movement between two shifted functions, characterized by the variation in their corresponding phase values. In other words, the phase shift quantifies the extent to which one function has been translated in relation to another based on the change in their phase values. This concept is employed in phase-based approaches to represent motion in signals, such as images or video frames, allowing for the generation of interpolated frames by adjusting the phase values of the original frames \cite{meyer2015phase}. These techniques have several potential applications in video processing and are emerging as a promising method for enhancing video quality. This technique improves motion magnification \cite{wadhwa2013phase} and view expansion \cite{didyk2013joint}. Although this approach can achieve impressive results, it may struggle to maintain high-frequency details in videos where temporal changes are significant \cite{meyer2015phase}. Recently, phase-based methods have emerged as a potential alternative to motion estimation-based methods for video processing.  Different methods, such as generating a flow field, estimating a convolution kernel that varies spatially, or combining predictions from both past and future, can be used for frame interpolation \cite{wu2018video} \cite{niklaus2017video}.

\begin{figure}[h]
\centering
\includegraphics[width=100mm]{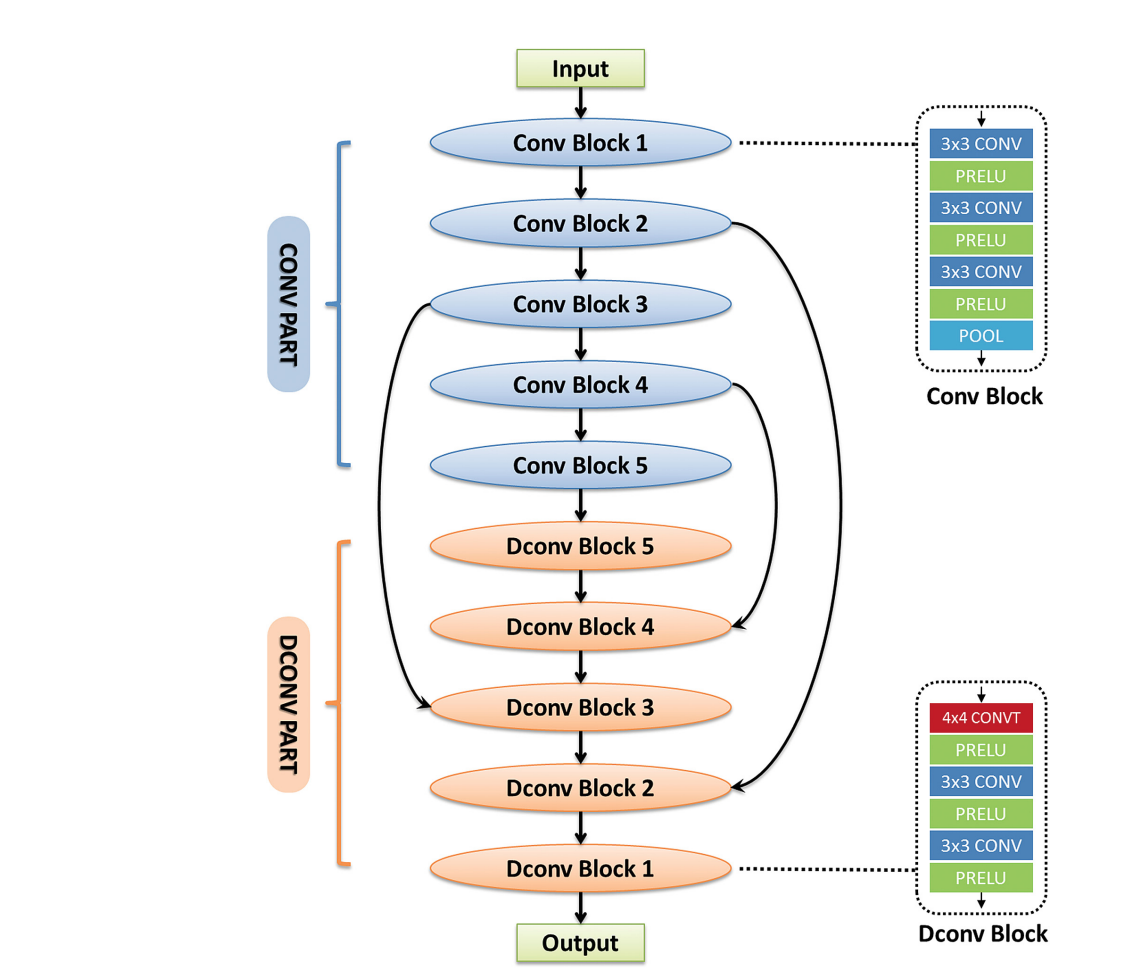}
\caption{Frame interpolation using CNN from \cite{long2016learning}.}
\label{fig:cnninetpolation}
\end{figure}

However, the accuracy of interpolation decreases as the time intervals between frames increase, making the process more challenging. To achieve frame interpolation, neural networks can be utilized to interpolate video frames temporally and encode the residuals \cite{djelouah2019neural}. Convolutional neural networks (CNNs) have the potential to predict optical flow \cite{dosovitskiy2015flownet}, indicating their ability to comprehend temporal motion. One example of using CNN for frame interpolation is the work of Long et al. \cite{long2016learning}. The model architecture includes convolutional and deconvolutional components, as illustrated in Figure \ref{fig:cnninetpolation}. The Convolution Block takes an input, performs a series of convolutional operations followed by activation using a parametric rectified linear unit (PRELU), and then performs pooling to produce an output. In the deconvolution block, the input data is first passed through a convolution transpose layer, followed by a PRELU. The output is then passed through two convolution layers, each followed by a PRELU activation. These operations aim to upsample the input data to a higher resolution while simultaneously maintaining intricate image features in the interpolated frame.  This model has been trained using KITTI RAW \cite{geiger2013vision} video sequences. Training triplets, composed of every three sequential frames, are employed for model training. The first and third frames serve the role of inputs, while the second frame functions as the output. This model is optimized using the charbonnier loss function $\rho(x) = \sqrt{x^2 + \epsilon^2}$,  with $\epsilon = 0.1$ \cite{sun2014quantitative}.

\subsection{Motion compensation via optical flow}
Motion information plays a crucial role in video compression techniques as it helps to eliminate temporal redundancy in video sequences. A method to address this problem involves using optical flow, a learning-based technique that captures motion information \cite{wadhwa2013phase}. Optical flow describes the 2D velocity field that illustrates the apparent motion within the image caused by independent object movements in the scene or observer motion \cite{barron1994performance}. Determining optical flow can be a demanding undertaking that is affected by multiple factors. These include occlusions, varying lighting conditions, and scene changes. As a result, it is often necessary to use robust techniques that can handle these challenges.
Compression techniques involve estimating and compressing the movement of objects in a video to create a prediction of the current frame. This prediction is created by applying a warping method to a previously decoded frame \cite{agustsson2020scale}. The difference between this prediction and the actual frame is compressed to reduce the overall distortion in the video. However, the motion-compensation method is likely to generate imprecise approximations of the target frame due to its inability to model intricate motion patterns. The residual signal that the algorithm is incapable of forecasting is then encoded \cite{gallego2018unifying}. The development of the FlowNet architecture \cite{dosovitskiy2015flownet} introduced a new paradigm in optical flow estimation, employing a CNN to acquire optical flow from the data. Despite initial challenges in competing with established methods, further advancements and refinements, such as FlowNet2 \cite{ilg2017flownet}, considerably improved its performance. By stacking multiple FlowNet modules, FlowNet2 outperformed numerous traditional approaches \cite{ilg2017flownet}. However, its extensive model design demanded considerable memory, rendering it unsuitable for mobile and embedded devices \cite{shah2021traditional}. As a result, subsequent research aimed at developing lightweight modules without sacrificing accuracy. This was achieved by incorporating well-known concepts from traditional methods into deep learning frameworks, as demonstrated in SpyNet \cite{ranjan2017optical}. 

\begin{figure}[h]
\centering
\includegraphics[width=100mm]{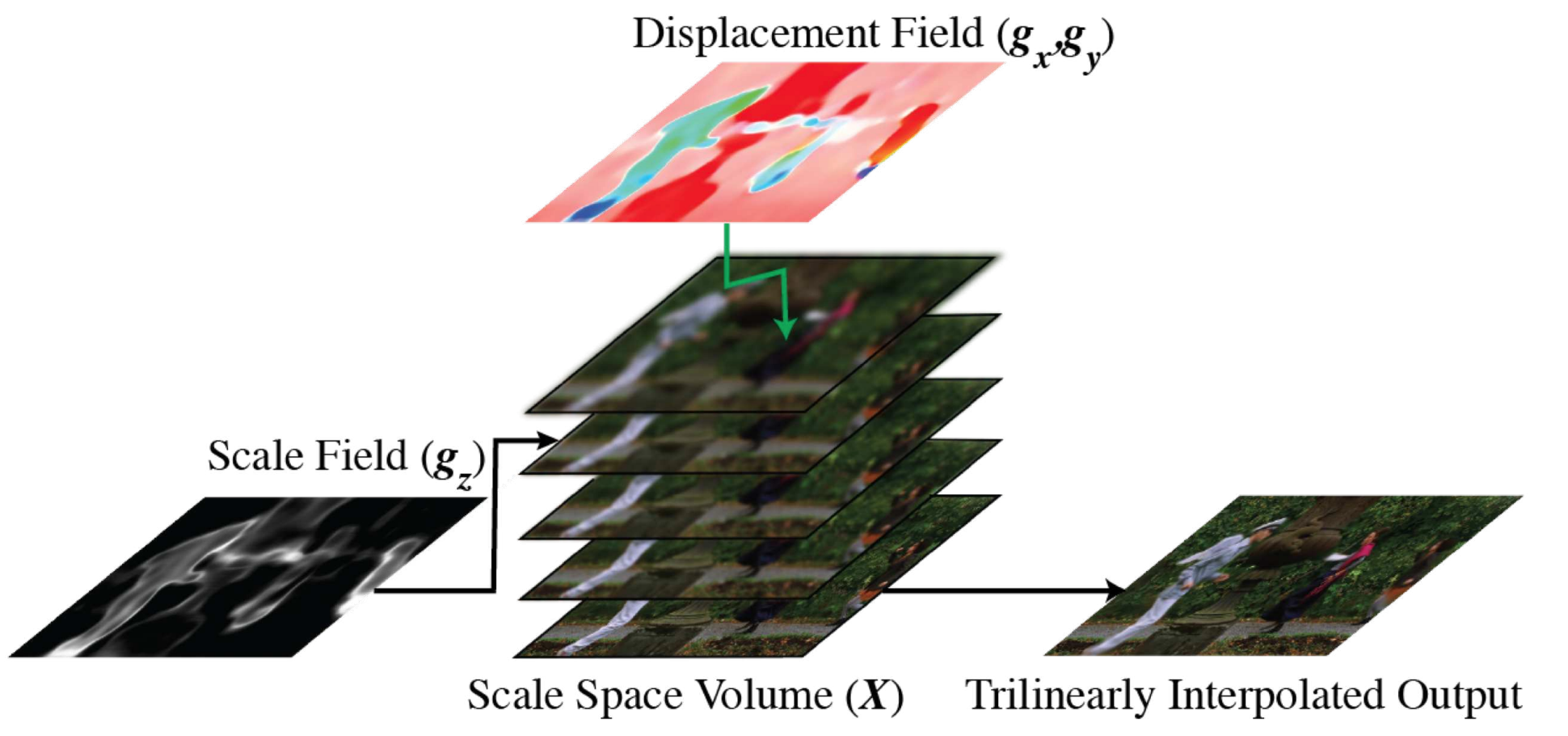}
\caption{Scale space wraping from \cite{agustsson2020scale}.}
\label{fig:ssw}
\end{figure}

 \section{Scale-space flow network} \label{chap:Scale-space flow}

The motion compensation via the optical flow method, one of the deep learning-based compression models discussed earlier, has its drawbacks. It relies on complex architectures and training schemes, making it more challenging to develop and implement. To address these limitations, E. Agustsson et al. proposed a scale-space flow model for end-to-end optimized video compression \cite{agustsson2020scale}. This method integrates a scale field into the traditional 2-channel flow field, leading to the formation of scale-space flow and scale-space warping. By employing an adaptive blurring procedure based on the prediction's accuracy, the model reduces intermediate residual error and enhances the efficiency of residual compression.

The proposed method for transforming images involves creating a volume of different versions of the same image, called a scale-space volume, as depicted in Figure \ref{fig:ssw}. Utilizing the suggested module, an image $x$ is transformed into a uniform-resolution volume $X$. This approach differs from bilinear warping, which generates the warped output by directly using a displacement field $(f_x, f_y)$ to process the 2D source image. Instead, the innovative approach leverages trilinear sampling, utilizing a 3-channel displacement field in the 3D scale-space domain $(g_x, g_y, g_z)$.

The primary focus of the scale-space flow model is to process input data that consists of a sequence of frames. The process begins by encoding the first frame to a latent space $[z_0]$ and obtaining a reconstruction $\hat{x}_0$. Then, for the current frame, the wrapped latents $[w_i]$ are encoded and estimated for decoding a scale-space flow using a network $g_i = g([w_i])$. The current frame is approximated by deforming the previous reconstruction using the scale-space flow. To improve the accuracy, the difference between the estimate and the actual frame is encoded as a latent $[v_i]$ and applied to the estimate to get a final reconstruction $\hat{x}_i = \bar{x}_i + \hat{v}_i$.

\begin{figure}[h]
\centering
\includegraphics[width=100mm]{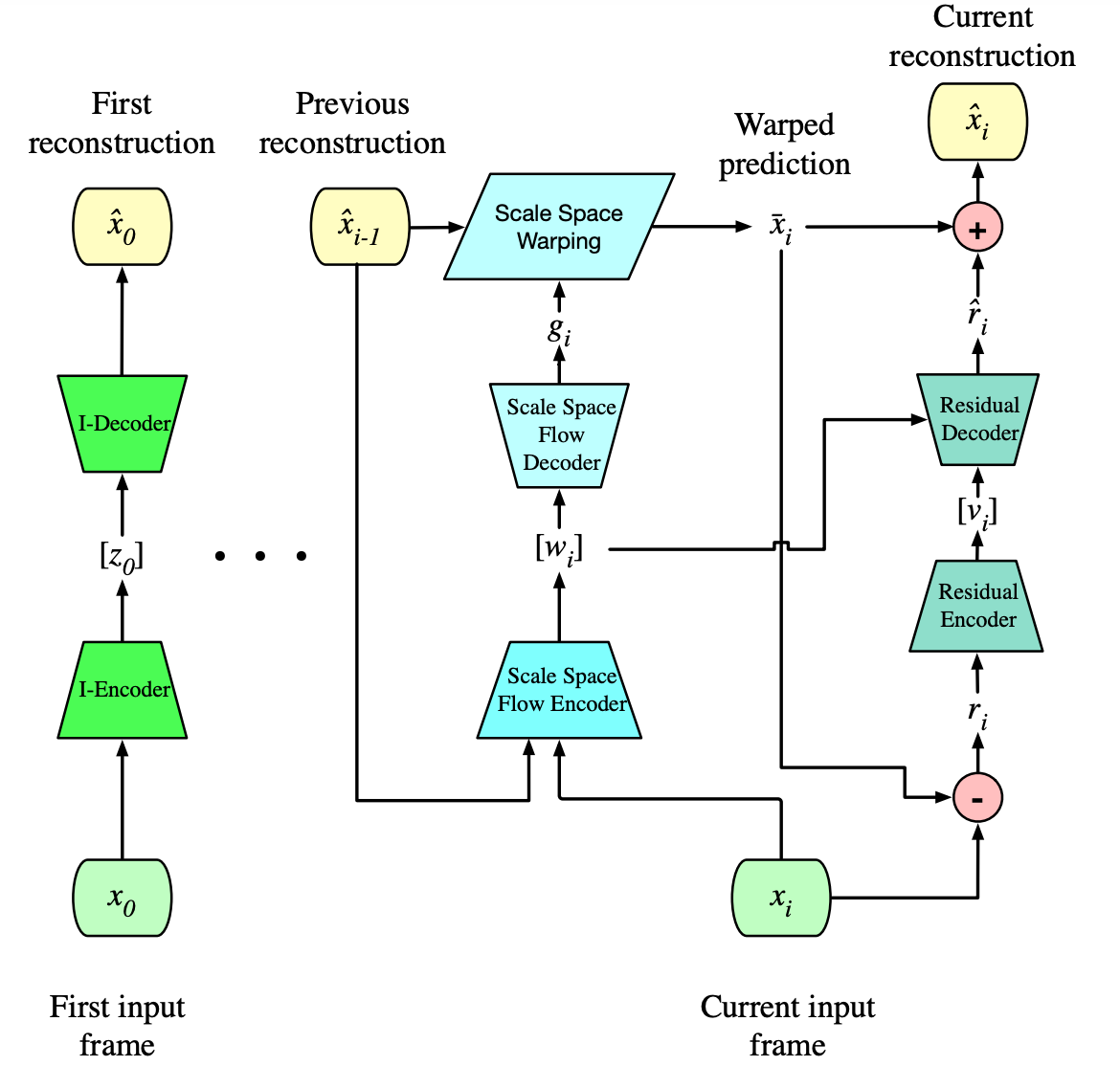}
\caption{Scale space model architecture from \cite{agustsson2020scale}.}
\label{fig:sswm}
\end{figure}

Figure \ref{fig:sswm} illustrates how scale-space warping is used in a compression system. To model the density of each of latents, $z_0$, $v_i$, and $w_i$,  a distinct hyperprior \cite{minnen2018joint} \cite{balle2018variational} is used. The hyperprior is an end-to-end image compression model that uses a VAE that trains a  prior simultaneously with an autoencoder, resulting in improved rate-distortion based on PSNR \cite{balle2018variational}. The models were trained on approximately 700,000 $(1920 \times 1080)$ YouTube videos at a 30Hz frame rate. The training process involved nearly 278 hours of video, with a batch size of 8 frames and an average frame rate of 30 FPS. The training dataset comprised 60 consecutive frames from each video sequence, partitioned into temporal chunks of 3 frames each. According to the evaluations, the scale space flow model surpasses H.264 and HEVC, along with top-performing learning-based methods \cite{wu2018video} \cite{lu2019dvc} \cite{habibian2019video}.

    \chapter{360-degree Projection Formats}
\label{chap: 360-degree Projection Formats}

\begin{figure}[h]
\centering
\includegraphics[width=100mm]{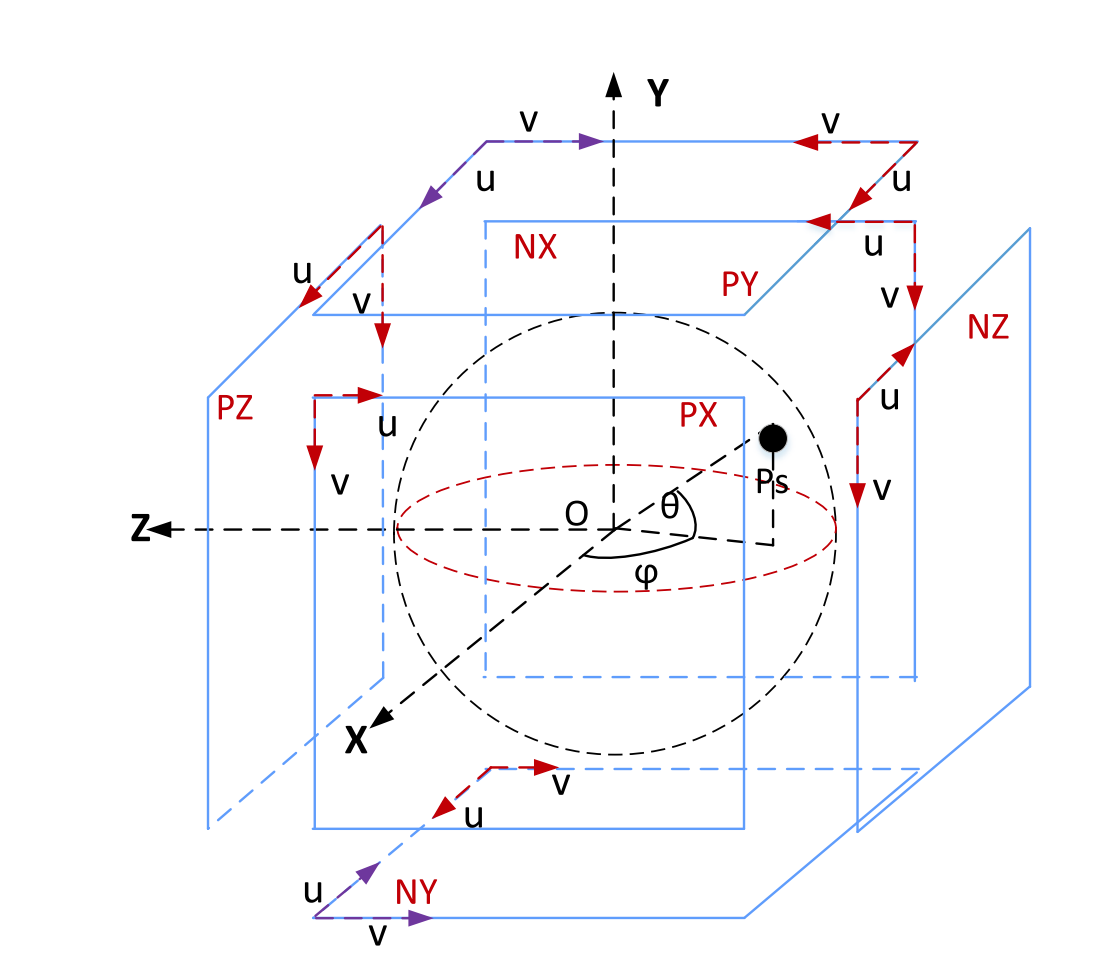}
\caption{X, Y, Z cordinate system used in 360lib app from \cite{ye2017jvet}.}
\label{fig:coordinates}
\end{figure}

 360-degree video is a way to provide an immersive viewing experience through virtual and augmented reality applications. Nonetheless, standard video coding techniques are not helpful for these types of video, so it is essential to project 360-degree content to a 2D plane before encoding \cite{xiu2017evaluation}. The efficiency of 360-degree video encoding can be enhanced by optimizing the coding techniques in line with the projection method employed. However, the projection process may generate redundancy or shape distortion issues \cite{hussain2021evaluation}. Therefore, it is essential to compare different projections to design improved 360-degree image and video services. This is critical for ensuring users experience better quality and efficient use of resources. This section briefly illustrates the most commonly used projection formats provided by the 360Lib app \cite{he2017360lib} \cite{ye2017jvet} \cite{podborski2017virtual}.

360Lib employed the right-hand coordinate system as depicted in Figure \ref{fig:coordinates} \cite{ye2017jvet}.

The coordinates on the unit sphere (X, Y, Z) are deduced from the longitude values $\phi \in [-\pi, \pi]$ and latitude $\theta \in [-\pi/2, \pi/2]$ from \begin{equation}
\label{equ:coordinates}
(X, Y, Z)=\begin{cases}
			X = \cos(\theta)\cos(\phi) \\
                Y = \sin(\theta)  \\
                Z = -\cos(\theta)\sin(\phi) \\ 
		 \end{cases}.
\end{equation}
Longitude is defined as the angle that starts from the X-axis in a counterclockwise direction, while latitude is the angle from the equator toward the Y-axis.

Each projection is defined by the function that applies the conversion from (f, m, n) to (X, Y, Z) and vice versa, where f is the face index which shows the number of faces that every projection possesses a mapping position represented by column and row coordinates denoted as m and n, respectively For each face in the 2D projection plane, a 2D plane coordinate system is established, and a face index is defined to generalize the 2D coordinate system. The image sampling grid is represented in a u-v plane as depicted in Figure \ref{fig:coordinates}, where the position of a sampling point is specified by (m, n) \cite{ye2017jvet}.

\section{Equi-rectangular projection format }
\begin{figure}[h]
\centering
\includegraphics[width=100mm]{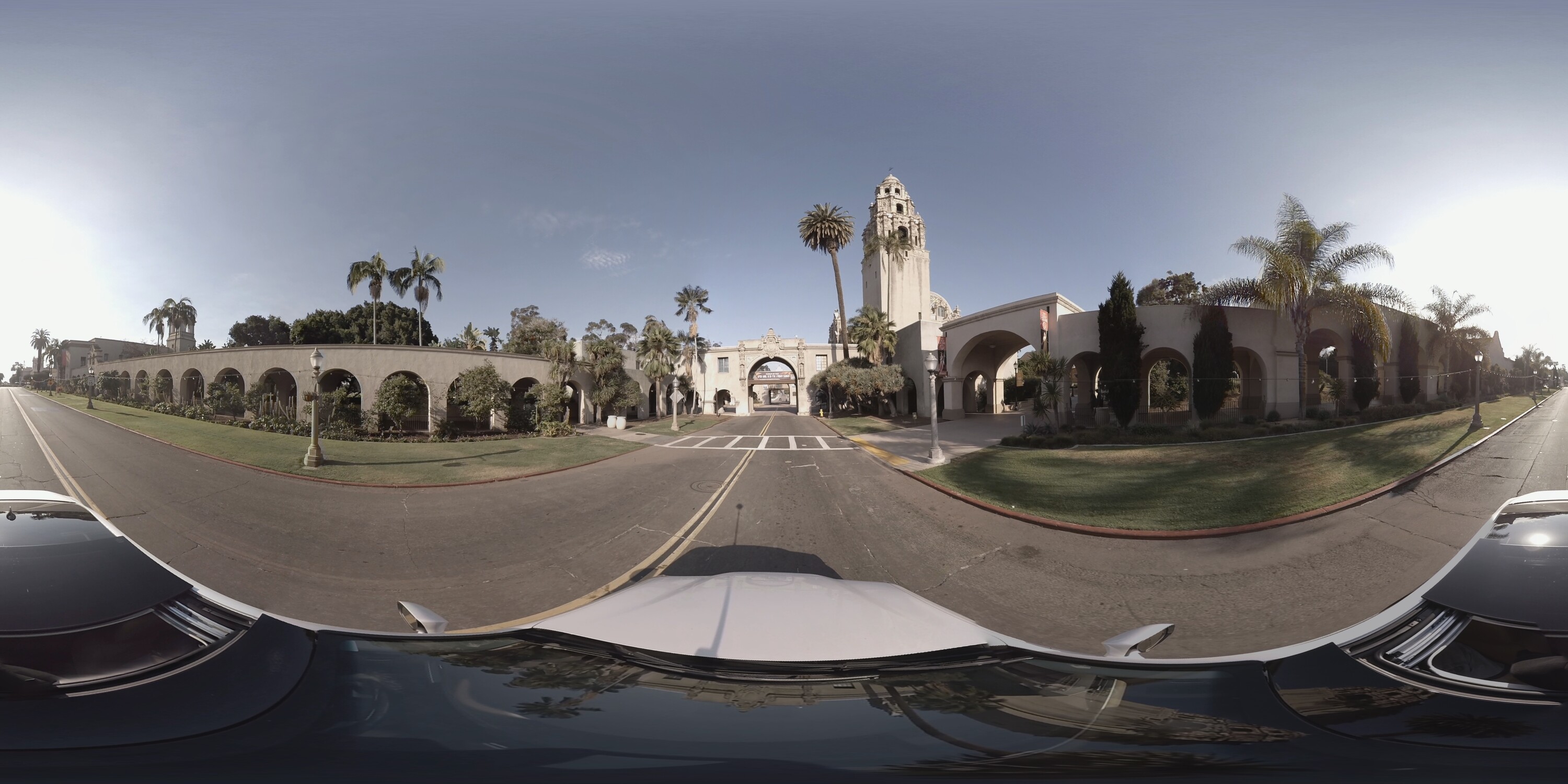}
\caption{Balboa JVET sequence in ERP format.}
\label{fig:erp_balboa}
\end{figure}

Since the Equi-rectangular projection (ERP) has only one face, the face index, denoted by f, is zero. Given the height and width of the face as H and W respectively, the mapping position from a 2D plane with coordinates (m,n) to a 3D plane is computed by finding the corresponding position on the (u,v) plane using \begin{equation}
\label{equ:erp1}
(u, v)=\begin{cases}
			u = \frac{m + 0.5}{W}, \quad 0 \leq m < W \quad   \\
                v = \frac{n + 0.5}{H}, \quad 0 \leq n < H \quad
		 \end{cases}.
          \end{equation}
By utilizing these coordinates, the longitude and latitude are calculated from \begin{equation}
\label{equ:erplan}
\phi = (u - 0.5) \cdot (2\pi) \quad ,
\theta = (0.5 - v) \cdot \pi \quad, 
\end{equation} and finally, the coordinates (X, Y, Z) are obtained using (\ref{equ:coordinates}). For converting 3D to 2D, the longitude and latitude are calculated using (\ref{equ:coordinates}), and then the (u,v) coordinates are obtained using (\ref{equ:erplan}) \cite{ye2017jvet}.
Figure \ref{fig:erp_balboa} depicts a JVET sequence in ERP format.

\begin{figure}[h]
\centering
\includegraphics[width=150mm]{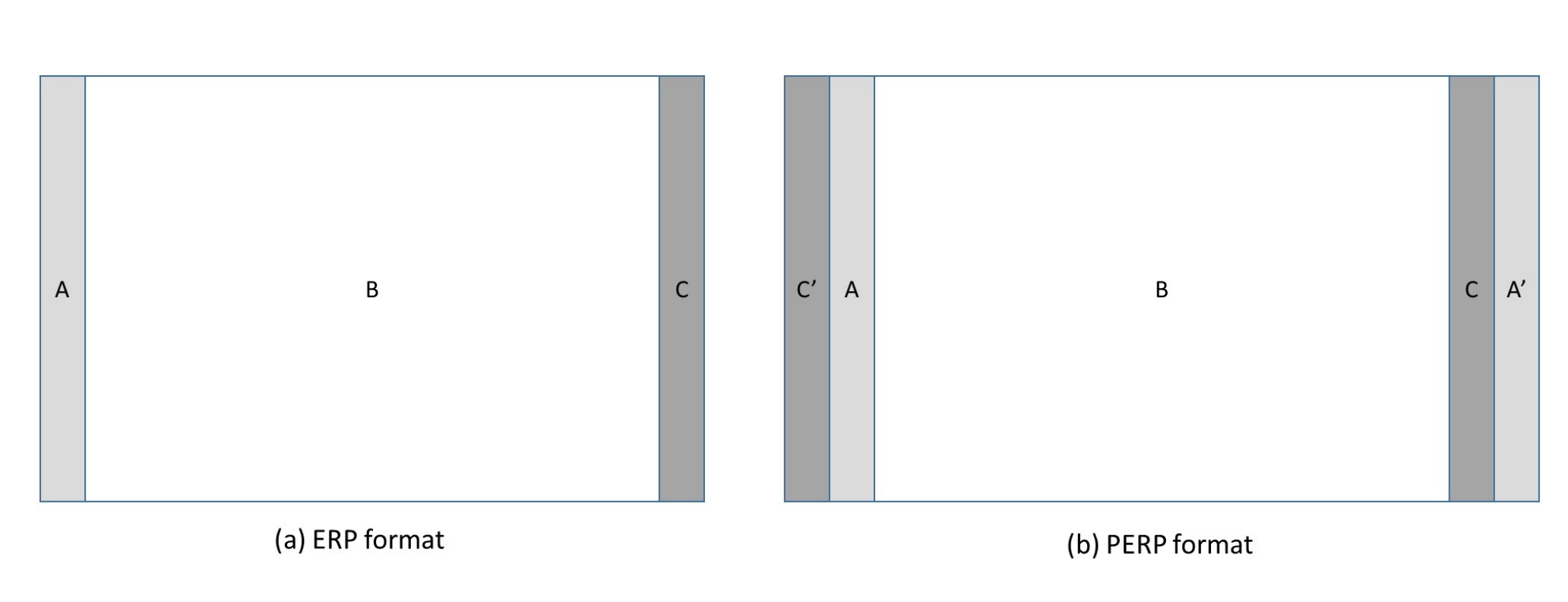}
\caption{ERP to PERP conversion from \cite{ye2019omnidirectional}.}
\label{fig:padded_erp}
\end{figure}

\section{Padded Equi-rectangular projection format}

\begin{figure}[h]
\centering
\includegraphics[width=100mm]{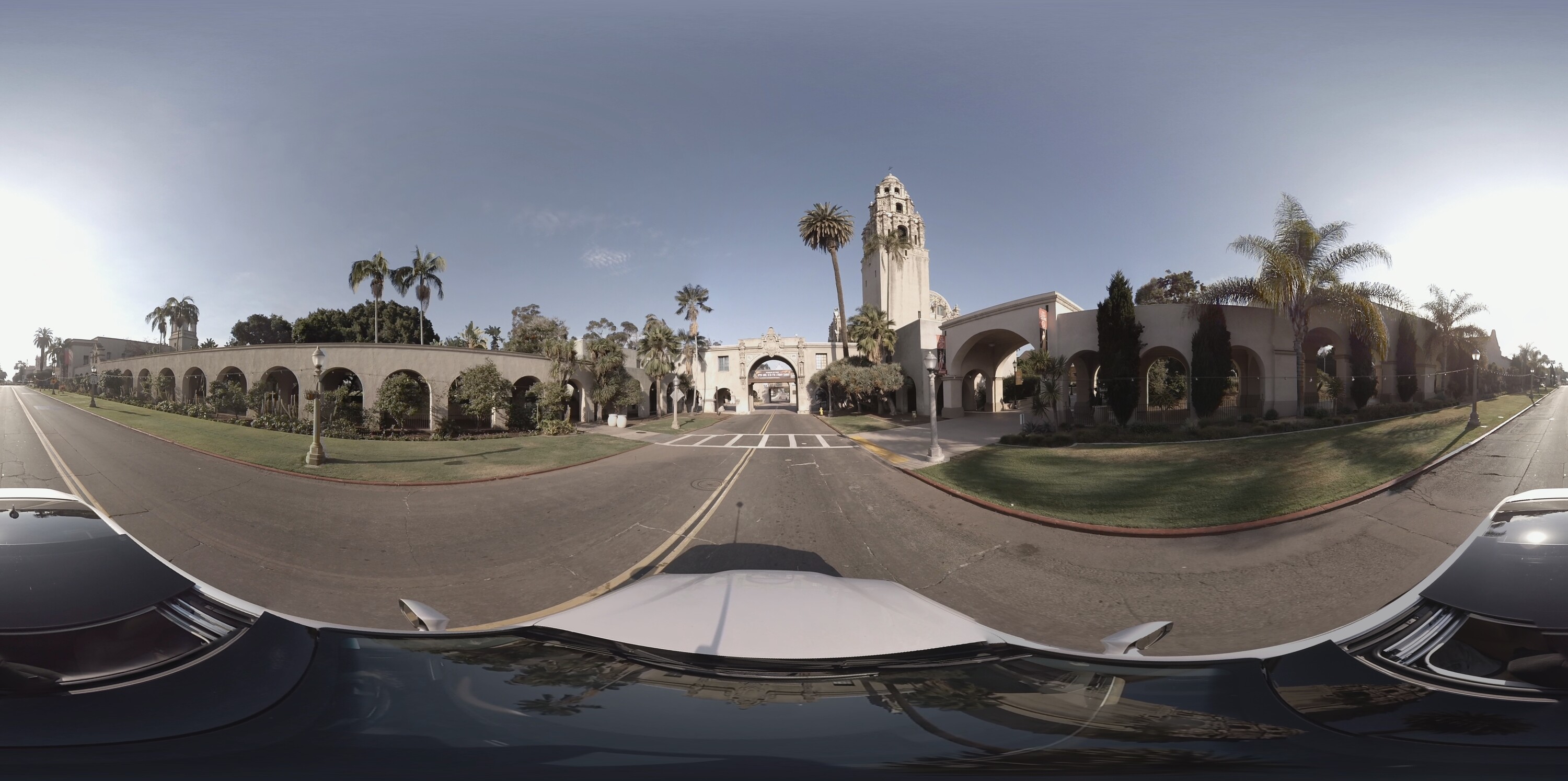}
\caption{Balboa JVET sequence in PERP format.}
\label{fig:perp_balboa}
\end{figure}

Converting 3D content to a 2D format can lead to discontinuities if the spherical representation samples do not align with the coded projection format. These discontinuities can manifest as a visible artificial line along the left and right extremities of the converted ERP frame \cite{ye2019omnidirectional}. To minimize the occurrence of seam artifacts in reconstructed viewports, it is suggested to duplicate samples across discontinuous edge boundaries, resulting in a padded ERP (PERP) format. Adding extra space around the edges of an image can make it look better by hiding the lines where different parts of the image are put together \cite{boyce2017ee4}. This projection format requires copying pixels from the left side of the image and pasting them onto the right side of the image to create padding as shown in Figure \ref{fig:padded_erp} \cite{ye2019omnidirectional}. Figure \ref{fig:perp_balboa} depicts a JVET sequence in PERP format.

\section{Cubemap projection format}
\begin{figure}[h]
\centering
\includegraphics[width=100mm]{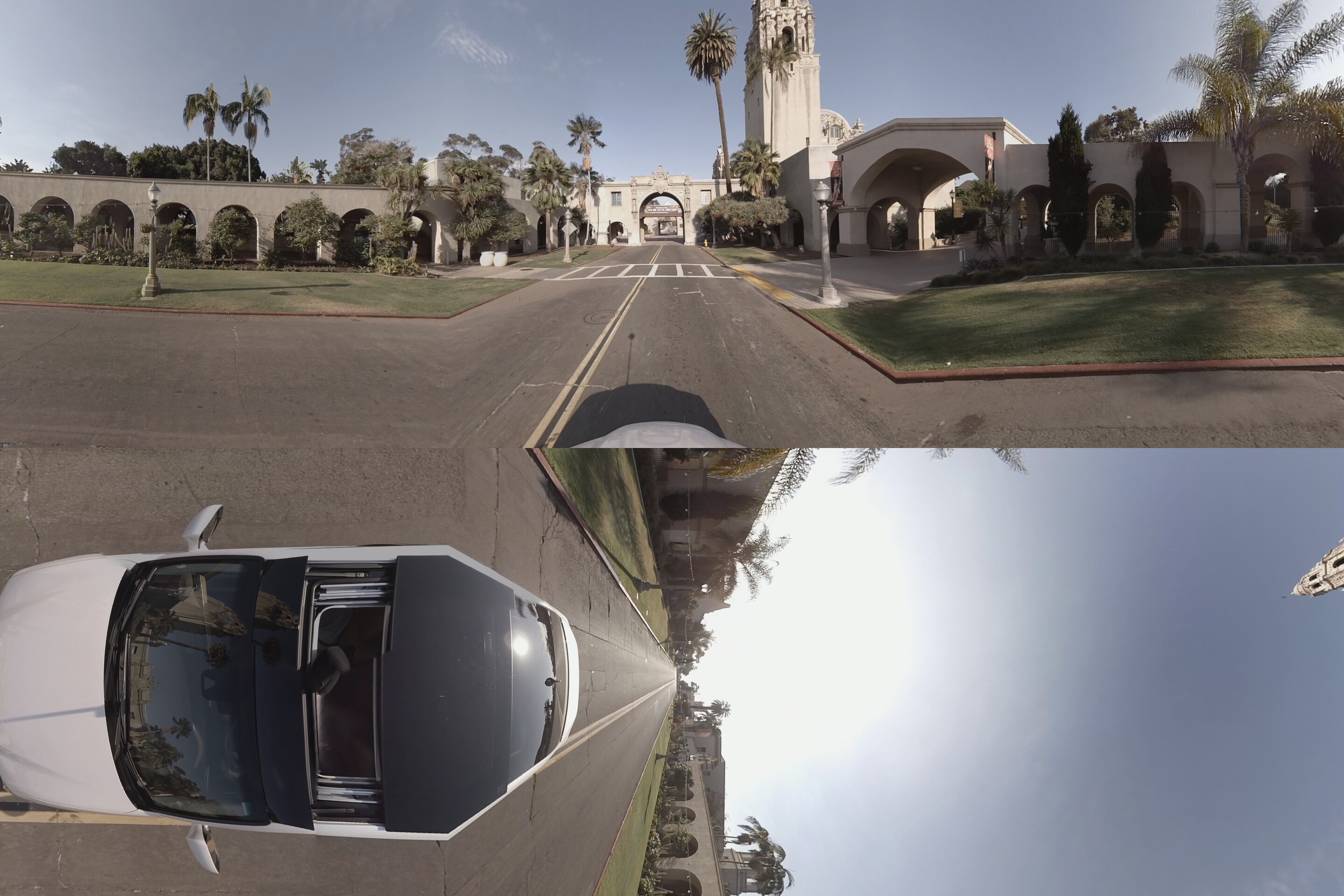}
\caption{Balboa JVET sequence in CMP format.}
\label{fig:cmp_balboa}
\end{figure}

The Cubemap projection (CMP) has six square faces, each of these faces in the (u, v) plane is a $2 \times 2$ square. Given  $A \times A$ as the dimension of each face, the mapping position from a 2D plane with coordinates (m,n) to a 3D plane is computed by finding the corresponding position on the (u,v) plane using 
\begin{equation}
\label{equ:cmp1}
(u, v)=\begin{cases}
			u = \frac{2(m + 0.5)}{A} - 1, \quad 0 \leq m < A    \\
                v = \frac{2(n + 0.5)}{A} - 1, \quad 0 \leq n < A  

		 \end{cases}.
\end{equation}Then the coordinates (X, Y, Z) are obtained separately using Table \ref{table:cmp} for each face \cite{ye2017jvet}. The uneven distribution of samples on the faces is the main drawback of CMP, with a more elevated concentration around the edges and a lower concentration around the center. This artifact has a negative impact on the efficiency of video representation and coding performance \cite{duanmu2018hybrid}. To overcome this issue, various variants of CMP have been proposed and will be discussed below. Figure \ref{fig:cmp_balboa} depicts a JVET sequence in CMP format.

\begin{table}[t]
\centering
\caption{CMP 2D to 3D mapping}
\begin{tabular}{|c|c|c|c|}

\hline
f & X & Y & Z \\
\hline
0 & 1.0 & $-v$ & $-u$ \\
1 & $-1.0$ & $-v$ & $u$ \\
2 & $u$ & 1.0 & $v$ \\
3 & $u$ & $-1.0$ & $-v$ \\
4 & $u$ & $-v$ & 1.0 \\
5 & $u$ & $v$ & $-1.0$ \\
\hline
\end{tabular}

\label{table:cmp}
\end{table}

\section{Equi-angular cubemap projection format}
Equi-angular cubemap (EAC) \cite{zhou2017ahg8} addresses the non-uniform sampling issue in CMP by implementing an additional coordinate mapping function, this method reallocates samples to achieve a more uniform distribution across the faces. To convert CMP to EAC, the $f(u,v)$ function is utilized from

\begin{equation}
\label{equ:eac1}
f(u, v)=\begin{cases}
			f_u = \tan\left(\frac{u \cdot \pi}{4}\right)    \\
                f_v = \tan\left(\frac{v \cdot \pi}{4}\right)  

		 \end{cases},
\end{equation}
 and for applying the reverse conversion the function  $g(u, v)$ is used as

\begin{equation}
\label{equ:eac2}
g(u, v)=\begin{cases}
			g_u  = \frac{4}{\pi} \arctan(u)    \\
                g_v = \frac{4}{\pi} \arctan(v)   

		 \end{cases}.
\end{equation}

\begin{figure}[h]
\centering
\includegraphics[width=100mm]{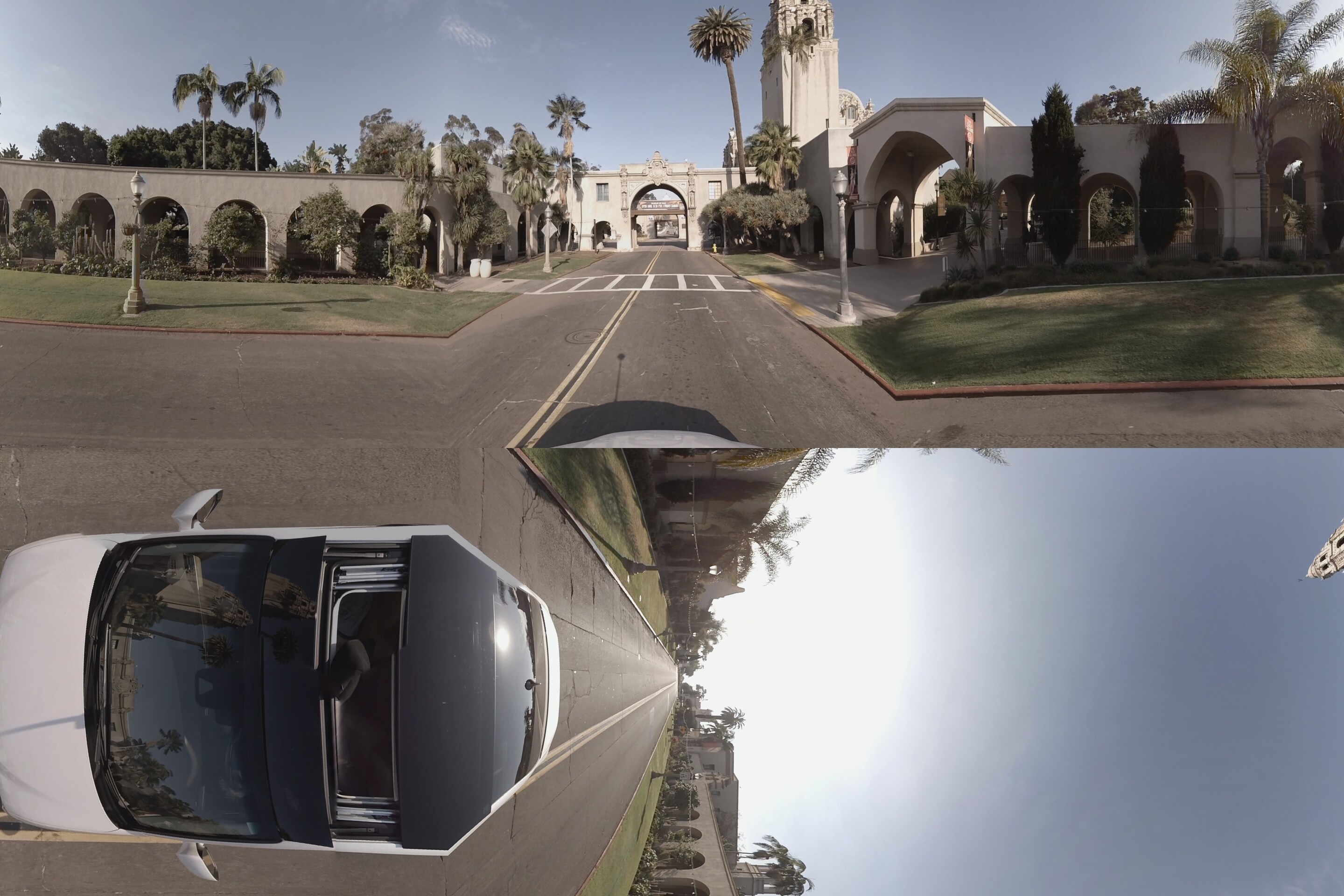}
\caption{Balboa JVET sequence in EAC format.}
\label{fig:eac_balboa}
\end{figure}

These functions are applied to both horizontal and vertical dimensions (u,v) as indicated in Figure \ref{fig:coordinates}. One of the benefits of EAC is that it maintains an approximately equal angle between two samples on a face in the spherical domain, regardless of their position on the face \cite{ye2019omnidirectional}. Figure \ref{fig:eac_balboa} depicts a JVET sequence in EAC format.

\section{Adjusted cubemap projection format}
The Adjusted Cubemap (ACP) format \cite{coban2017ahg8} addresses the non-uniform sampling issue in CMP by using quadratic functions to approximate the mapping functions. \\ 
The resulting sampling distribution is similar to EAC. To convert CMP to ACP, the $f(u,v)$ function is utilized from
\begin{equation}
\label{equ:acp1}
f(u, v)=\begin{cases}
			f_u = \mathrm{sgn}(u) \frac{0.34 - \sqrt{0.342 - 0.09|u|}}{0.18}      \\
                f_v = \mathrm{sgn}(v) \frac{0.34 - \sqrt{0.342 - 0.09|v|}}{0.18}

		 \end{cases},
\end{equation}
 and for applying the reverse conversion the function  $g(u, v)$ is used as

\begin{equation}
\label{equ:acp2}
g(u, v)=\begin{cases}
			g_u  = \mathrm{sgn}(u)\left(-0.36u^2 + 1.36|u|\right)    \\
                g_v = \mathrm{sgn}(v)\left(-0.36v^2 + 1.36|v|\right)  

		 \end{cases}.
\end{equation}

\begin{figure}[h]
\centering
\includegraphics[width=100mm]{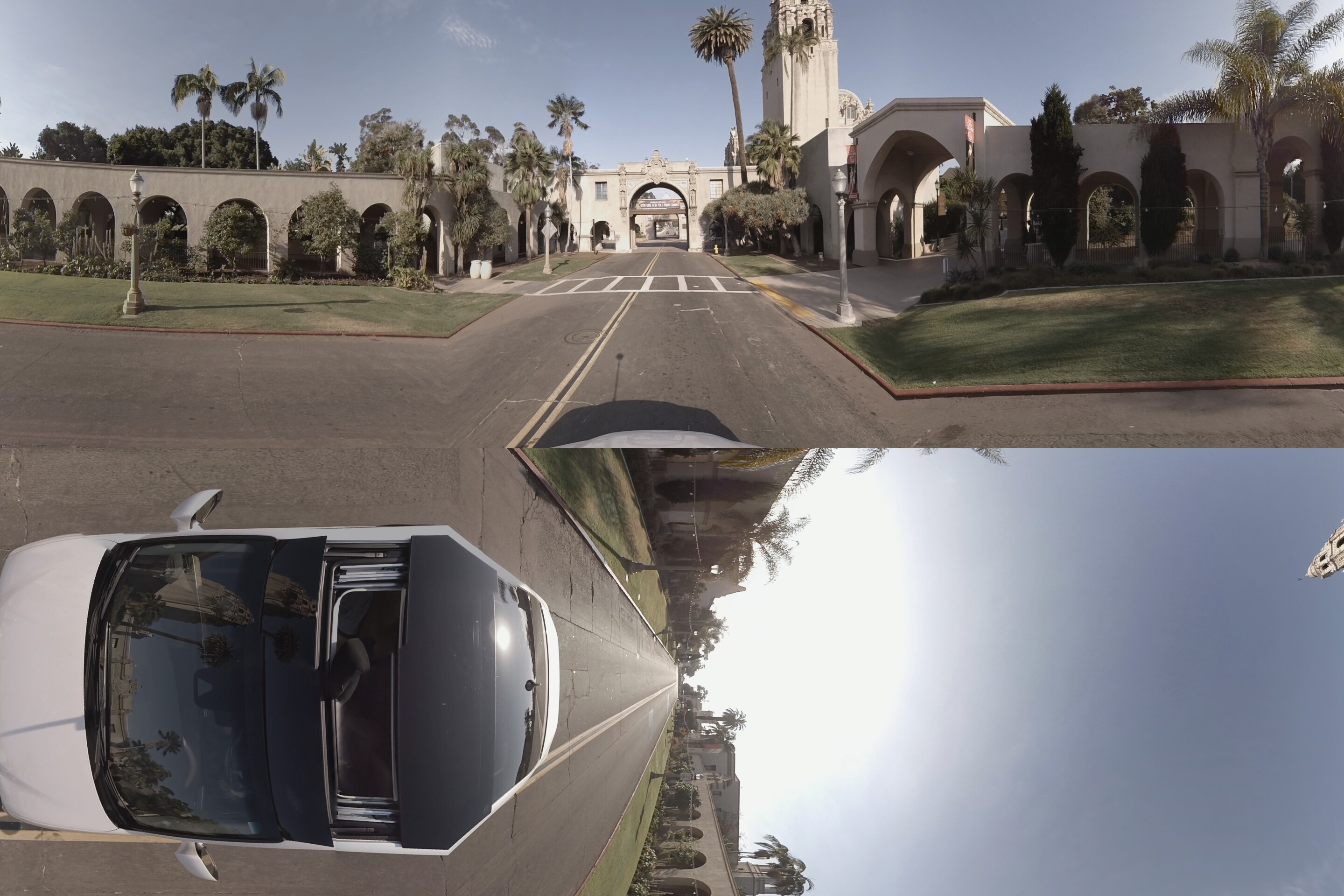}
\caption{Balboa JVET sequence in ACP format.}
\label{fig:acp_balboa}
\end{figure}

These functions are applied to both horizontal and vertical dimensions (u,v) as indicated in Figure \ref{fig:coordinates} \cite{ye2019omnidirectional}. Figure \ref{fig:acp_balboa} depicts a JVET sequence in ACP format.

\section{Hybrid Equi-angular cubemap projection format} 
Hybrid Equi-angular Cubemap (HEC) format \cite{duanmu2018hybrid} tries to further improve the issues in EAC by applying a new coordinate mapping function. The main drawback in EAC is that straight lines get distorted. HEC \cite{lin2019efficient} adds a mapping function $f_{v} (u,v)$ for the front, left, right, and back faces as
\begin{equation}
\label{equ:hec1}
f_{v}(u, v)=\frac{v}{1+0.4(1-u^2)(1-v^2)},
\end{equation}
while the u-axis remain unchanged 
\begin{equation}
\label{equ:hec2}
g_{v}(u, v)=
\begin{cases}
v, & \text{if } t=0 \\
\frac{1-\sqrt{1-4t(v-t)}}{2t}, & \text{otherwise}
\end{cases},
\end{equation}

where 

\begin{equation}
\label{equ:hec3}
t = 0.4v(g_{u}(u)^2 - 1).
\end{equation}

\begin{figure}[t]
\centering
\includegraphics[width=100mm]{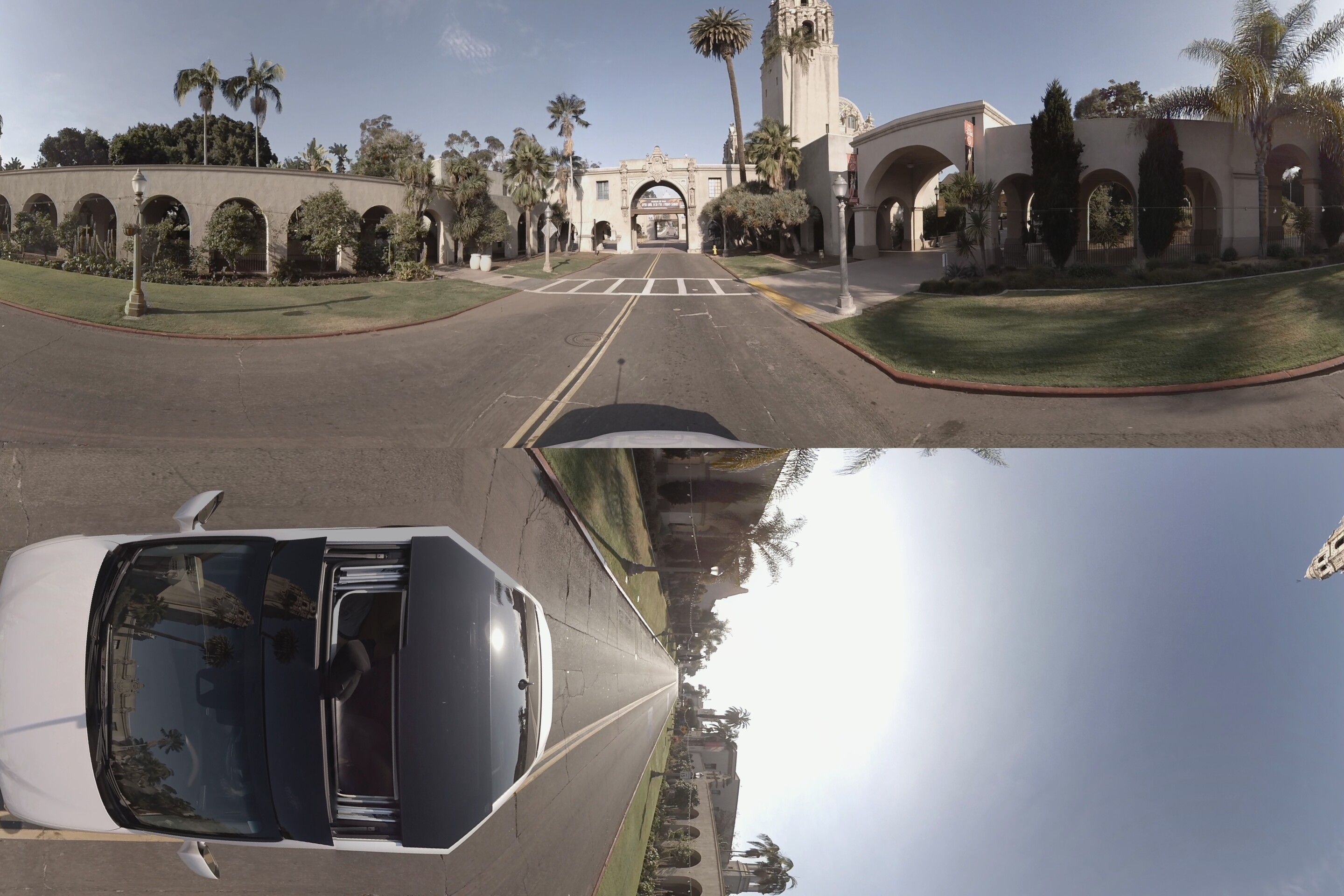}
\caption{Balboa JVET sequence in HEC format.}
\label{fig:hec_balboa}
\end{figure}

 Figure \ref{fig:hec_balboa} depicts a JVET sequence in HEC format.

\begin{figure}[h]
\centering
\includegraphics[width=150mm]{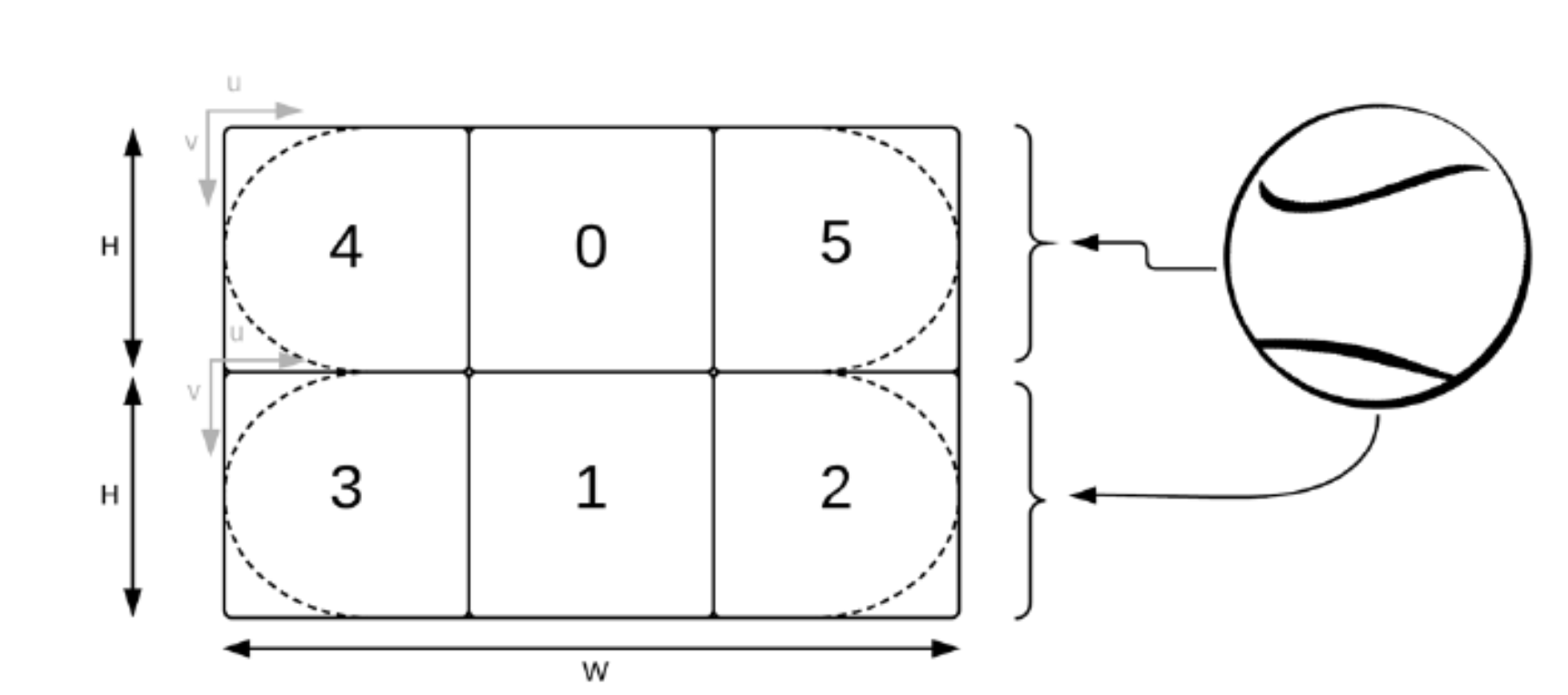}
\caption{Rotated sphere projection format from \cite{ye2017jvet}.}
\label{fig:rsp}
\end{figure}

\begin{figure}[h]
\centering
\includegraphics[width=100mm]{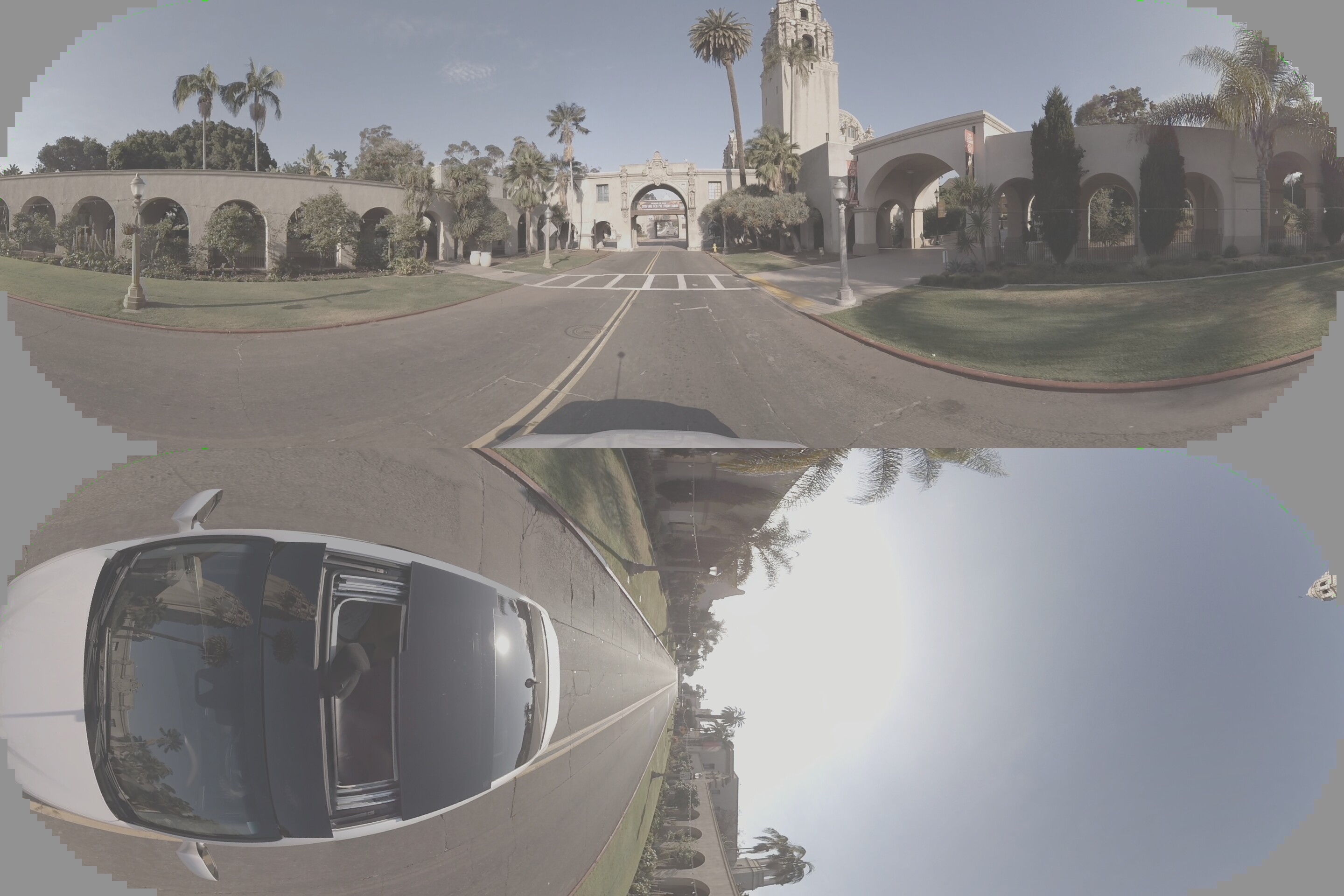}
\caption{Balboa JVET sequence in RSP format.}
\label{fig:rsp_balboa}
\end{figure}

\section{Rotated sphere projection format} 
The Rotated sphere projection (RSP) \cite{ye2017jvet} format uses two face rows to project a sphere onto a 2D plane. The first row is the center of an ERP picture spanning a 270 × 90 area. The formation of the second row involves rotating the initial ERP image and extracting the central area of the rotated ERP, corresponding to a region spanning 90 × 270. The two rows overlap at the corners with almost six percent redundant samples, which are deactivated to enhance coding efficiency. 360Lib employs six faces to implement RSP, where faces 0, 4, and 5 denote the top and bottom parts represented by faces 1, 2, and 3. Figure \ref{fig:rsp_balboa} depicts a JVET sequence in HEC format.

    \chapter{Material and methods}

\begin{figure}[h]
\centering
\includegraphics[width=150mm]{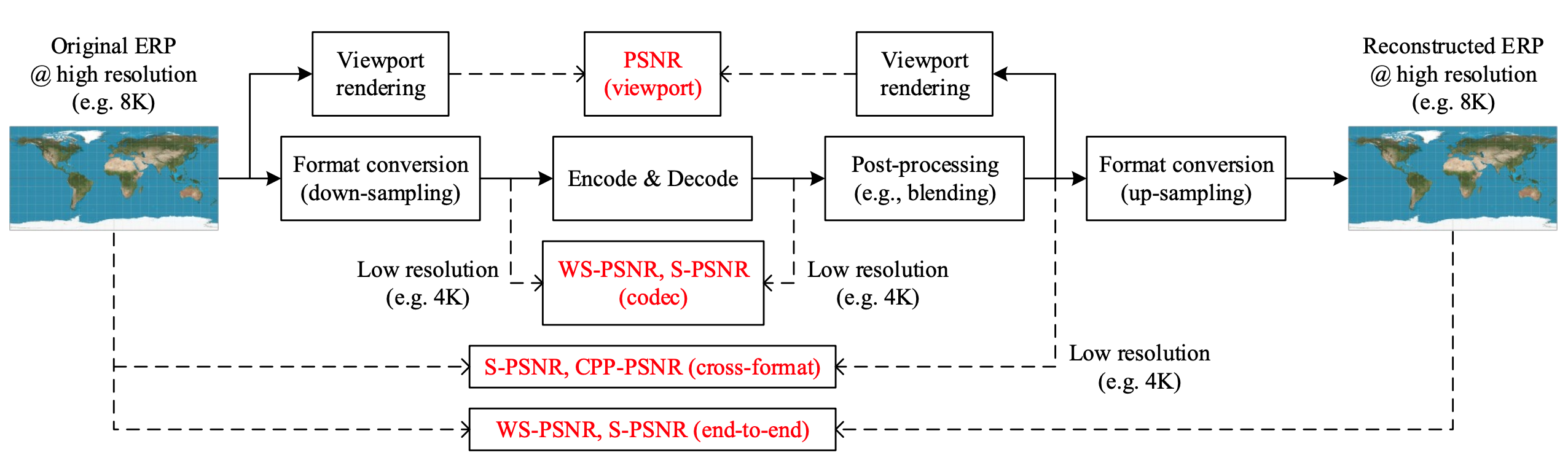}
\caption{360 video testing procedure from \cite{hanhart2018jvet}}
\label{fig:360testing}
\end{figure}

\label{chap: Material and methods}
This chapter explains the critical steps needed for preparing the pipeline for testing 360-degree videos with different projection formats based on JVET common test conditions and evaluation procedures for 360-degree video \cite{hanhart2018jvet}. Moreover, it presents methods for utilizing the scale space flow end-to-end model for compressing the projected videos.

\section{360-degree content testing procedures} \label{sec: 360-degree content testing precedur}

To assess how well 360-degree images and videos can be compressed, JVET has devised a testing protocol that evaluates their performance \cite{hanhart2018jvet}. Figure \ref{fig:360testing} depicts the JVET 360-degree common test condition procedure. The process begins by transforming the original projection format into the format used for coding. Then, this projected version is compressed using an end-to-end compression technique. It is essential to emphasize that while evaluating various methods of compressing 360-degree videos, ensuring a fair and impartial comparison is crucial. One way to do this is by making the compressed version of the video lower in quality than the original version. This ensures that the comparison is based on how well each compression method works, rather than on the quality of the original video. 

\begin{figure}[h]
\centering
\includegraphics[width=150mm]{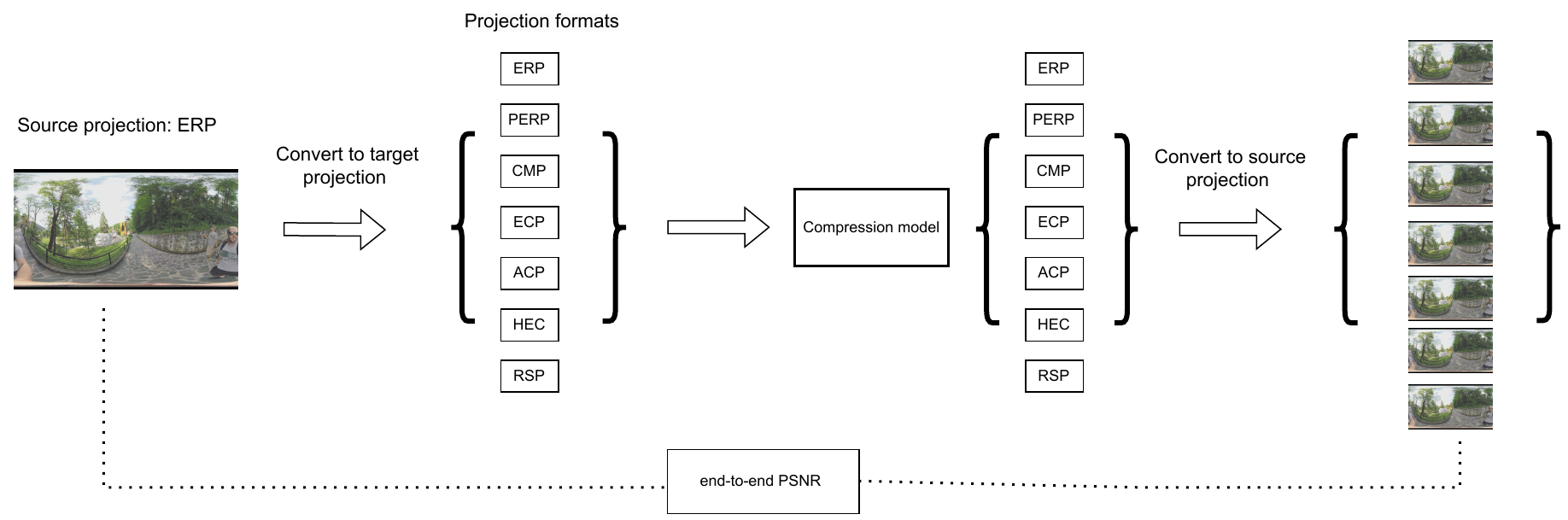}
\caption{Projection comparison pipeline}
\label{fig:Projection}
\end{figure}

After compression is done, in order to evaluate the projection efficiency, the compressed file should be converted back to the original projection with the same size as the input sequence.

\section{Implementation pipeline} \label{sec: Implementation pipeline}

This project evaluates the JVET 360-degree testing sequences listed in Table \ref{tab:sequences}. These sequences are in 8K resolution and use the YUV(4:2:0) format, with the reference projection in ERP. The primary goal is to compare the performance of coding projection formats outlined in Table \ref{tab:projection}. The aim is to identify which projections can achieve lower bitrates while maintaining acceptable video quality. To achieve this, a pipeline depicted in Figure \ref{fig:Projection} has been developed.

First, the testing sequence projection is converted to one of the projections listed in Table \ref{tab:projection}. Next, the source projected sequence is compressed using a scale-space-flow model with 8 levels of qualities (1: lowest, 8: highest). Based on the model output, the bitrate for each compression level is calculated. Then, each of these compressed sequences is converted back to the reference projection format (ERP). To evaluate the performance of the pipeline, the end-to-end PSNR between the reference sequence and the compressed sequences in the reference projection format is calculated. Several objective metrics developed for this purpose, including PSNR, WS-PSNR \cite{sun2017weighted}, and S-PSNR-I, are employed, all of which are included in the 360Lib software. Further details about these metrics will be provided later. Finally, the compression performance of the projections is analyzed using Bjontegaard Delta Bitrate (BD-Rate) \cite{herglotz2022beyond}. Overall, this project aims to provide a comprehensive evaluation of coding projection formats for 360-degree videos in the context of learning-based compression networks.

\begin{table}[!htbp]
\centering
\caption{JVET testing sequances}
\begin{tabular}{|c|c|c|c|c|c|}
\hline
Sequence name & Width & Height & Frame
count & Frame
rate & Bit
depth \\
\hline
SkateboardInLot & 8192 &4096 & 300 & 30fps  & 10 \\
ChairLift  & 8192 &4096 & 300 & 30fps  & 10 \\
KiteFlite & 8192 &4096 & 300 & 30fps  & 8\\
Harbor & 8192 &4096 & 300 & 30fps  & 8\\
GasLamp & 8192 &4096 & 300 & 30fps  & 8 \\
Balboa & 6144 &3072 & 600 & 60fps  & 8 \\
Broadway & 6144 &3072 & 600 & 60fps  & 8  \\
Landing2 & 6144 &3072 & 300 & 30fps  & 8  \\
BranCastle2 & 6144 &3072 & 300 & 30fps  & 8 \\
AerialCity & 3840 &1920 & 300 & 30fps  & 8 \\
PoleVault\_le & 3840 &1920 & 300 & 30fps  & 8 \\

\hline
\end{tabular}
\label{tab:sequences}
\end{table}

\begin{table}[htbp]
\centering
\caption{Projection formats}
\begin{tabular}{|c|c|c|c|c|}
\hline
Format & Face width & Face height & Width &
Height \\
\hline
ERP & 4096& 2048& 4096& 2048 \\
Padded ERP& 4096& 2048& 4096& 2048\\
CMP & 960& 960 &2880& 1920 \\
ACP & 960& 960& 2880& 1920 \\
EAC & 960& 960& 2880& 1920 \\
HEC &960& 960& 2880& 1920 \\
RSP& 960& 960& 2880& 1920\\
\hline
\end{tabular}
\label{tab:projection}
\end{table}

\section{Projection Conversion using 360Lib} \label{sec:Projection Conversion}
In Section \ref{sec: Implementation pipeline}, it was stated that the first step in the pipeline involves converting the reference project into a coding projection from Table \ref{tab:projection}. This conversion is accomplished using 360Lib, a C++ based computer vision software designed to evaluate the performance of 360-degree content.

In 360Lib \cite{yejoint}, the projection conversion is performed face-by-face. For a point (mc, nc) on face fc in the coding projection format and a point (mr, nr) on face fr in the reference projection format, the corresponding coordinates in the 3D XYZ space are denoted as (X, Y, Z). The conversion process initiates by associating each sample position (fc, mc, nc) on the coding projection plane with its corresponding (X, Y, Z) coordinates in the 3D system. Next, the relevant sample position (fr, mr, nr) on the reference projection plane is identified, and the sample value at (fc, mc, nc) is established based on the sample value at (fr, mr, nr).

360Lib allows users to define the relative rotation between source and destination 3D coordinates by specifying a set of 3D rotation parameters (yaw, pitch, roll) within the configuration file, thus offering flexibility. These rotation parameters allow for non-aligned XYZ axes between the reference and coding projections. In the 360Lib coordinate system, the yaw parameter indicates the degree of counterclockwise rotation around the Y axis, the pitch parameter determines the degree of counterclockwise rotation around the negative Z axis, and the roll parameter defines the degree of counterclockwise rotation along the X axis.

To perform the projection format conversion from the reference format to the coding format, the coding 2D sampling point (fc, mc, nc) is first mapped to 3D space coordinates (X, Y, Z) based on the coding projection format. If necessary, (X, Y, Z) is then rotated according to the three rotation angles yaw, pitch and roll $(\beta, \alpha, \gamma)$ to (X', Y', Z') using 

\begin{equation} \label{equ:rxyz}
R_{XYZ} = R_{Y}(\beta) \cdot R_{Z}(-\alpha)\cdot R_{x}(\gamma) 
\end{equation}
where
\begin{equation} \label{equ:rz}
 R_{Z}(\alpha)  = \begin{pmatrix}
\cos({\alpha}) & -\sin(\mathrm{\alpha}) & 0 \\
\sin(\mathrm{\alpha}) & \cos(\mathrm{\alpha}) & 0 \\
0 & 0 & 1
\end{pmatrix},
\end{equation}

\begin{equation} \label{equ:ry}
 R_{Y}(\beta)  =
\begin{pmatrix}
\cos(\mathrm{\beta}) & 0 & \sin(\mathrm{\beta}) \\
0 & 1 & 0 \\
-\sin(\mathrm{\beta}) & 0 & \cos(\mathrm{\beta})
\end{pmatrix},
\end{equation} 

and 

\begin{equation} \label{equ:rx}
 R_{ X }(\gamma)  =
\begin{pmatrix}
1 & 0 & 0 \\
0 & \cos(\mathrm{\gamma}) & -\sin(\mathrm{\gamma}) \\
0 & \sin(\mathrm{\gamma}) & \cos(\mathrm{\gamma})
\end{pmatrix}
\end{equation}

Subsequently, the coordinates (X', Y', Z') should be converted to the 2D sampling point (fr, mr, nr) in accordance with the reference projection format. In the final step, (fr, mr, nr) is illustrated by adjacent frame interpolation samples at whole number positions on face fr, while the interpolated sample value is assigned to (fc, mc, nc) in the encoding projection format.

\section{Compression model} \label{sec:Compression model}
To compress the coding projection sequence, a scale-space-flow model proposed by E. Agustsson et al. \cite{agustsson2020scale} was employed. The CompressAI API provides pre-trained weights for this scale-space model, enabling close replication of the original publication's outcomes \cite{begaint2020compressai}. The model expects input batches of RGB image tensors in the format of (N, C, H, W). Here, N denotes the batch size, and C represents the number of input channels, which is 3 for RGB images. The spatial dimensions of the images are denoted by (H, W). Input image data must fall within the [0, 1] range, and there is no need for normalization beforehand. The dimensions H and W should be at least 64 pixels in length, and the input frame's height and width should be a power of 2 \cite{begaint2020compressai}. The model's loss is determined by the total rate-distortion loss, which is optimized across $N$ frames. The loss equation is defined as

\begin{equation}
L = \sum_{i=0}^{N-1} d(x_i, \hat{x}_i) + \lambda \left[H(z_0) + \sum_{i=1}^{N-1} H(v_i) + H(w_i)\right],
\label{equ:loss}
\end{equation}
 where $L$ represents the total rate-distortion loss, $d(x_i, \hat{x_i})$ is the distortion metric (e.g., mean squared error (MSE) or multiscale structural similarity (MS-SSIM) \cite{wang2003multiscale}) between the actual frame $x_i$ and its reconstructed version $\hat{x}_i$, the Lagrange multiplier, represented by $\lambda$, governs the balance between distortion and bitrate in the compression process, the notation $H(\cdot)$ represents the estimated entropy of the associated latent variable, factoring in side information from the hyperprior, $z_0$ is the image latent, $v_i$ refers to the residual latent, and $w_i$ stands for the motion compensation latent.

The rate points are influenced by the $\lambda$ parameter, which balances the trade-off between distortion and bitrate. A higher $\lambda$ value prioritizes lower bitrate at the cost of increased distortion, while a lower $\lambda$ value emphasizes lower distortion but results in a higher bitrate. The model is fine-tuned for both MSE and MS-SSIM distortion metrics at 9 distinct levels, covering bitrates between 0.025 and 0.8 bpp. For MSE, $\lambda$ is configured as $0.01 \cdot 2^i$, and for MS-SSIM, $\lambda$ is set to $10 \cdot 2^i$, with $i$ ranging from -3 to 5. Through this procedure, two separate models are created, with one being optimized for MSE and the other tailored for MS-SSIM. The optimization procedure employs the Adam optimizer \cite{kingma2014adam}, using a batch size of 8 and a learning rate of $10^{-4}$. This approach facilitates effective training and convergence of the model parameters.

\section{Evaluation metrics} \label{sec:Evaluation metrics}

Considering the spherical nature of 360-degree video, traditional evaluation metrics must be adapted to accurately assess the differences between the original and the compressed frame in the reference projection format. 360Lib offers three objective spherical quality metrics for evaluating the quality of 360-degree videos, which are discussed in detail below.

\subsection{Peak-Signal-to-Noise ratio}
Peak-signal-to-noise ratio (PSNR) \cite{huynh2008scope} refers to a metric used to evaluate the level or degree of excellence in a signal by evaluating the power of the noise that interferes with the signal against the maximum power of the signal itself. In other words, it represents the ratio of the highest possible signal power to the power of the noise that affects the signal. PSNR serves as a frequently employed measure to evaluate codec performance or to conduct comparisons between various video codecs. PSNR is calculated using \begin{equation}
PSNR = 10 \log_{10} \left(\frac{255^2}{MSE}\right),
\label{equ:psnr}
\end{equation}
where  \begin{equation}
MSE = \frac{\sum_{i=1}^{M} \sum_{j=1}^{N}(x(i,j)-\hat{x}(i,j))^2}{N\cdot M}.  
\label{equ:psnr2}
\end{equation}
$MSE$ represents the mean-squared error of the pixel luminance values between the matching frames in the compressed $\hat{x}(i,j)$ and reference videos $x(i,j)$ where $(i, j)$ is the coordinate of a pixel. Please denote that $(M, N)$ is the frame size \cite{yejoint}.

\subsection{Weighted-to-Spherically-Uniform PSNR}

Another assessment metric for 360-degree videos is the Weighted-to-Spherically-Uniform Peak-Pignal-to-Noise ratio (WS-PSNR)\cite{sun2017weighted} 

\begin{equation}
\text{WS-PSNR} = 10 \log_{10} \left(\frac{{255}^2}{\text{WS-MSE}}\right),
\label{equ:wspsnr1}
\end{equation}

where

\begin{equation}
\text{WS-MSE} = \frac{1}{\sum\limits_{i=0}^{M-1}\sum\limits_{j=0}^{N-1} s(i,j)}\sum\limits_{i=0}^{M-1}\sum\limits_{j=0}^{N-1} {(x(i,j) - \hat{x}(i,j))^2} \cdot s(i,j).
\label{equ:wspsnr2}
\end{equation}
The spherical factor, denoted as $s(i, j)$, is an essential element for analyzing geometric distortions associated with projections along an axis in a spherical system. It signifies the impact of the sphere's curvature on the distortion encountered at each position (i, j) within the given projection. However, this factor is not consistent across different projections \cite{yejoint}.
For ERP, it is illustrated as

\begin{equation}
s(i, j)_{\text{ERP}} = \cos\left(\frac{(j + 0.5 - \frac{N}{2})\pi}{N}\right).
\end{equation}

The spherical factor $s(i, j)$ for CMP-based projection is defined as follows

\begin{equation}
s(i, j)_{\text{CMP}} = \left(1 + \frac{d^2(i, j)}{r^2}\right)^{-\frac{3}{2}}.
\end{equation}

In this equation, $d(i, j)$ represents the distance of each point from the center of the cube map face, which is illustrated as

\begin{equation}
d^2(i, j) = (i + 0.5 - \frac{A}{2})^2 + (j + 0.5 - \frac{A}{2})^2,
\end{equation}

where each face of the cube map projection has a resolution of $(A\times A)$ and a radius $r = \frac{A}{2}$.

The ACP format features uniform weight distribution across all its faces, so only the weights on one face are calculated. As ACP is an extension of CMP its weights are calculated based on CMP as 

\begin{equation}
s(i, j)_{\text{ACP}}  = \frac{1}{\left(1 + 4 \cdot \left[\left(\frac{i_{CMP} + \frac{1}{2} - \frac{A}{2}}{A}\right)^{2} + \left(\frac{j_{CMP} + \frac{1}{2} - \frac{A}{2}}{A}\right)^{2}\right]\right)^{\frac{3}{2}} \cdot 16 \cdot t_{i} \cdot t_{j}}
,
\end{equation}
where

\begin{equation}
t_i = \sqrt{0.342 - 0.09 \cdot \left|2\left(\frac{i + 0.5}{A} \right) - 1\right|}
,
\end{equation}
and 

\begin{equation}
t_j = \sqrt{0.342 - 0.09 \cdot \left|2\left(\frac{j + 0.5}{A} \right) - 1\right|}.
\end{equation}
In order to map each position $(i,j)$ from ACP face to CMP, these conversions should be applied as
\begin{equation}
i_{CMP} = 0.5 \cdot A \cdot \left(1.0 + \frac{0.34 - \sqrt{0.342 - 0.09 \cdot \left|2\left(\frac{i + 0.5}{A}\right) - 1\right|}}{0.18}\right)
, 
\end{equation}
and 

\begin{equation}
j_{CMP} = 0.5 \cdot A \cdot \left(1.0 + \frac{0.34 - \sqrt{0.342 - 0.09 \cdot \left|2\left(\frac{j + 0.5}{A}\right) - 1\right|}}{0.18}\right)
.
\end{equation}

For EAC similar formulation as ACP and CMP applies with minor modification due to its different forward mapping with the spherical factor 

\begin{equation}
s(i, j)_{\text{EAC}}  = \frac{\pi ^2}{\left(1 + (\tan(t_i))^2 + (\tan(t_j))^2\right)^\frac{3}{2} \cdot 16 \cdot (\cos(t_i))^2 \cdot (\cos(t_j))^2},
\end{equation}
where

\begin{equation}
t_i = \frac{\pi}{4} \cdot \left(\frac{2(i + 0.5)}{A} - 1\right),
\end{equation}
and
\begin{equation}
t_j = \frac{\pi}{4} \cdot \left(\frac{2(j + 0.5)}{A} - 1\right).
\end{equation}

As stated in the previous chapter, HEC has a similar forward mapping for top and bottom faces as EAC, which means that for these faces, the spherical factor is the same as EAC. However, for the back, right, and front faces, the spherical factor is illustrated as

\begin{equation}
s(i, j)_{\text{HEC}} = \frac{1}{\left(1 + i_{HEC}^2 + j_{HEC}^2\right)^{\frac{3}{2}}} \cdot \frac{\pi\left(1 + 0.4(1 - u^2)(1 + v^2)\right)}{4\cos^2\left(\frac{\pi}{4}u\right)(1 + 0.4(1 - u^2)(1 - v^2))^2}
\end{equation}
where 

\begin{equation}
u = 2\left(\frac{i + 0.5}{A}\right) - 1,
\end{equation}
and

\begin{equation}
v = 2\left(\frac{j + 0.5}{A}\right) - 1.
\end{equation}

Give $(i,j)$ as the pixel position on each face

\begin{equation}
i_{HEC} = \tan\left(\frac{\pi}{4}u\right),
\end{equation}
and 
\begin{equation}
j_{HEC} = \frac{v}{1 + 0.4(1 - u^2)(1 - v^2)}.
\end{equation}

RSP has a different structure compared to the previous projections mentioned thus far. The RSP consists of two circular faces of equal size. As illustrated in Figure \ref{fig:rsp}, when the circles are collapsed, some pixel positions are not selected, referred to as inactive sample positions. Consequently, the spherical factor is only calculated for the active samples within the circles as

\begin{equation}
s(i, j)_{RSP} = \begin{cases}
0, & (i, j) \in \text{inactive sample} \\
\cos(\theta), & (i, j) \in \text{active sample}
\end{cases}.
\end{equation}

\subsection{Spherical PSNR}
Spherical PSNR (S-PSNR) is a method for evaluating the quality of 360-degree videos by employing uniformly distributed points on a sphere. To compute S-PSNR, the first step involves choosing a point $s$ on the sphere. \\

\begin{figure}[t]
\centering
\includegraphics[width=100mm]{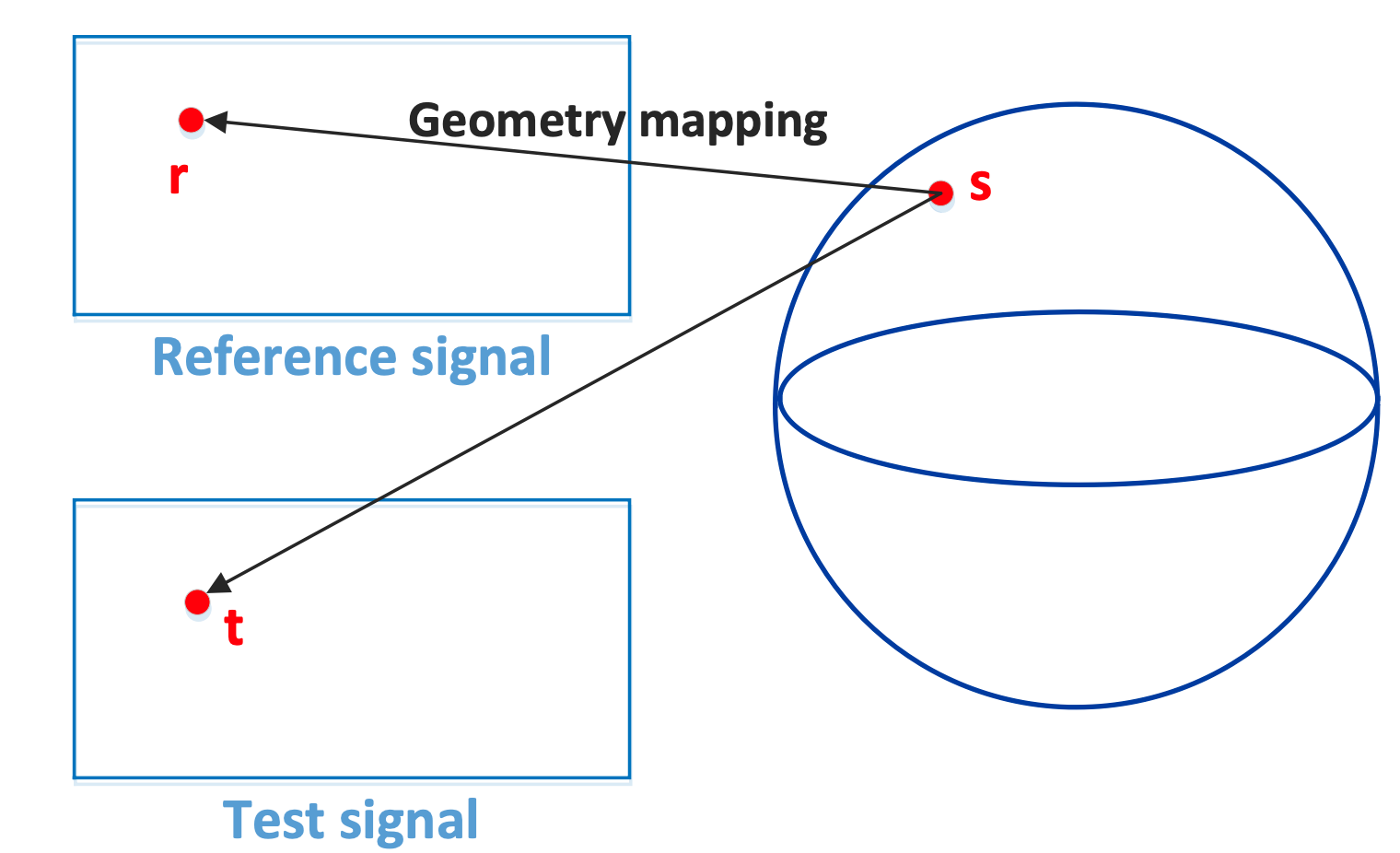}
\caption{Sampled points on sphere from \cite{yejoint}.}
\label{fig:spsnr}
\end{figure}

Subsequently, the associated positions $r$ in the reference signal and $t$ in the compressed frame are determined using a 3D-to-2D coordinate mapping technique as depicted in Figure \ref{fig:spsnr}. The distortion is subsequently determined based on the sample values at positions $r$ and $t$. This procedure is executed for each point $s$ in the set, with the total distortion being accumulated. Lastly, PSNR is derived from the accumulated distortion \cite{yejoint}.

\begin{figure}[t]
\centering
\includegraphics[width=150mm]{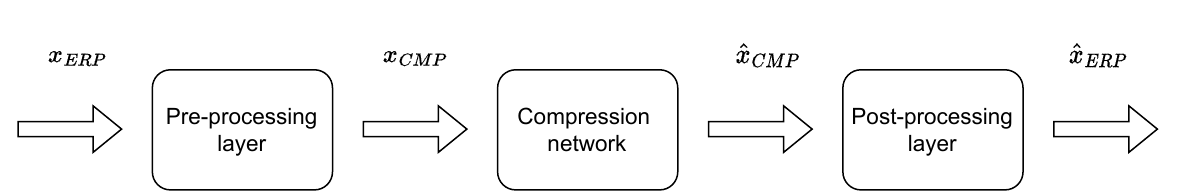}
\caption{Projection model forward pass.}
\label{fig:forward}
\end{figure}

\begin{figure}[h]
\centering
\includegraphics[width=100mm]{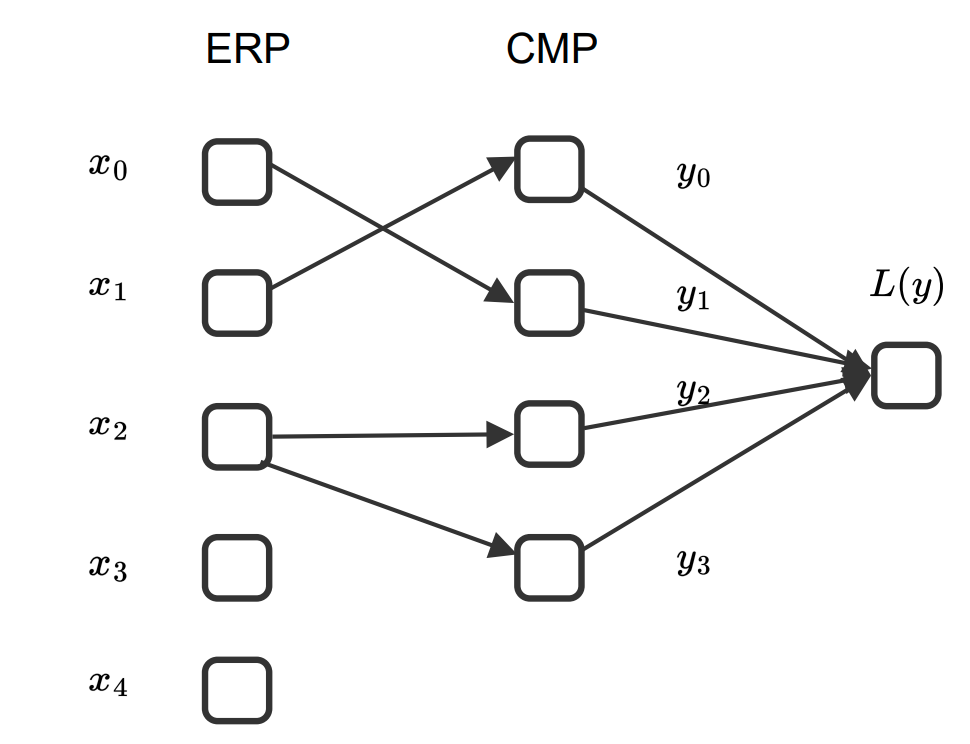}
\caption{CMP layer forward pass}
\label{fig:cmpforward}
\end{figure}

\section{Projection conversion using neural networks}
In addition to the method discussed in Section \ref{sec:Projection Conversion} for converting projections using 360Lib software, a reprojection layer with fixed parameters has been implemented in Python. This approach employs a model consisting of three layers. The first layer transforms the reference projection into the coding projection, while the middle layer is the compression model, compressing the coding projection. Finally, the third layer reverts the coding projection back to the reference projection. It is crucial to emphasize that the projection conversion layers do not contain any learning parameters; instead, they function as pre- and post-processing stages for the input and output of the compression layer.

As illustrated in Figure \ref{fig:forward}, the forward pass of the model begins with the conversion of ERP frames to CMP frames using the first layer, which applies pre-processing to the original ERP frames. Following this step, the CMP frames are compressed using the compression layer. Subsequently, a post-processing layer is employed to convert the CMP frames back to ERP frames. At this stage, the evaluation metrics are calculated in the same manner as before, and the loss between the compressed and uncompressed ERP frames is determined. The post-processing and preprocessing layers function similarly, with the only difference being the reversal of projection types. In this section, the pre-processing layer is explained, which involves converting a set of ERP frames into CMP frames.

Figure \ref{fig:cmpforward} demonstrates the functioning of the pre-processing layer, referred to as the CMP layer. In a nutshell, the process begins by generating a set of points with the same size as the ERP frames, \begin{equation}
 x= \begin{pmatrix}
x_{1} & x_{2} & x_{3} \\
x_{4}& x_{5} & x_{6} \\
\end{pmatrix}
\label{equ:erp}
\end{equation} representing the ERP projection. The same procedure is carried out for the CMP as well \begin{equation}
 y =\begin{pmatrix}
y_{1} & y_{2}  \\
y_{3}& y_{4} \\
\end{pmatrix}.
\label{equ:cmp}
\end{equation} 
These points indicate a set of positions in a 2D plane. To generate these points, a Python function is employed, which creates points based on the respective projections. Subsequently, another function is used to map the CMP points to the ERP points. As shown in Figure \ref{fig:cmpforward}, this mapping is not one-to-one, and some ERP points may not be mapped at all. The mapping coordinates from ERP to CMP, generated by the Python function, may not be integer values. Therefore, they must be rounded

\begin{equation}
f(x_i)=\text{round} (y_i), 
\label{equ:map2}
\end{equation} as these points represent pixel locations. Once rounded, the positions are extracted from the ERP frame and represented in the CMP form. The mapping process can be formulated as 
\begin{equation}
y_j = \sum_{i=1}^n w_i x_i,
\label{equ:mapfor}
\end{equation} where 
\begin{equation}
{w_i}=\begin{cases}
			1 & \text{if } x_i = \text{round}(y_j) \\
                0 & \text{otherwise}
		 \end{cases}
\label{equ:w}
\end{equation} represents the weights of each neuron. It is important to note that in these equations, the indices $(i = 1,.., N)$ and $(j = 1,..., M)$ correspond to the number of ERP points $N$ and CMP points $M$, respectively. The rounding process mentioned earlier, which maps non-integer coordinates to integer values, essentially involves selecting the nearest neighbors of each point. In other words, when multiple points are located close to each other, the nearest neighbor interpolation identifies the point that is closest to all the other points in the group. This interpolation method ensures that each non-integer coordinate is assigned to the closest integer coordinate, representing the pixel location in the image. By doing so, the mapping process can efficiently convert points between the ERP and CMP projections while preserving the spatial relationships between the points in the image. 

\begin{figure}[t]
\centering
\includegraphics[width=150mm]{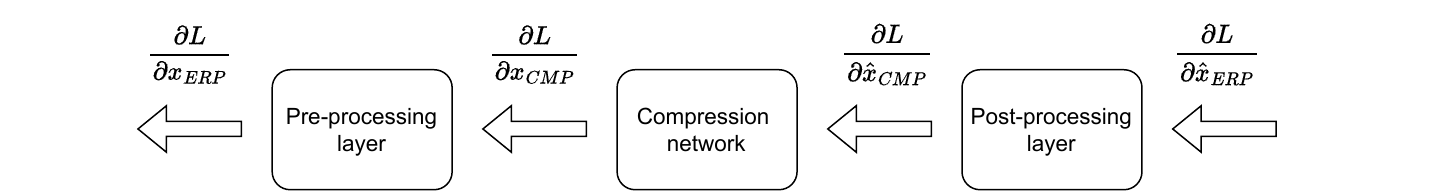}
\caption{Projection model backward pass}
\label{fig:modelbackward}
\end{figure}

Figure \ref{fig:modelbackward} depicts the backward pass of the model through all three layers. During the backpropagation process, the weights of the model are updated to minimize the loss function. Here is a high-level overview of the backpropagation process for this model, treating each layer as a black box. The post-processing layer receives the gradient of loss with respect to the post-processing layer's output, which is the compressed ERP frames $\frac{\partial L}{\partial \hat x_{ERP}}$. It then calculates the gradient of the loss with respect to its input, which is the compressed CMP frames $\frac{\partial L}{\partial \hat x_{CMP}}$. This process continues for all layers based on the chain rule, considering the weights and activation functions of each layer.
\clearpage

\begin{figure}[t]
\centering
\includegraphics[width=100mm]{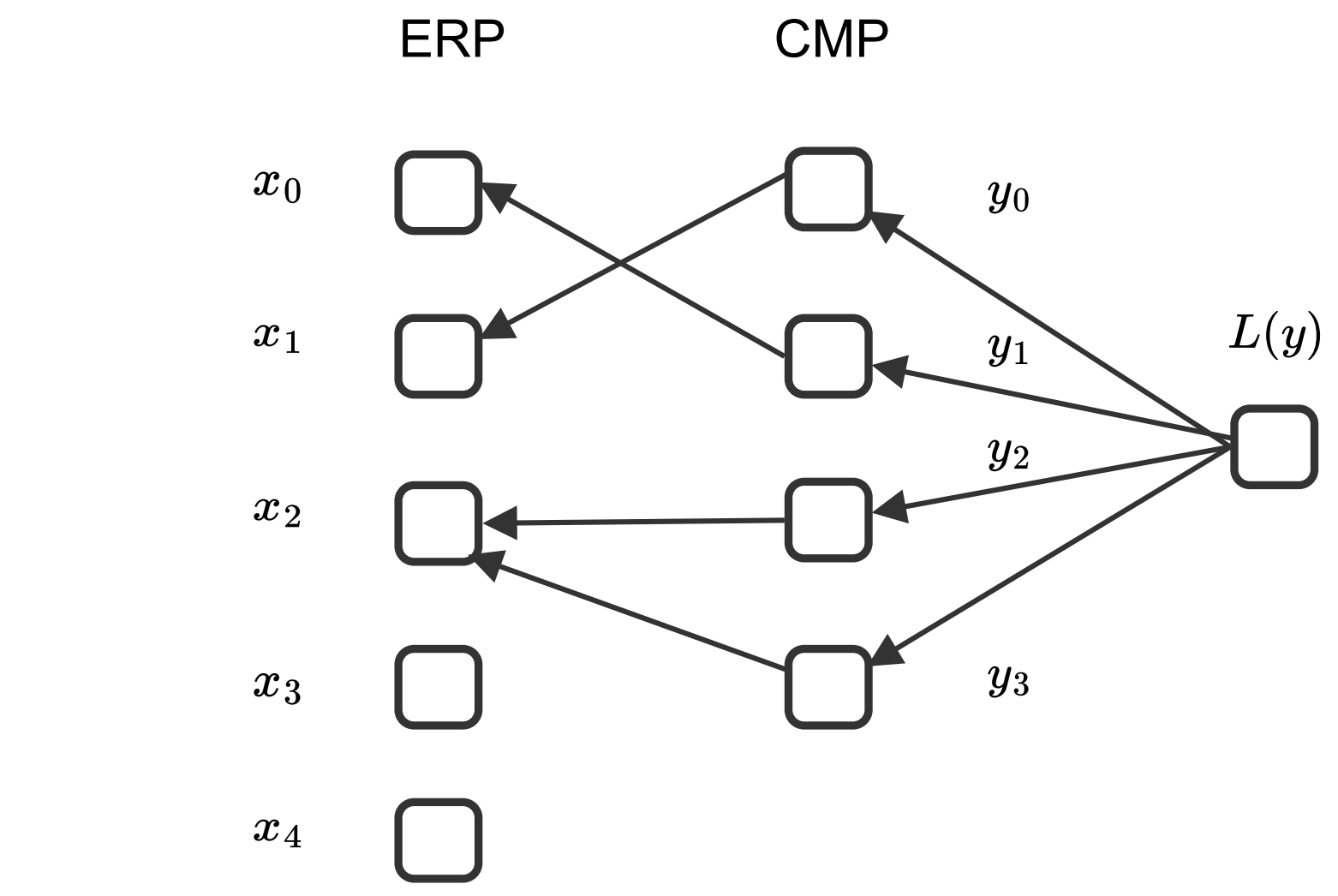}
\caption{CMP layer backward pass}
\label{fig:backwardcmp}
\end{figure}

The goal is to obtain the gradient of loss with respect to the model input $\frac{\partial L}{\partial x_{ERP}}$, which corresponds to the uncompressed ERP frames. Finally, the model's weights are updated based on these gradients to minimize the loss function. Figure \ref{fig:backwardcmp} demonstrates the backward pass of the CMP layer. The forward pass mapping function is not one-to-one, so the backward pass must model the relationship between each CMP point and its corresponding ERP points. To achieve this, the gradient of the loss with respect to the input must be computed, denoted as $\frac{\partial L}{\partial x}$. The chain rule is applied to achieve this calculation. The gradient of the loss with respect to the input can be determined by 

\begin{equation}
\frac{\partial L}{\partial x_0} = \frac{\partial L}{\partial y_1} \cdot \frac{\partial y_1}{\partial x_0}= \frac{\partial L}{\partial y_1}, 
\label{equ:gl0}
\end{equation}

\begin{equation}
\frac{\partial L}{\partial x_1} = \frac{\partial L}{\partial y_0} \cdot \frac{\partial y_0}{\partial x_1} = \frac{\partial L}{\partial y_0}, 
\label{equ:gl1}
\end{equation}

\begin{equation}
\frac{\partial L}{\partial x_2} = \frac{\partial L}{\partial y_2} \cdot \frac{\partial y_2}{\partial x_2} + \frac{\partial L}{\partial y_3} \cdot \frac{\partial y_3}{\partial x_2} =\frac{\partial L}{\partial y_2}+\frac{\partial L}{\partial y_3}, 
\label{equ:gl2}
\end{equation}
and 
\begin{equation}
\frac{\partial L}{\partial x_3} = \frac{\partial L}{\partial y_3} \cdot \frac{\partial y_3}{\partial x_3} = 0, 
\label{equ:gl3}
\end{equation} 

as shown in Figure \ref{fig:backwardcmp}. The gradient of the output with respect to the input, $\frac{\partial y_i}{\partial x_i}$, is either zero or one. If a point from the ERP projection $x_i$ is mapped to a point in the CMP projection $y_i$, the gradient is one; otherwise, it is zero. Using (\ref{equ:mapfor}), the gradient of the output with respect to the input can be found as $w_i = \frac{\partial L}{\partial x_i}$ for each ERP point $x_i$. Each point is associated with a weight vector $w_i$ of size $N\times1$, where its elements are either zero or one. The gradient of the loss with respect to the output is a matrix

\begin{equation}
\frac{\partial L}{\partial y} =\begin{pmatrix}
\frac{\partial L}{\partial y_1} & \frac{\partial L}{\partial y_2}  \\
\frac{\partial L}{\partial y_3} & \frac{\partial L}{\partial y_4}  \\
\end{pmatrix}.
\label{equ:gly1}
\end{equation}

By employing (\ref{equ:gly1}), the gradient of the loss with respect to the input can be determined as
\begin{equation}
\frac{\partial L}{\partial x} =\begin{pmatrix}
w_1 \cdot \frac{\partial L}{\partial y} & w_2 \cdot \frac{\partial L}{\partial y}  ..... & w_N \cdot \frac{\partial L}{\partial y}  \\
\end{pmatrix}.
\label{equ:glx}
\end{equation}

    \chapter{Evaluation and Discussion}
\label{chap:Evaluation and Discussions}

\begin{figure}[h]
\centering
\includegraphics[width=100mm]{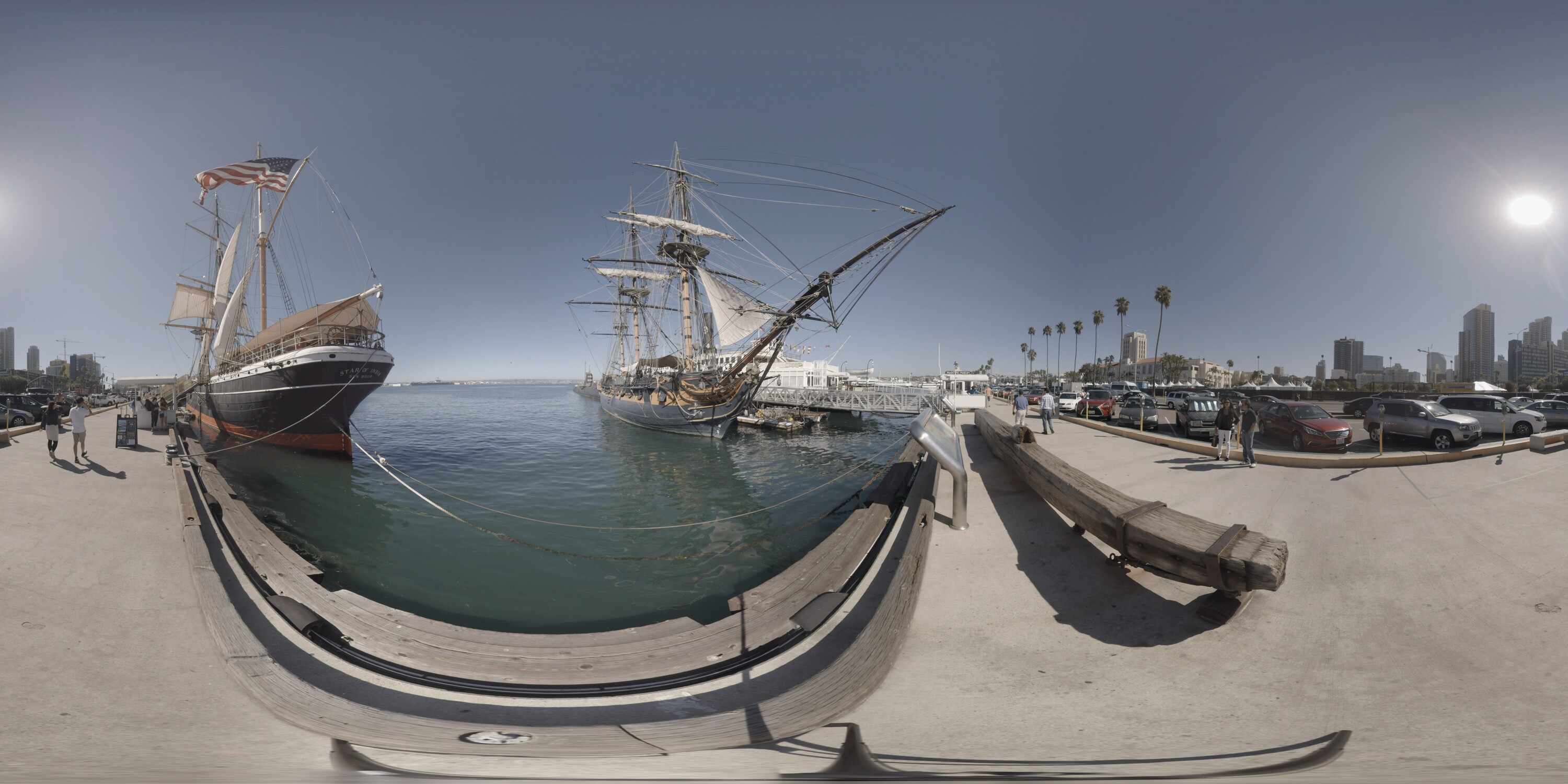}
\caption{Harbor reference projection ERP}
\label{fig:Harbor_ERP}
\end{figure}

In this chapter, the results of various stages of the proposed pipeline will be discussed. The process starts with converting the original projection ERP to each of the coding projections. This is followed by compressing the converted projection using eight different compression levels. The bpp (bits per pixel) is calculated for each of these levels. The projection is then converted back to the reference projection (ERP), and the end-to-end PSNR is calculated by comparing the reference ERP frames to the compressed ERP frames. 

\section{Evaluation of the 360lib pipeline}

\subsection{Projection conversion}

For each sequence, the reference projection undergoes conversion to each of the seven projections listed in Table \ref{tab:projection}. Figure \ref{fig:Harbor_ERP} depicts the Harbor sequence in reference projection (ERP). To gain a better understanding of the effects of each projection, each projection and their compressed versions are analyzed at both the highest and lowest compression levels. The projections used can be classified into three categories: ERP-based, cube map-based, and RSP-based projections. 

ERP and Padded ERP projections are depicted in Figure \ref{fig:Harbor_ERP} (a),(d) respectively. Artifacts can be seen in subfigures (b) and (e) for ERP and PERP respectively at the lowest compression level. These artifacts are caused by the compression process, which can lead to loss of image quality and the introduction of visible distortions in the output. To gain a deeper understanding of the impact of compression on image quality, one can compare the original ERP and PERP projections (subfigures (a) and (d)) with their compressed versions (subfigures (b) and (e)). This comparison can reveal the specific types of artifacts introduced and assess the overall performance of the compression algorithm. In this sequence, blockiness within the water and around the boat is easily noticeable. At the highest compression level, the model's performance can be observed, and the difference is not significant. This holds true for other projection formats as well.

\begin{figure}
    \centering
    \begin{subfigure}[b]{0.32\textwidth}
        \includegraphics[width=\textwidth]{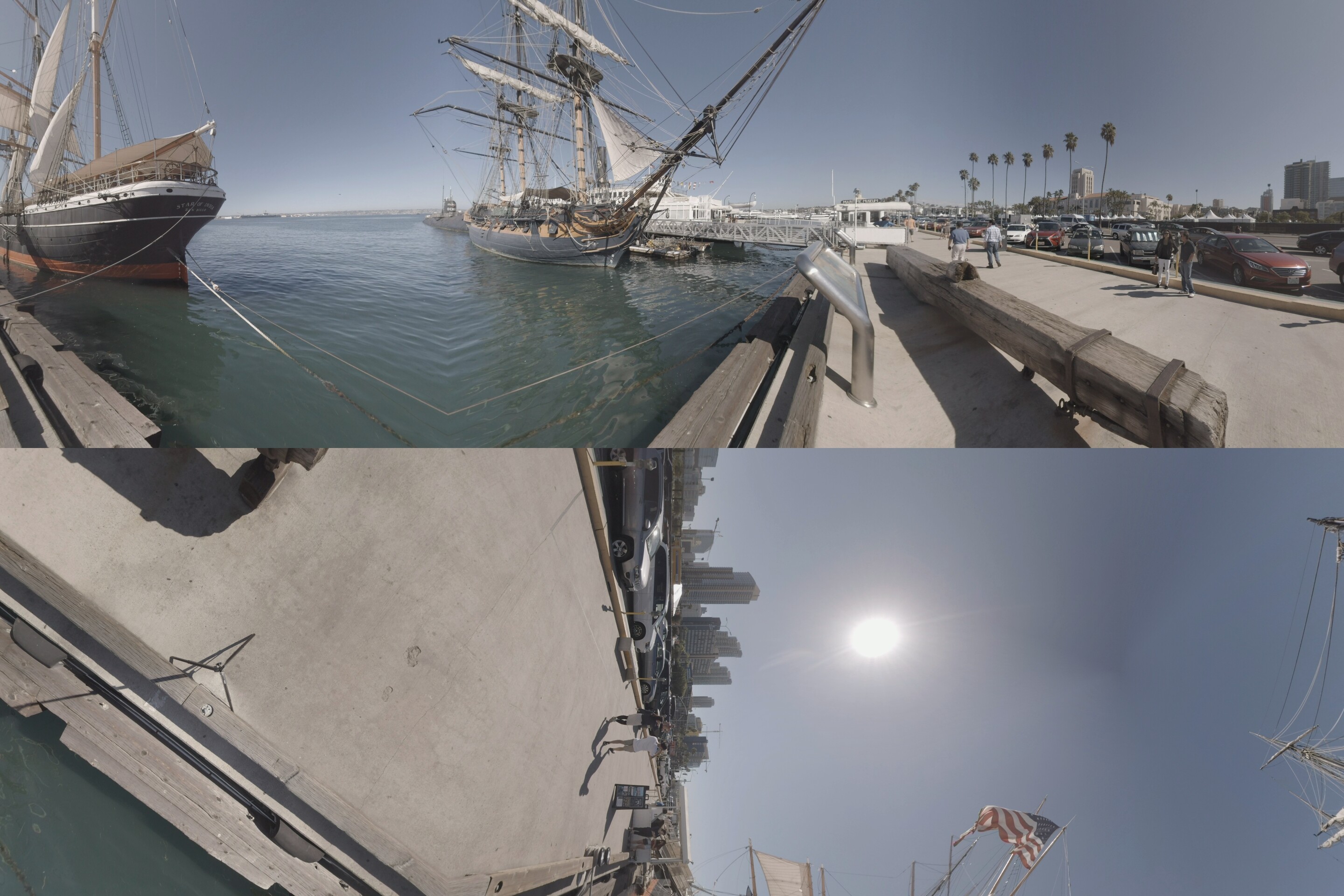}
        \caption{CMP before applying compression}
    \end{subfigure}
    \begin{subfigure}[b]{0.32\textwidth}
        \includegraphics[width=\textwidth]{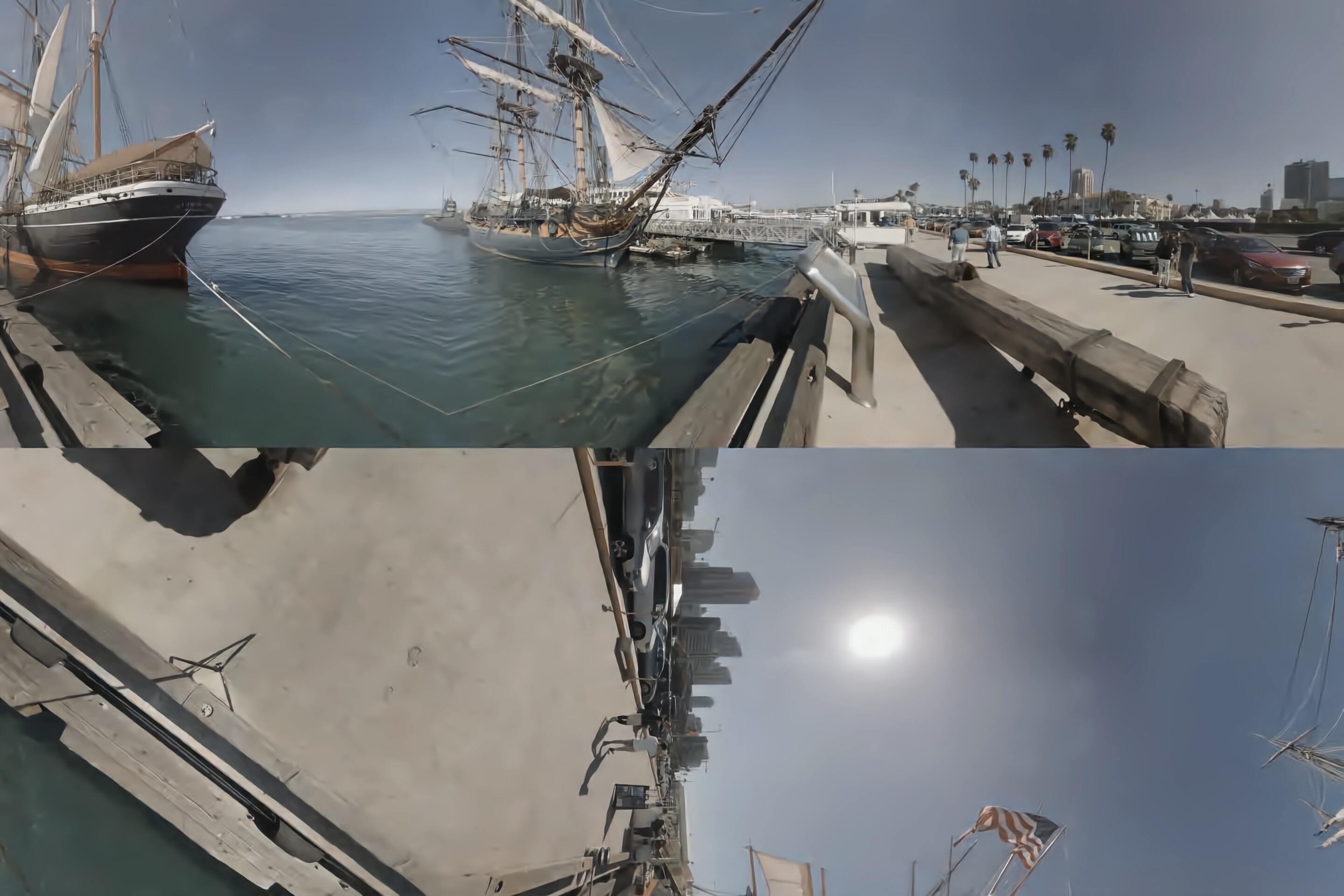}
        \caption{CMP lowest 
        compression level}
    \end{subfigure}
    \begin{subfigure}[b]{0.32\textwidth}
        \includegraphics[width=\textwidth]{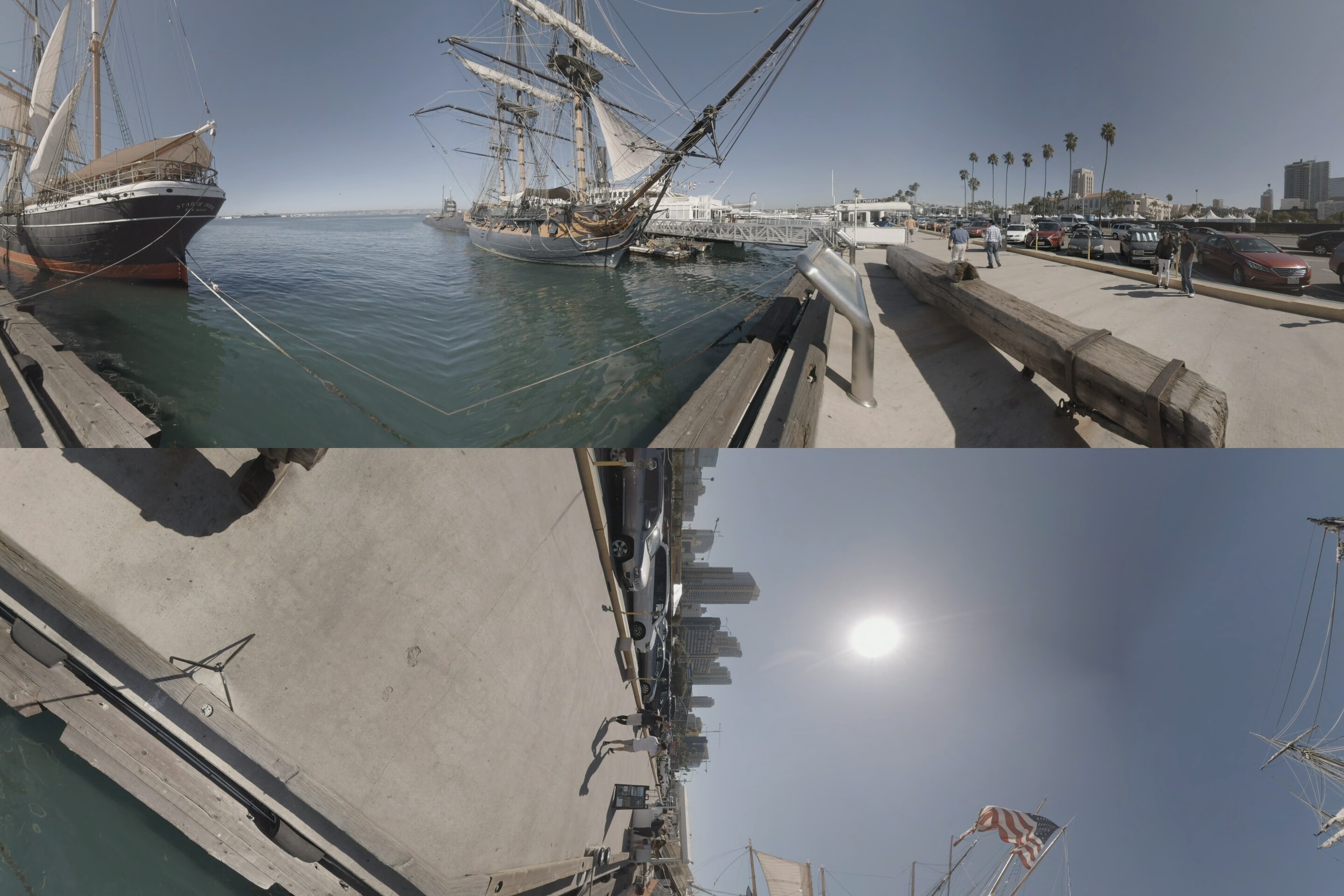}
        \caption{CMP highest
        compression level}
    \end{subfigure}

        \begin{subfigure}[b]{0.32\textwidth}
        \includegraphics[width=\textwidth]{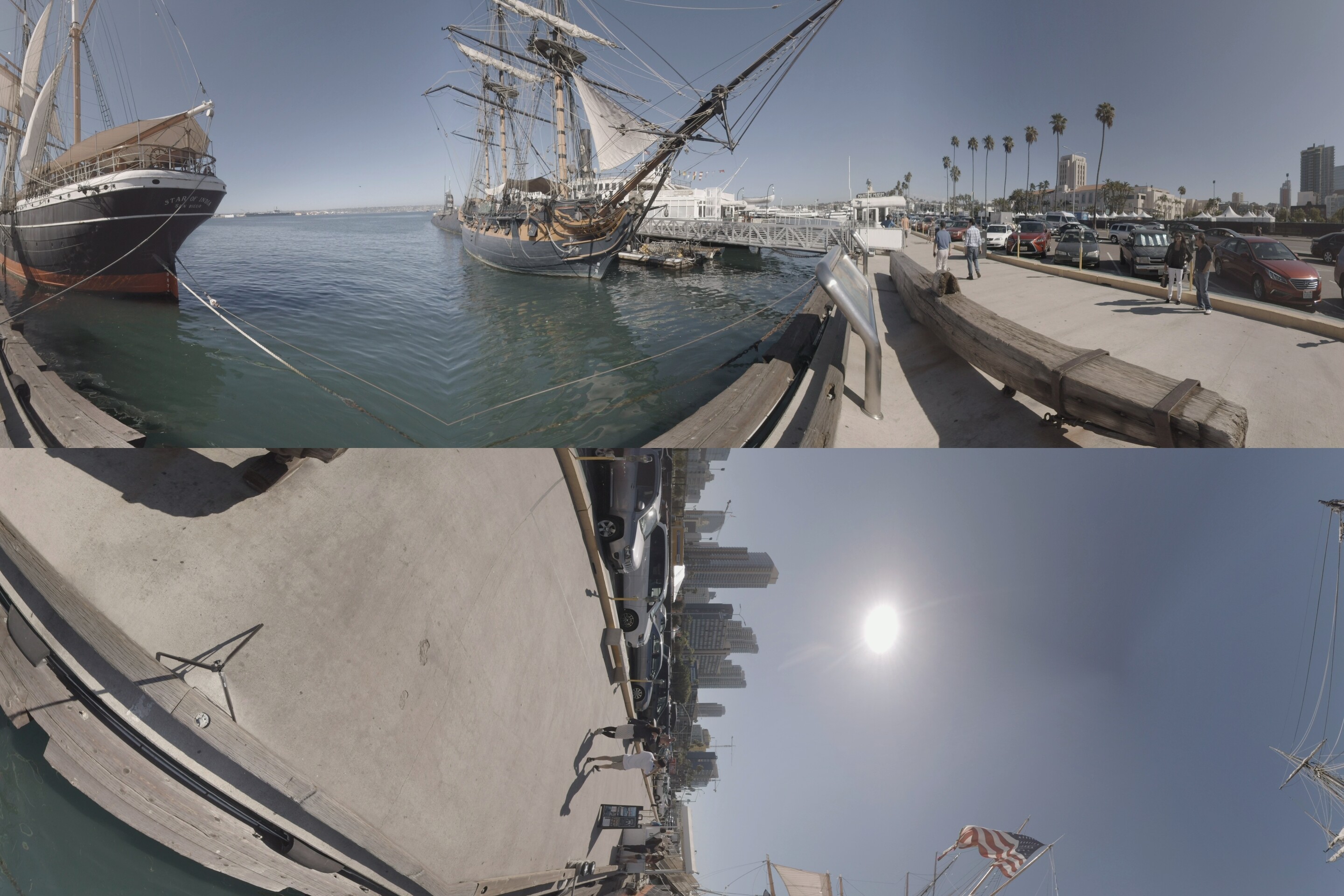}
        \caption{EAC before applying compression}
    \end{subfigure}
    \begin{subfigure}[b]{0.32\textwidth}
        \includegraphics[width=\textwidth]{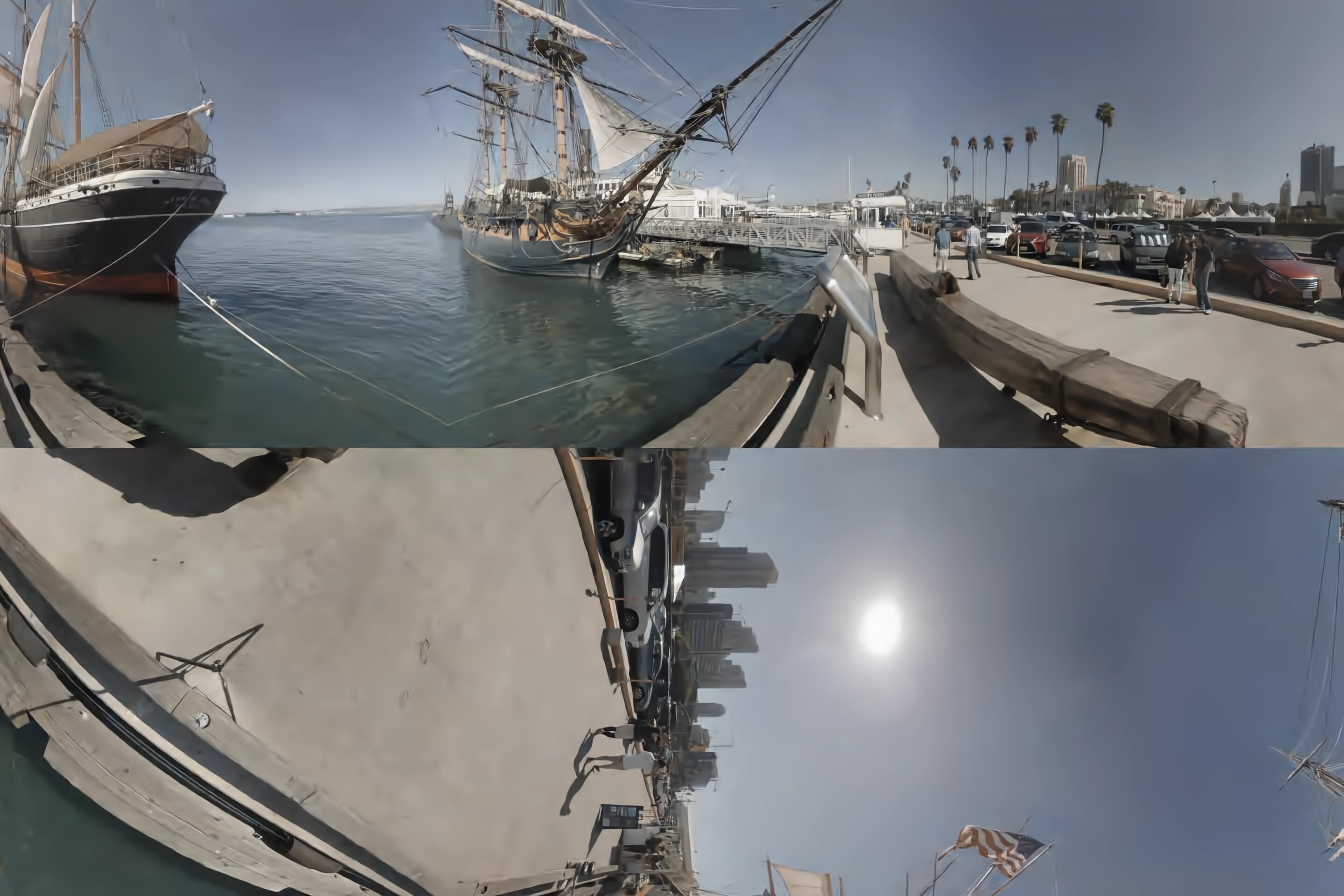}
        \caption{EAC lowest 
        compression level}
    \end{subfigure}
    \begin{subfigure}[b]{0.32\textwidth}
        \includegraphics[width=\textwidth]{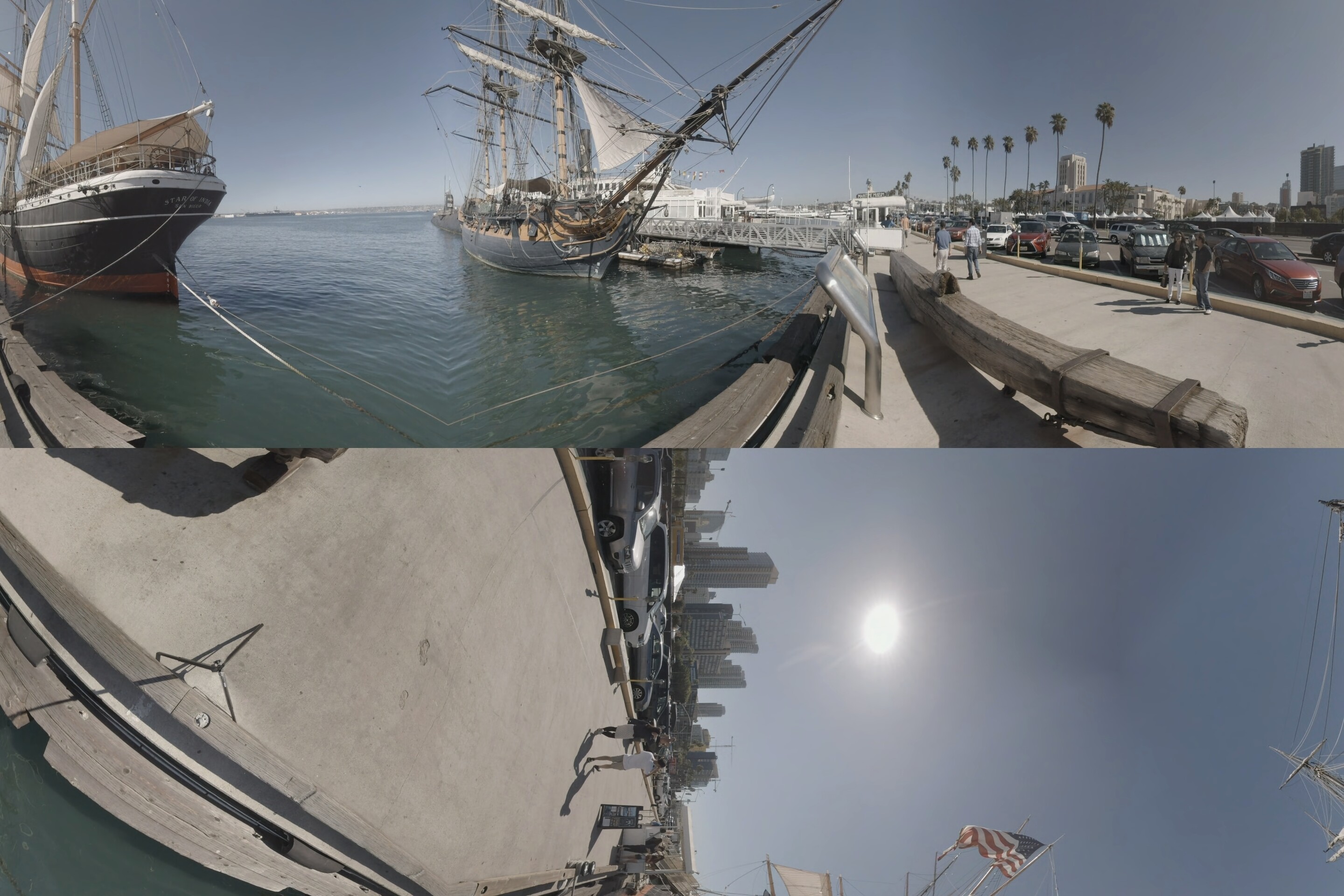}
        \caption{EAC highest
        compression level}
    \end{subfigure}

    \begin{subfigure}[b]{0.32\textwidth}
        \includegraphics[width=\textwidth]{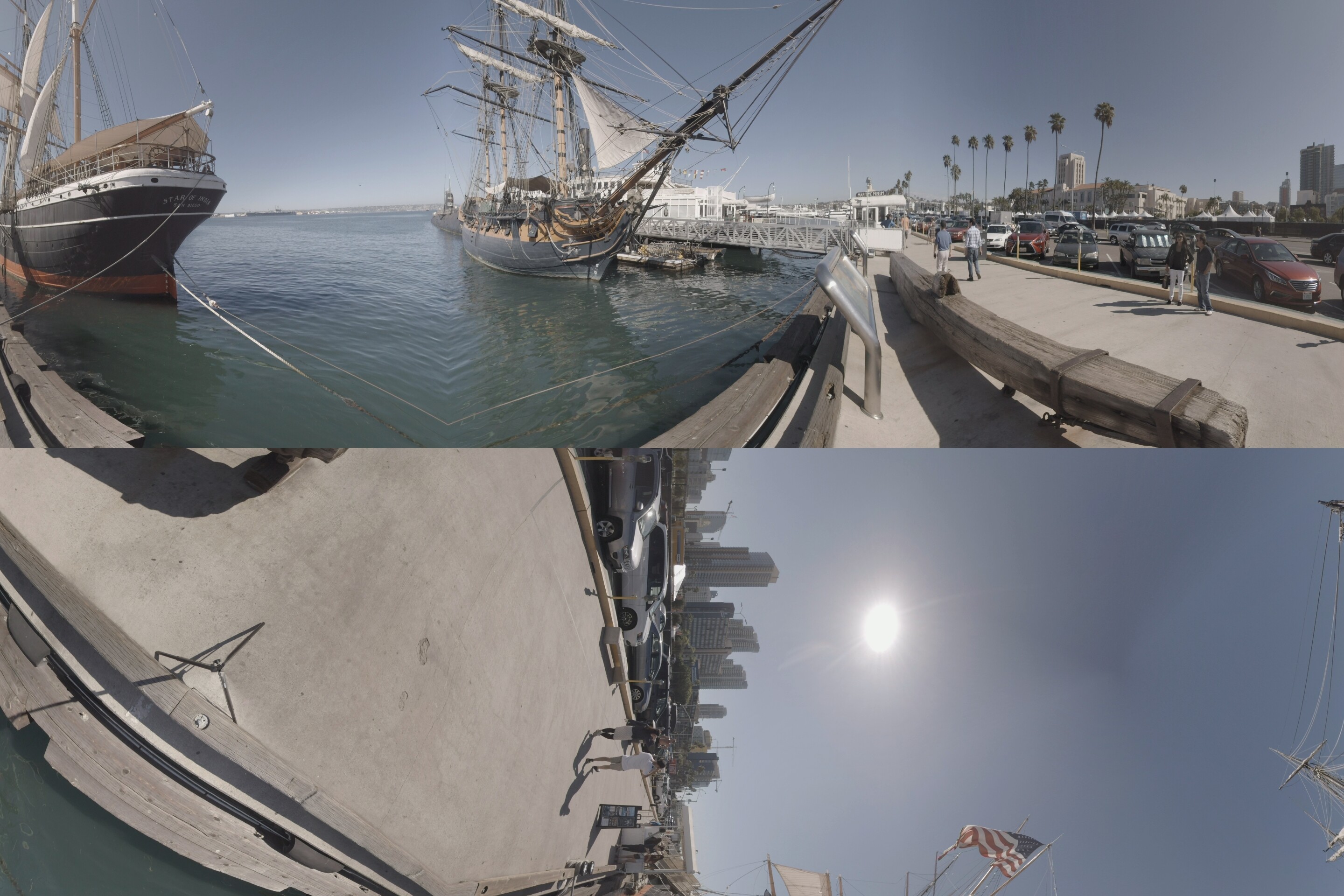}
        \caption{ACP before applying compression}
    \end{subfigure}
    \begin{subfigure}[b]{0.32\textwidth}
        \includegraphics[width=\textwidth]{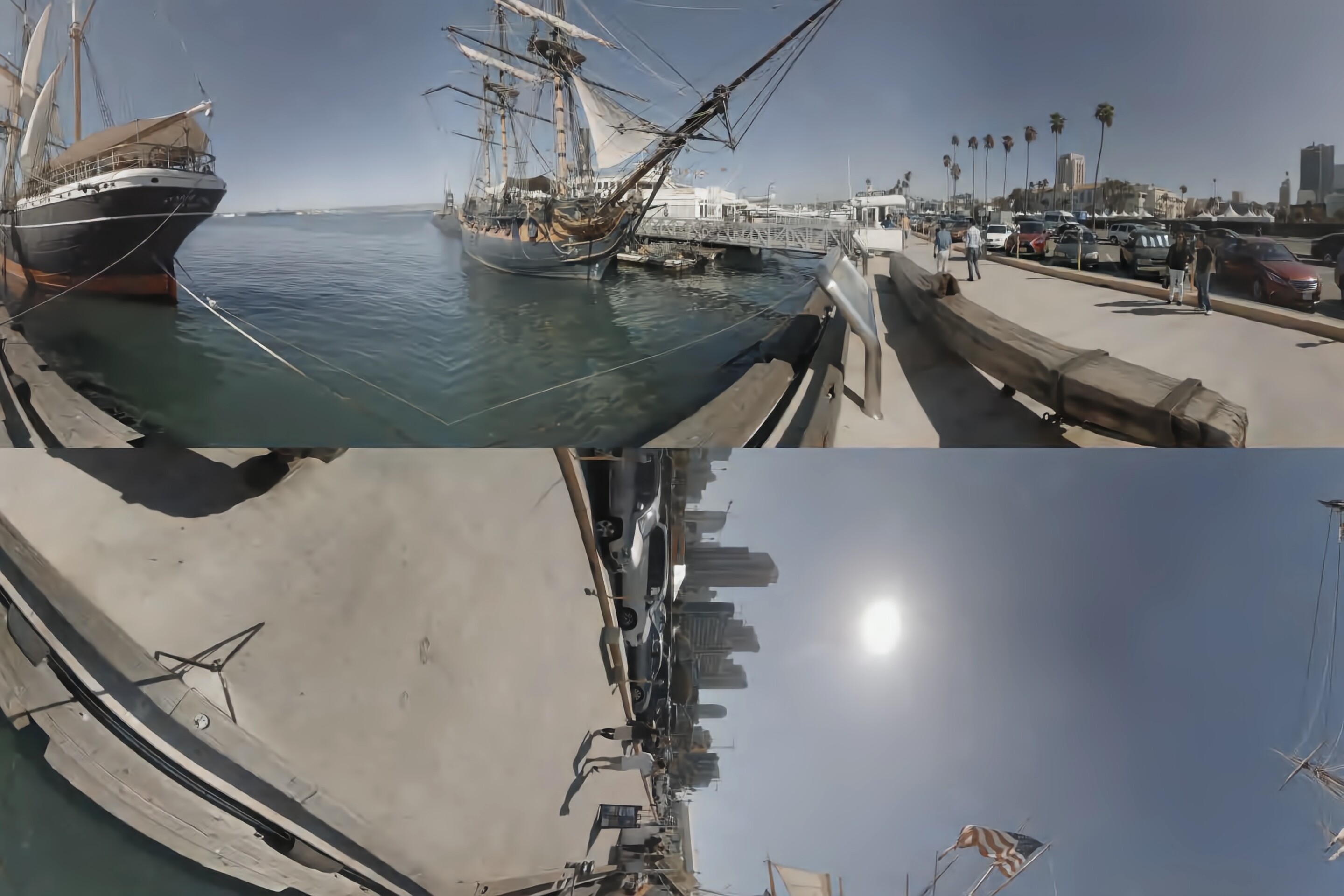}
        \caption{ACP lowest 
        compression level}
    \end{subfigure}
    \begin{subfigure}[b]{0.32\textwidth}
        \includegraphics[width=\textwidth]{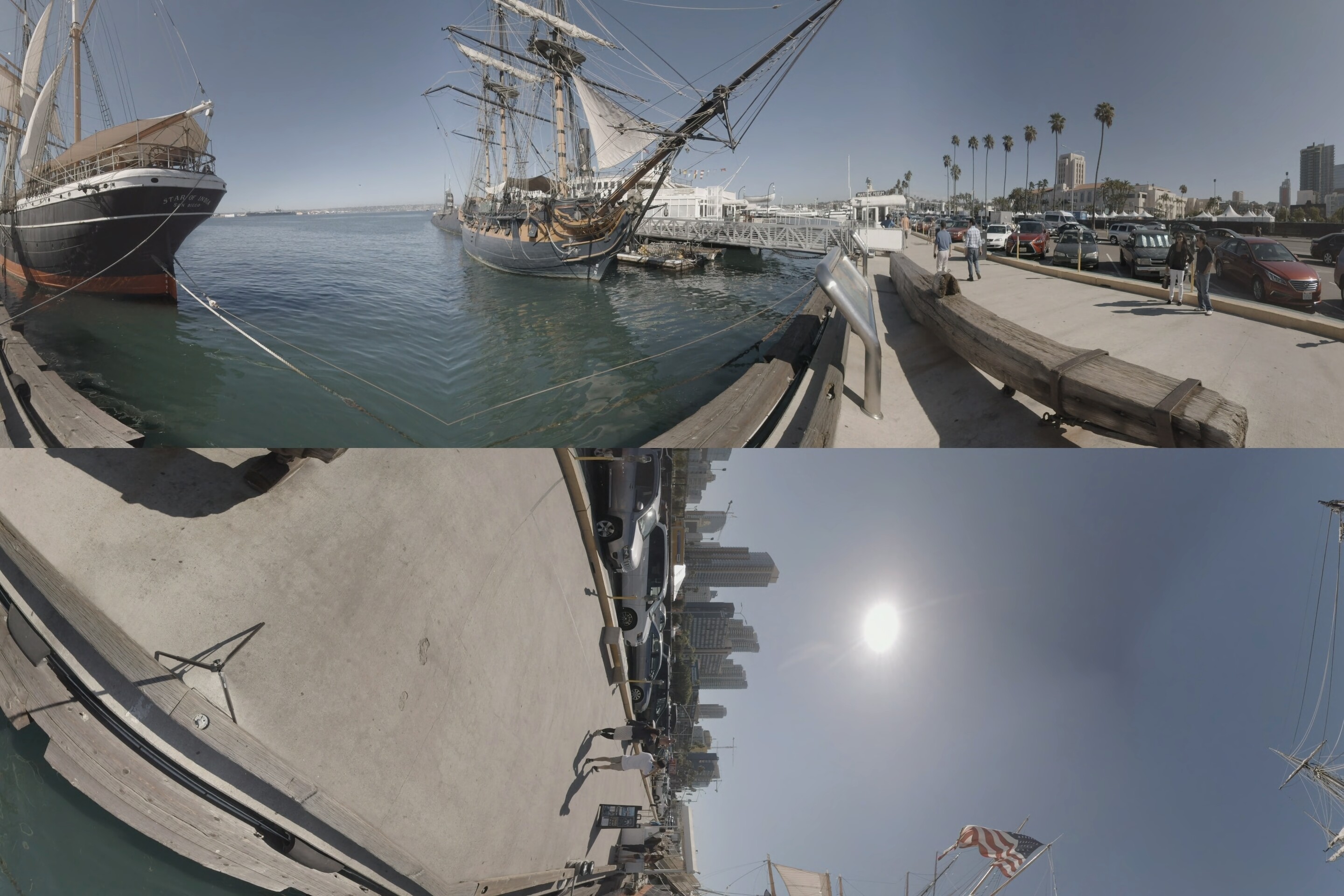}
        \caption{ACP highest
        compression level}
    \end{subfigure}

    \begin{subfigure}[b]{0.32\textwidth}
        \includegraphics[width=\textwidth]{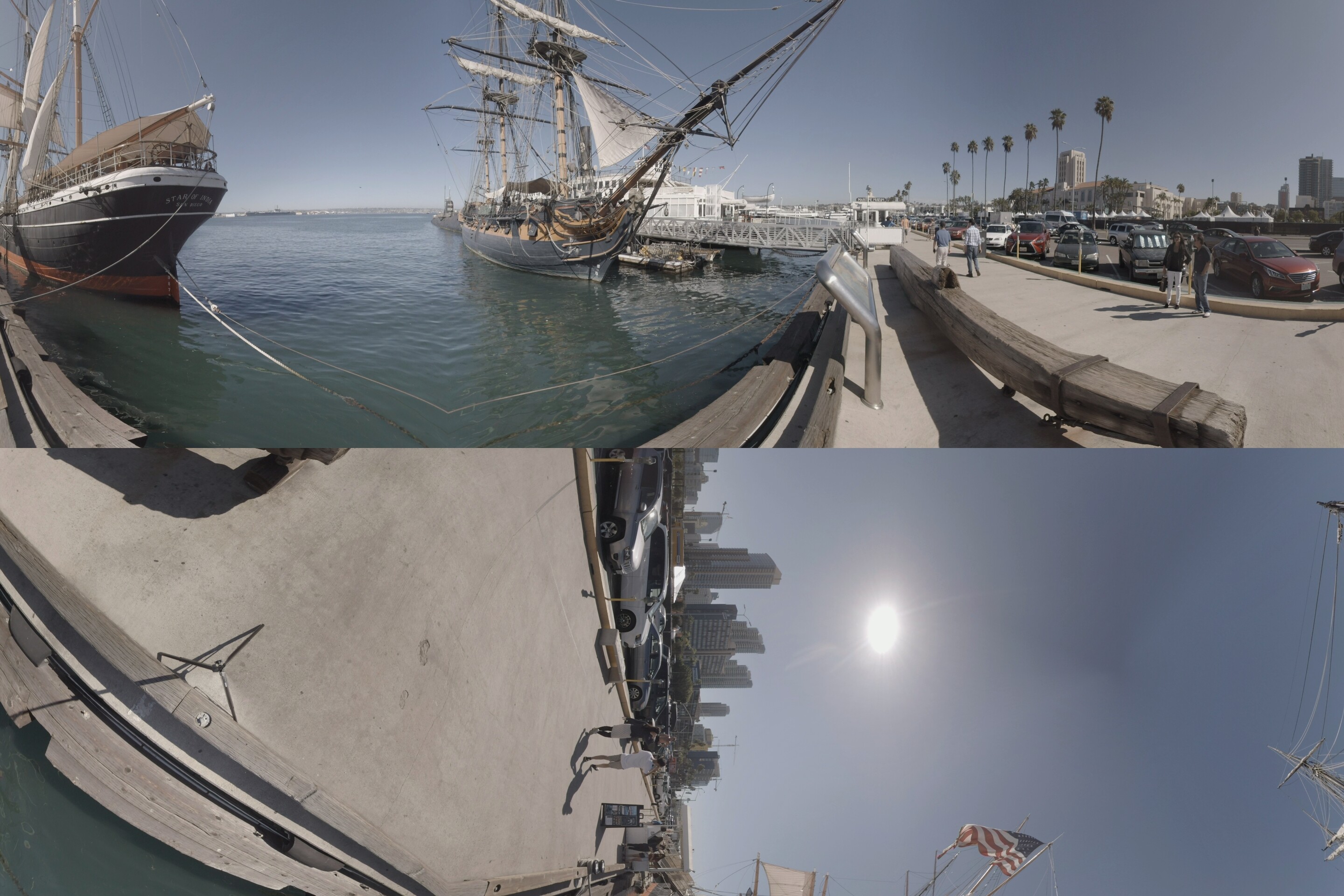}
        \caption{HEC before applying compression}
    \end{subfigure}
    \begin{subfigure}[b]{0.32\textwidth}
        \includegraphics[width=\textwidth]{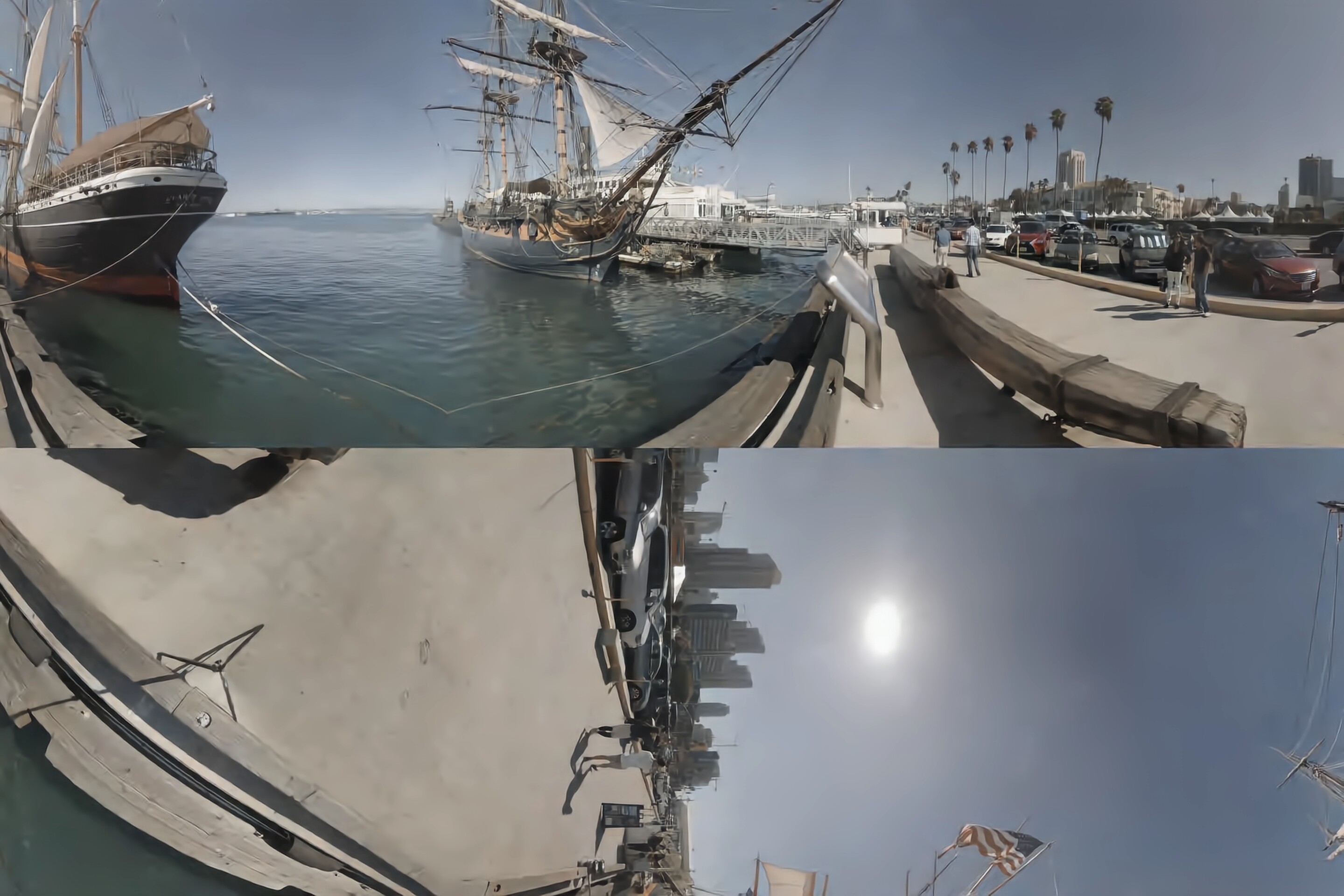}
        \caption{HEC lowest 
        compression level}
    \end{subfigure}
    \begin{subfigure}[b]{0.32\textwidth}
        \includegraphics[width=\textwidth]{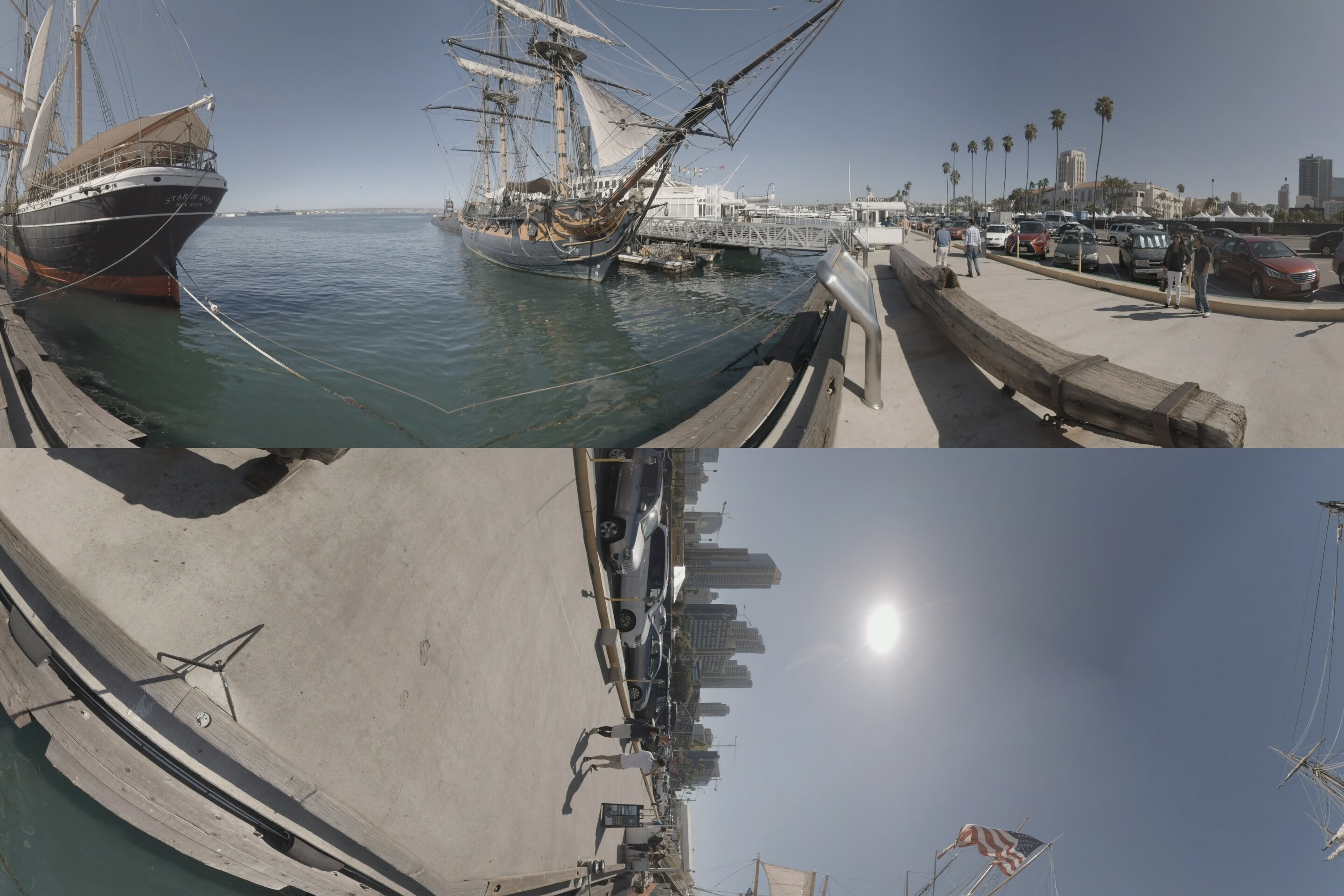}
        \caption{HEC highest
        compression level}
    \end{subfigure}
    
    \caption{Harbor cubemap based projection.}
    \label{fig:Harborcmp}
\end{figure}

Figure \ref{fig:Harborcmp} displays cube map-based projections (CMP, EAC, APC, HEC). In these projections, the frame is transformed into the six faces of a cube. As discussed in Chapter 3, CMP(a) shows uneven sampling on each face due to its rectilinear structure, characterized by a higher density of samples near the edges of the face which decreases in the center. This uneven spherical sampling leads to distorted objects appearing on the CMP faces, negatively impacting the coding efficiency for 360-degree videos. The discontinuous boundaries between the six planar square faces in CMP, with a straight line crossing two adjacent faces appearing to curve at their shared edge, further detract from the overall quality and efficiency of the 360-degree video representation. To address these artifacts, alternative projection methods such as EAC, ACP, and HEC have been proposed. In Figure \ref{fig:Harborcmp} (d), (g), and (j), the EAC, ACP, and HEC projections are shown respectively. These alternative projection methods improve the uniformity of spherical sampling by adjusting the cube faces, with coefficients specifically designed to approximate uniform sphere sampling.

\begin{figure}
    \centering
    \begin{subfigure}[b]{0.32\textwidth}
        \includegraphics[width=\textwidth]{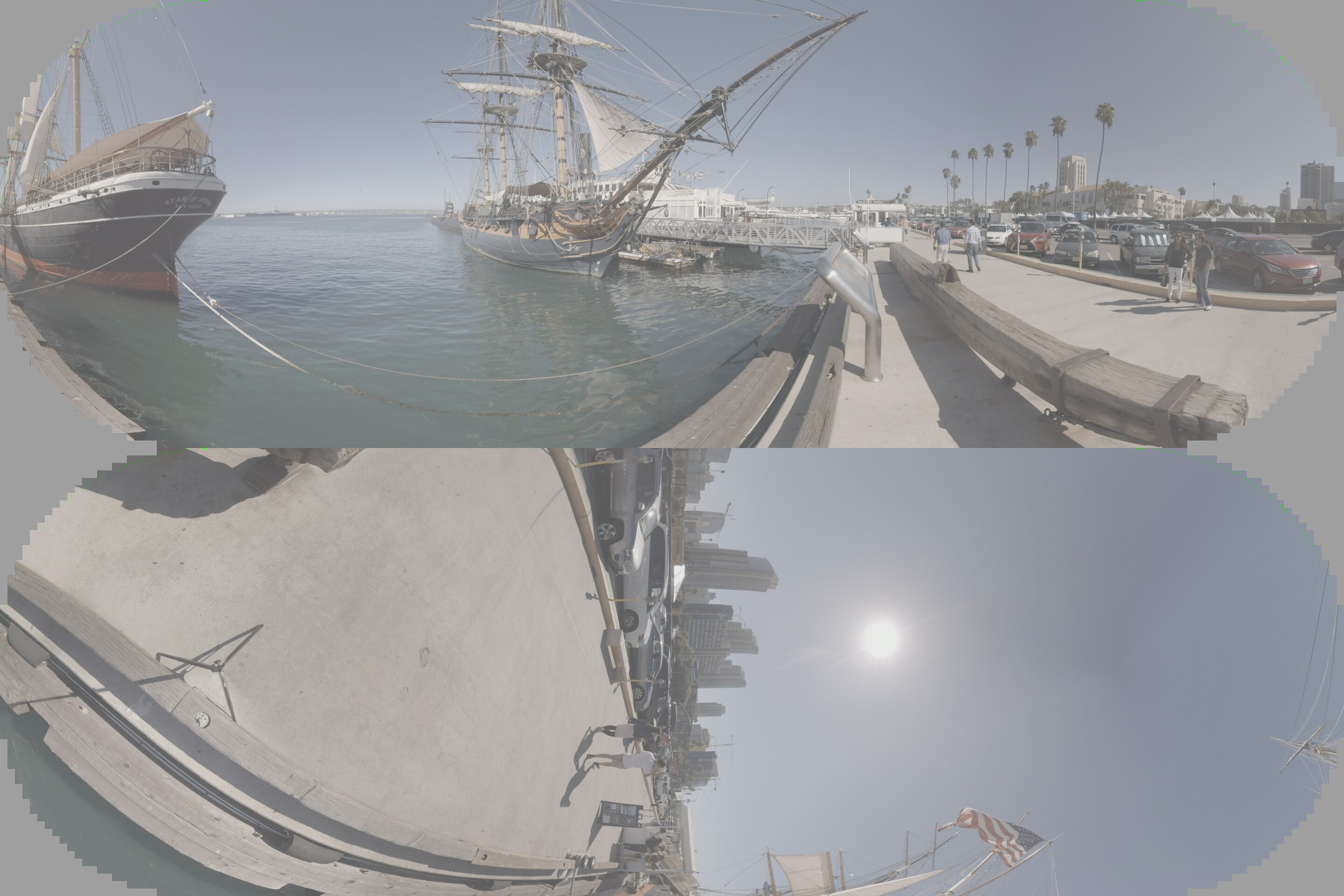}
        \caption{RSP before applying compression}
    \end{subfigure}
    \begin{subfigure}[b]{0.32\textwidth}
        \includegraphics[width=\textwidth]{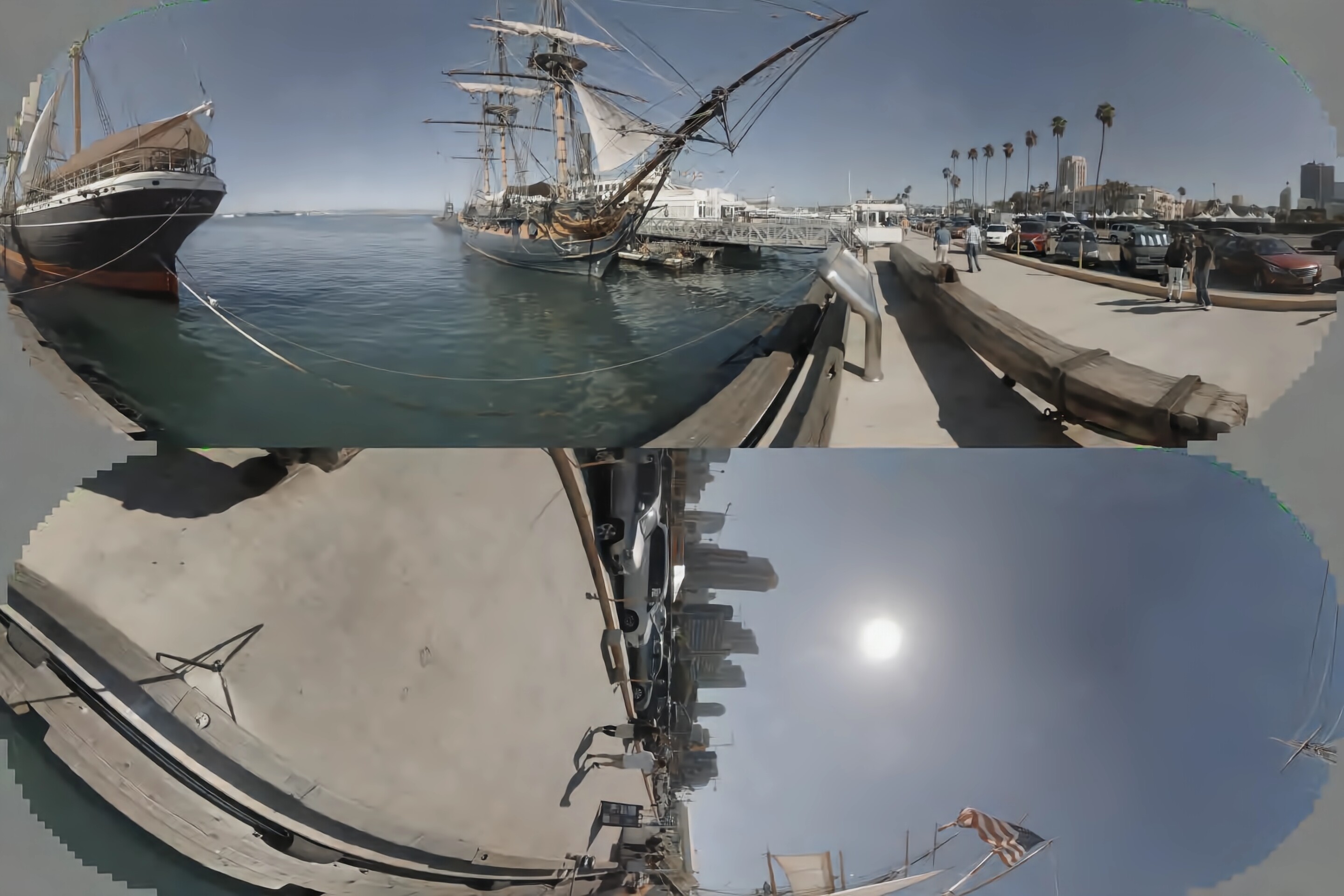}
        \caption{RSP lowest 
        compression level}
    \end{subfigure}
    \begin{subfigure}[b]{0.32\textwidth}
        \includegraphics[width=\textwidth]{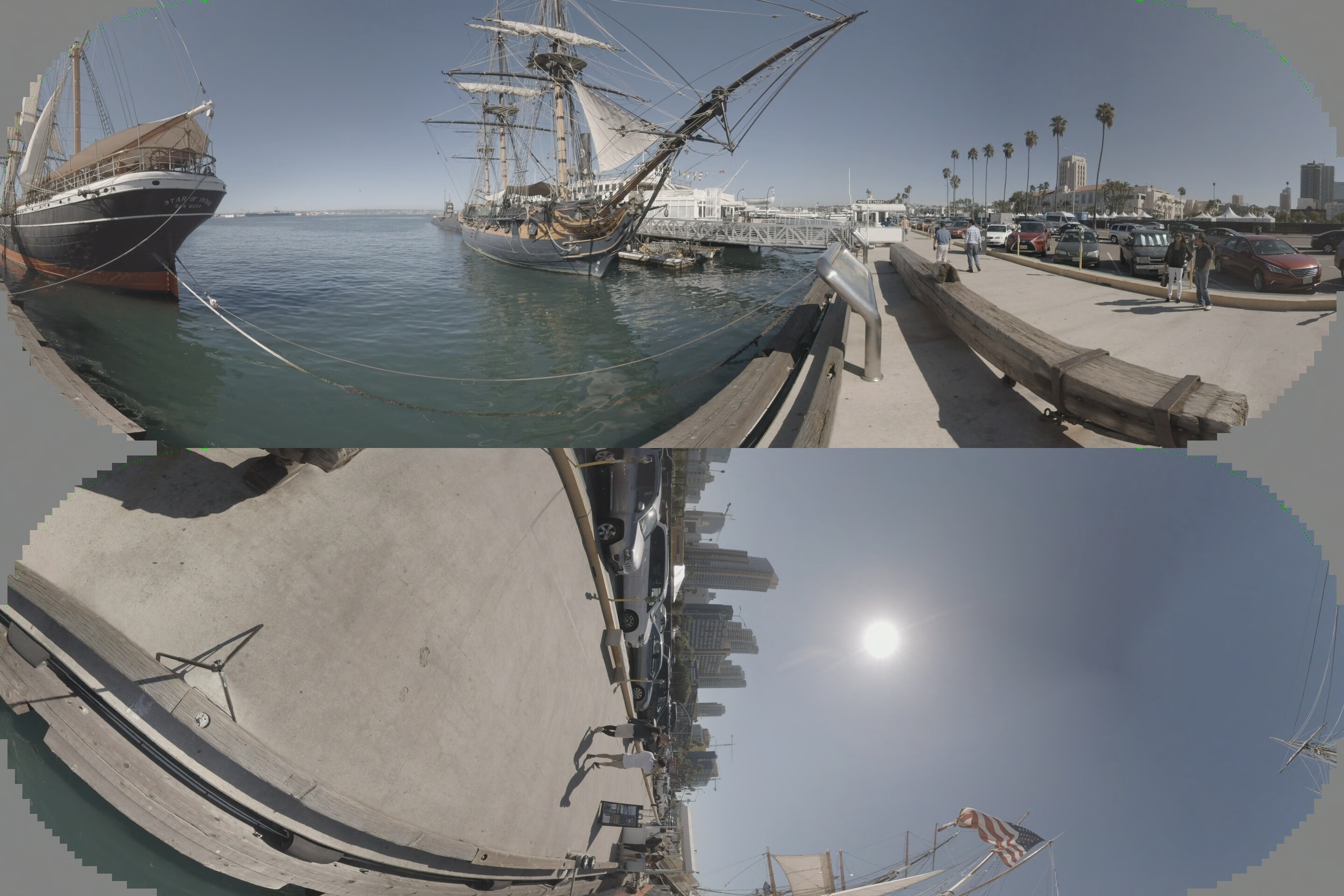}
        \caption{RSP highest
        compression level}
    \end{subfigure}
    
    \caption{Harbor RSP projection.}
    \label{fig:Harborrsp}
\end{figure}

Figure \ref{fig:Harborrsp} presents the RSP projection, which is characterized by round-shaped faces. The RSP layout features two rows of faces. To produce the first row, the central part of an ERP image is cropped, and for the second row, the original ERP picture undergoes spherical rotation, positioning the poles to coincide with the equator and rotating the sphere's back portion to the front.

\begin{figure}
    \centering
    \begin{subfigure}[b]{0.32\textwidth}
        \includegraphics[width=\textwidth]{Harbor_ERP.jpg}
        \caption{ERP before applying compression}
    \end{subfigure}
    \begin{subfigure}[b]{0.32\textwidth}
        \includegraphics[width=\textwidth]{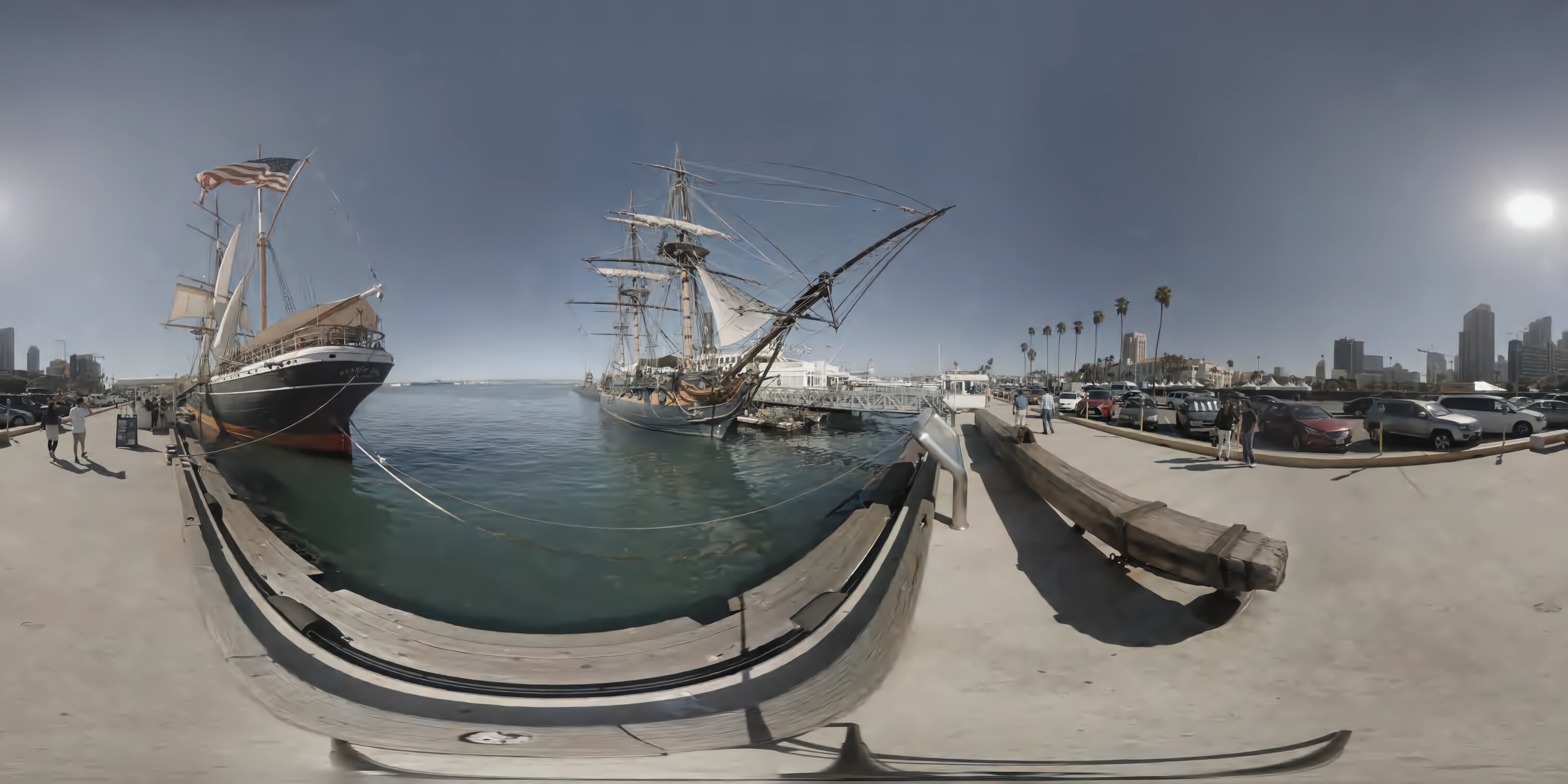}
        \caption{ERP lowest 
        compression level}
    \end{subfigure}
    \begin{subfigure}[b]{0.32\textwidth}
        \includegraphics[width=\textwidth]{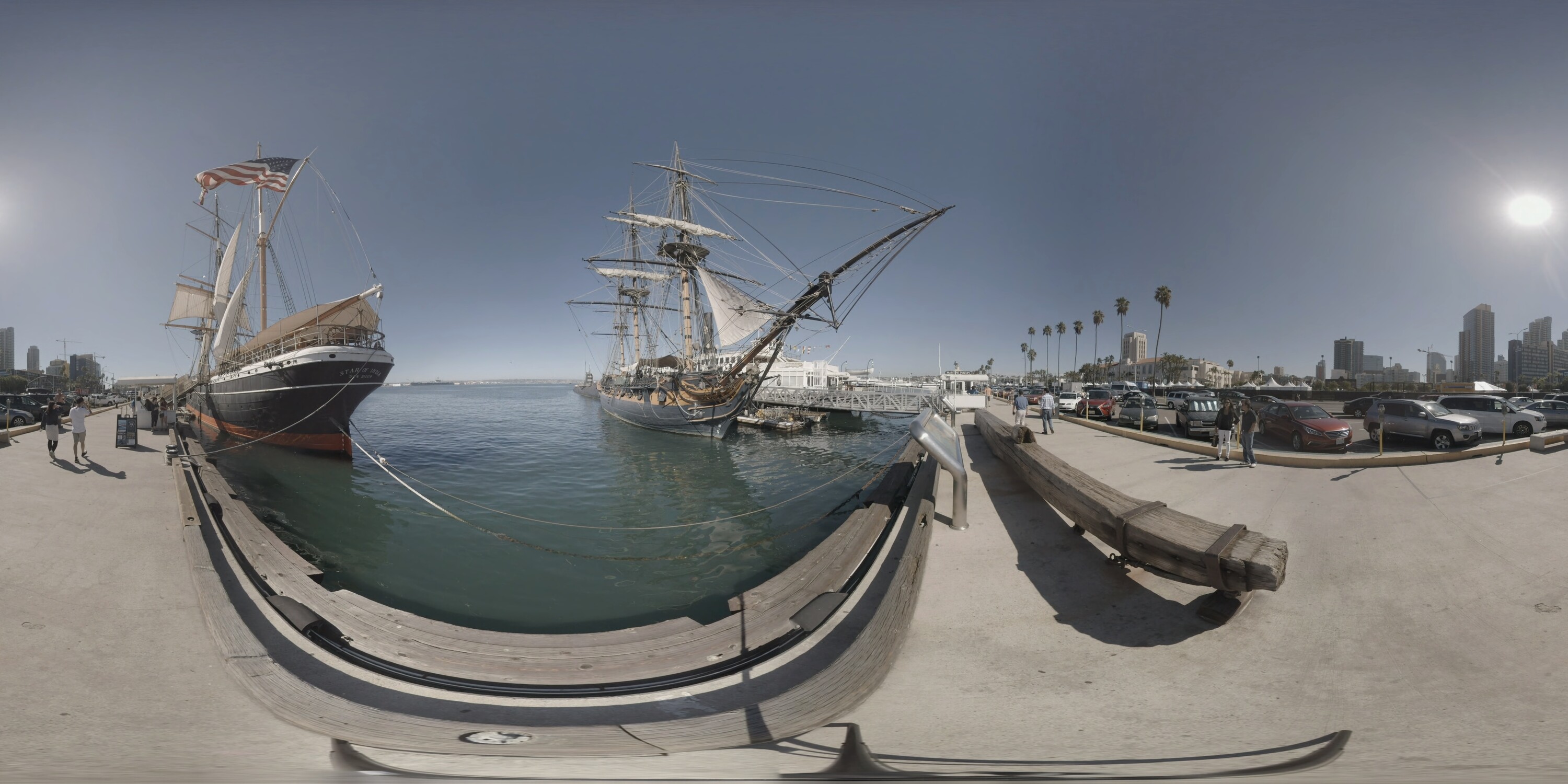}
        \caption{ERP highest
        compression level}
    \end{subfigure}

        \begin{subfigure}[b]{0.32\textwidth}
        \includegraphics[width=\textwidth]{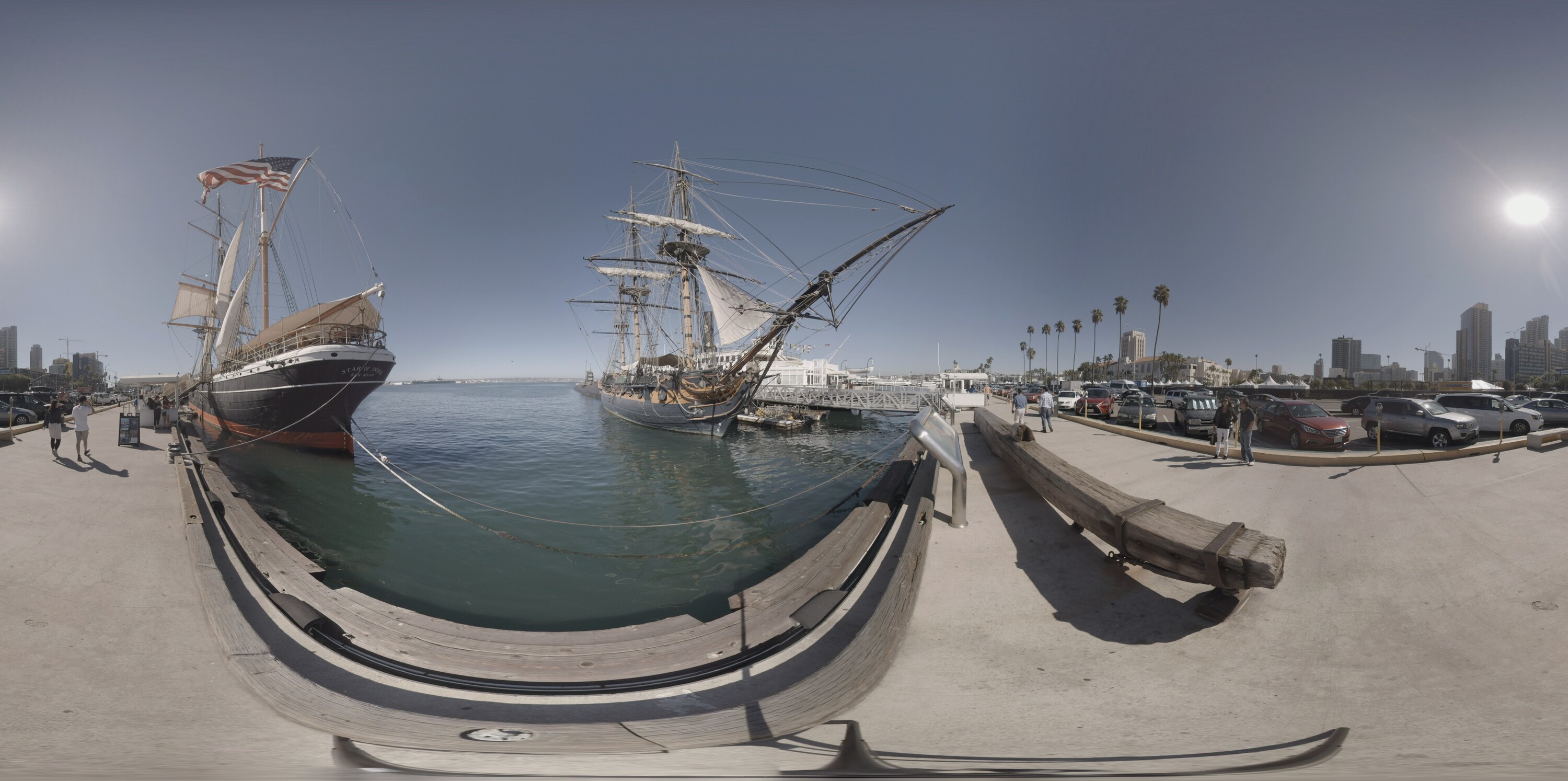}
        \caption{PERP before applying compression}
    \end{subfigure}
    \begin{subfigure}[b]{0.32\textwidth}
        \includegraphics[width=\textwidth]{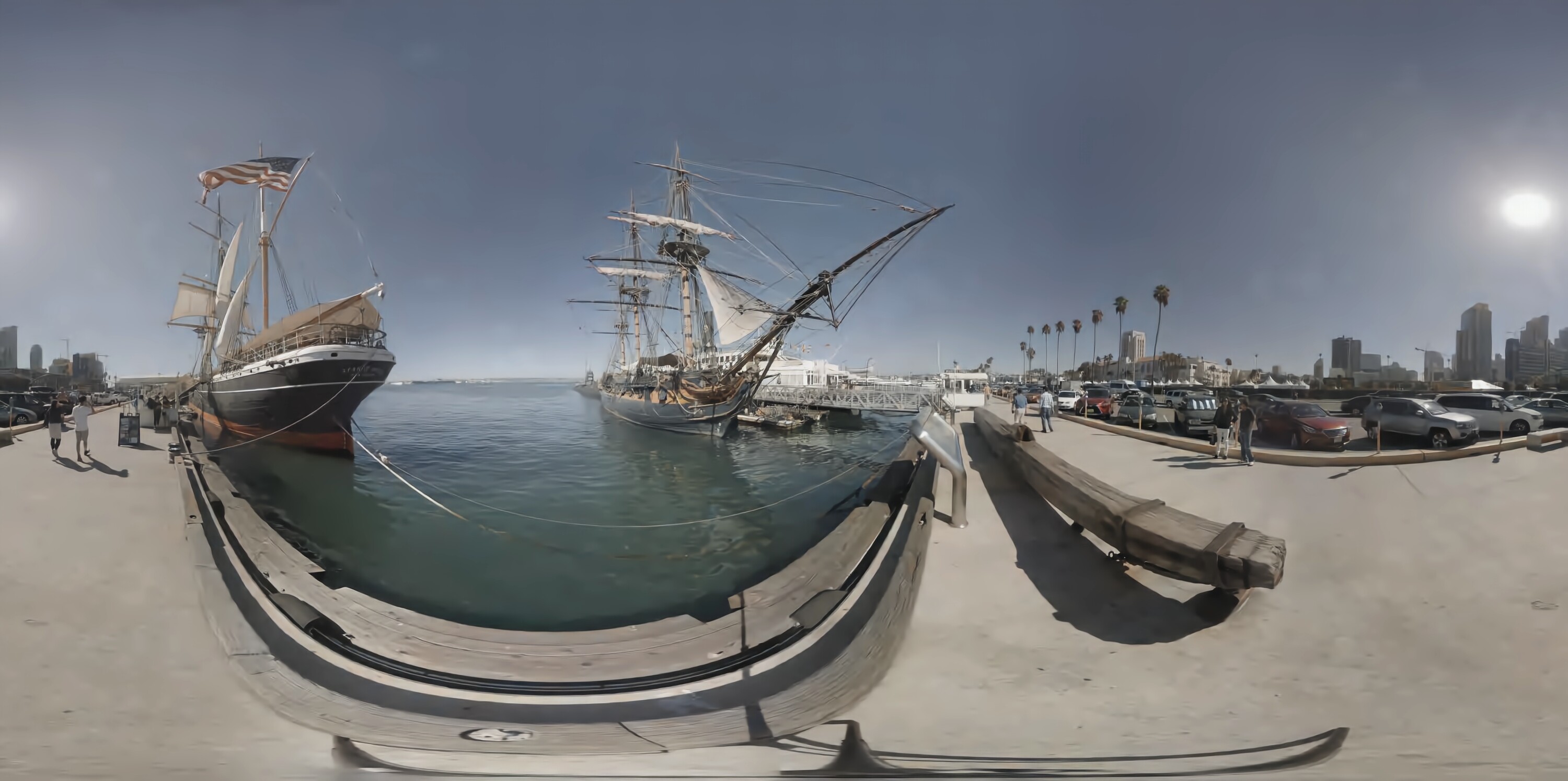}
        \caption{PERP lowest 
        compression level}
    \end{subfigure}
    \begin{subfigure}[b]{0.32\textwidth}
        \includegraphics[width=\textwidth]{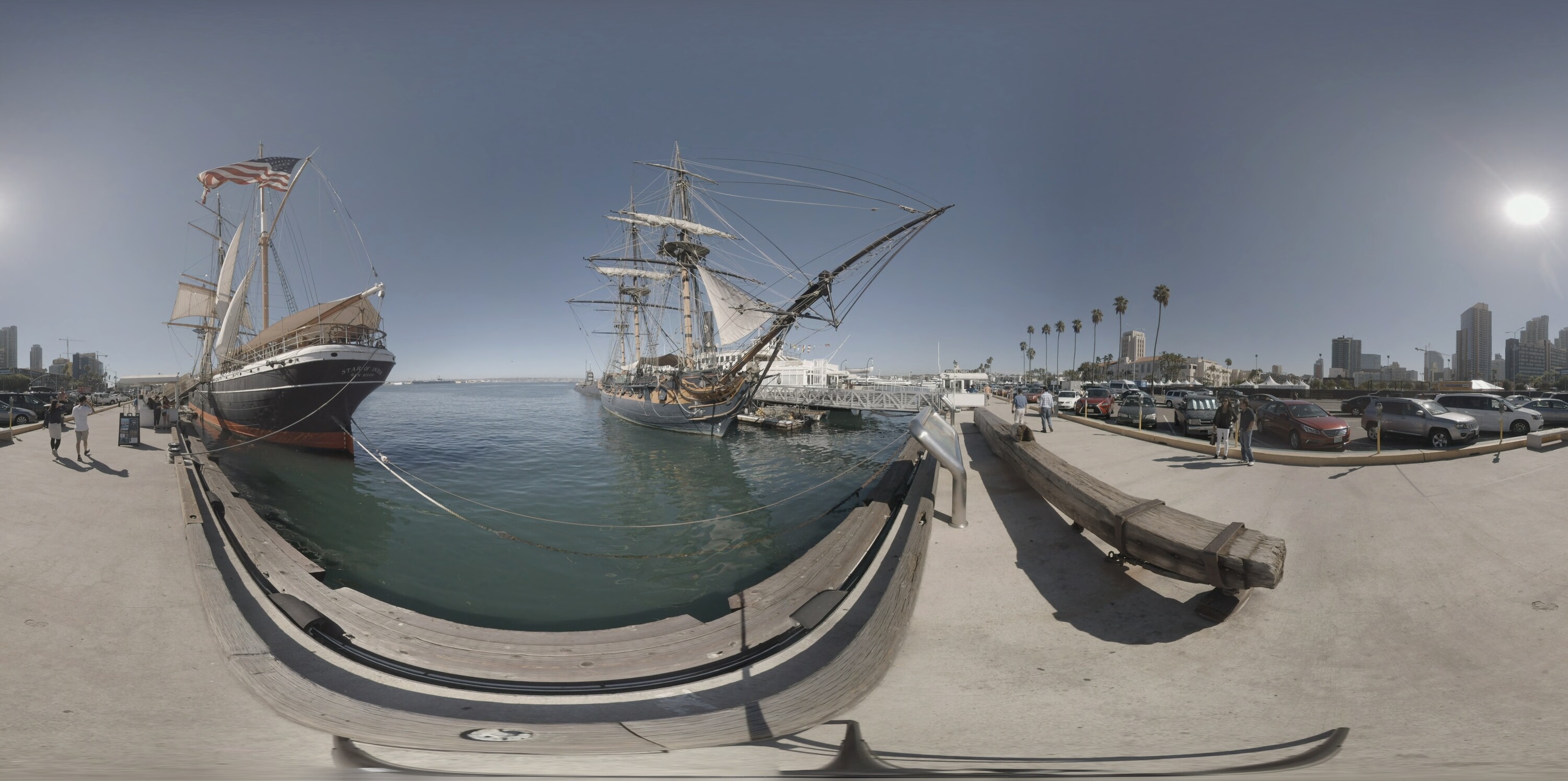}
        \caption{PERP highest
        compression level}
    \end{subfigure}

    \caption{Harbor ERP-based projections.}
    \label{fig:Harborerp}
\end{figure}

\begin{figure}
    \centering
    \begin{subfigure}[b]{0.3\textwidth}
        \includegraphics[width=\textwidth]{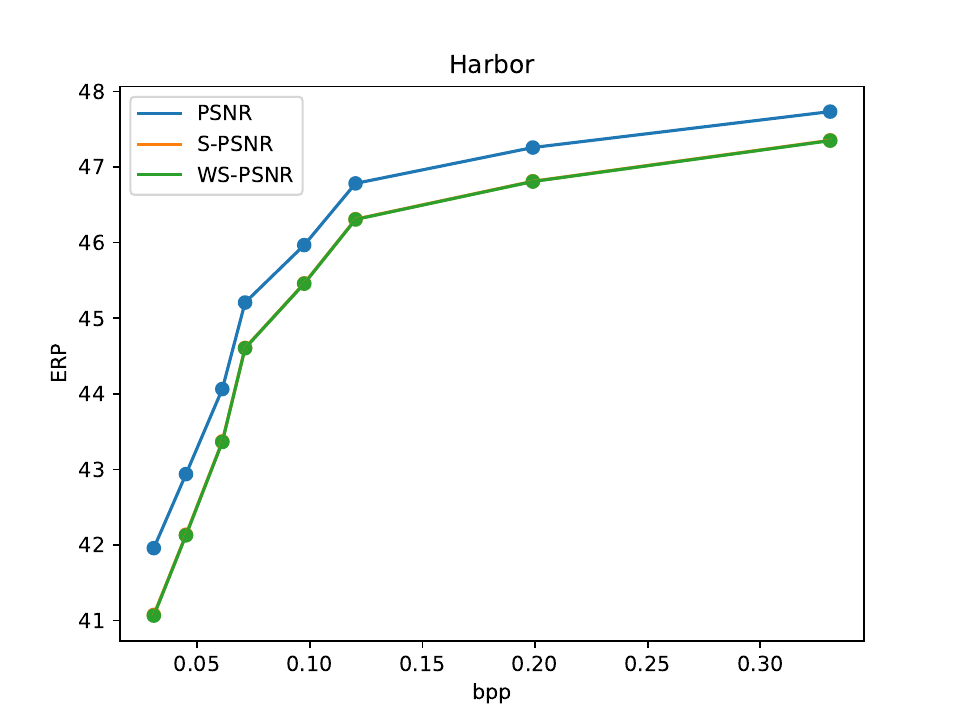}
        \caption{ERP}
    \end{subfigure}
    \begin{subfigure}[b]{0.3\textwidth}
        \includegraphics[width=\textwidth]{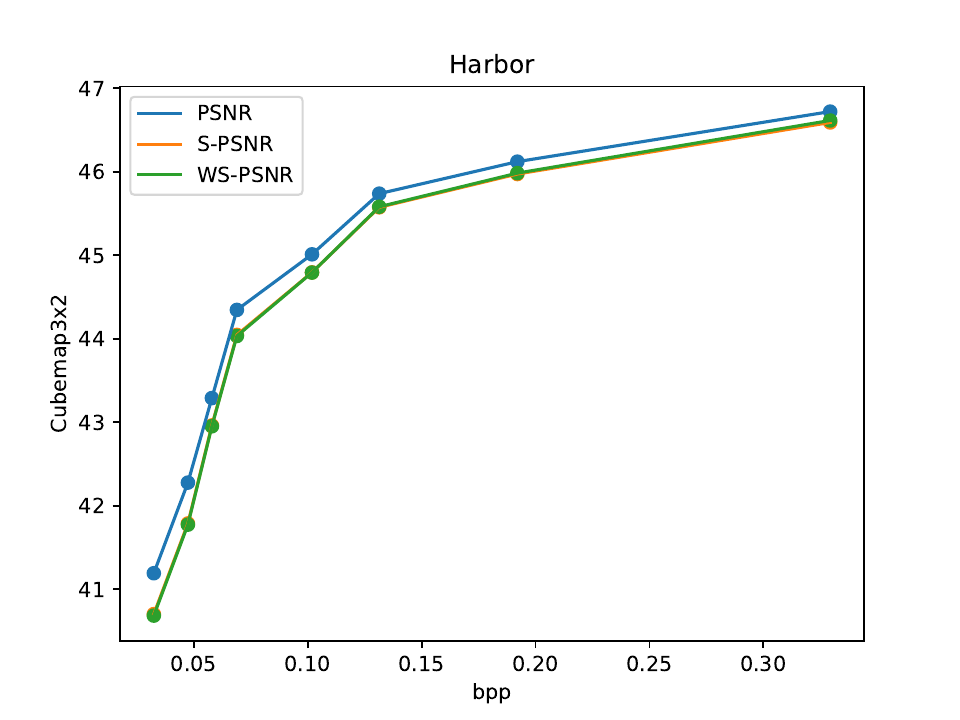}
        \caption{CMP}
    \end{subfigure}
    \begin{subfigure}[b]{0.3\textwidth}
        \includegraphics[width=\textwidth]{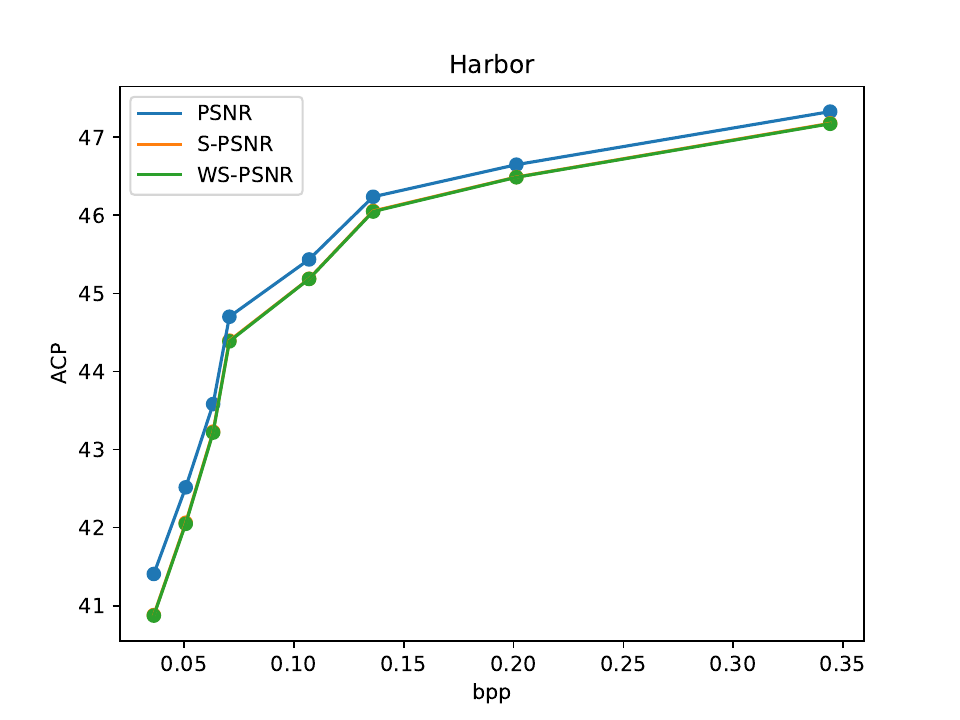}
        \caption{ACP}
    \end{subfigure}
    
    \begin{subfigure}[b]{0.3\textwidth}
        \includegraphics[width=\textwidth]{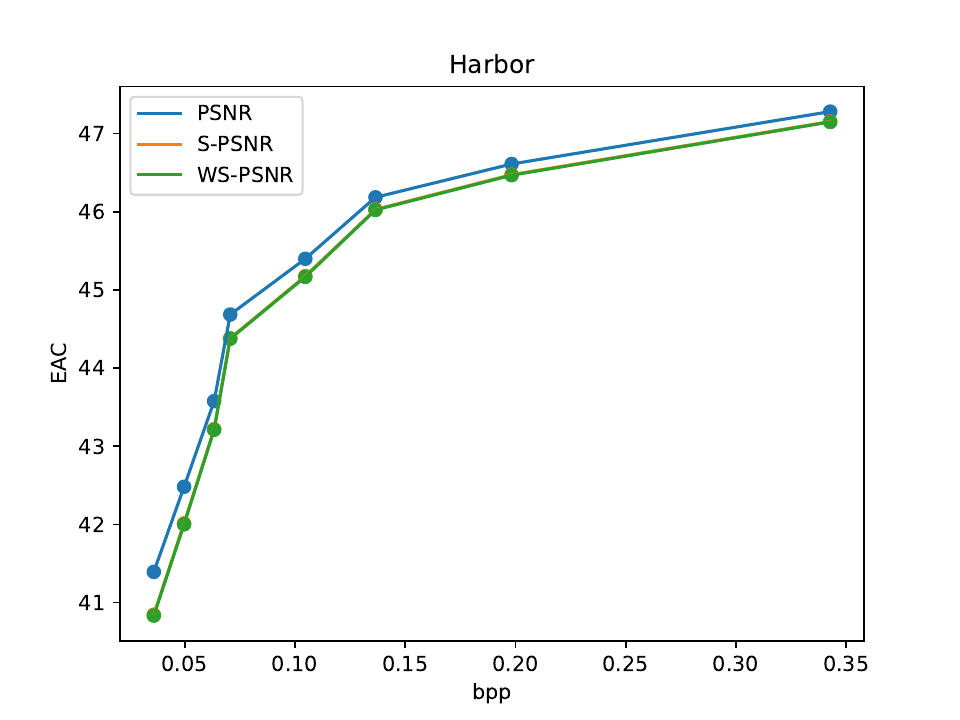}
        \caption{EAC}
    \end{subfigure}
    \begin{subfigure}[b]{0.3\textwidth}
        \includegraphics[width=\textwidth]{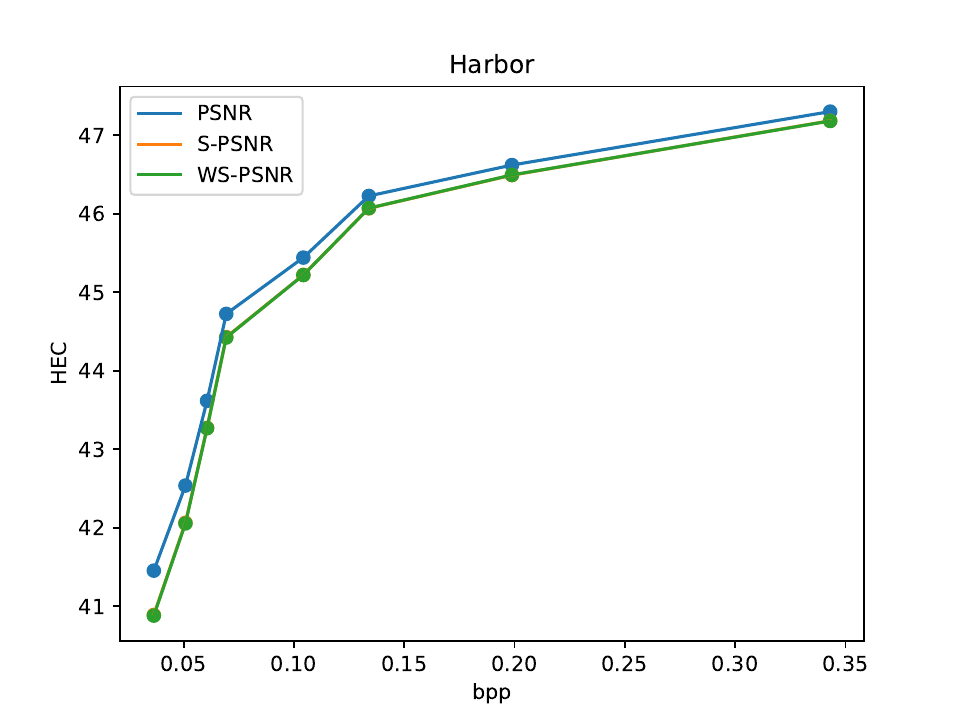}
        \caption{HEC}
    \end{subfigure}
    \begin{subfigure}[b]{0.3\textwidth}
        \includegraphics[width=\textwidth]{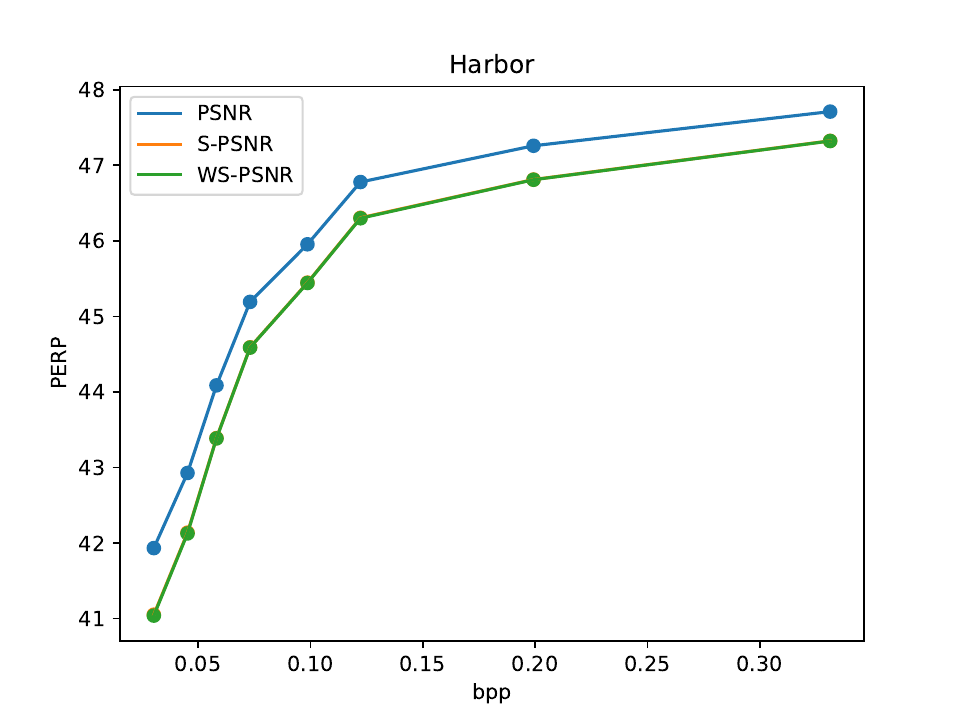}
        \caption{PERP}
    \end{subfigure}
    
    \begin{subfigure}[b]{0.3\textwidth}
        \includegraphics[width=\textwidth]{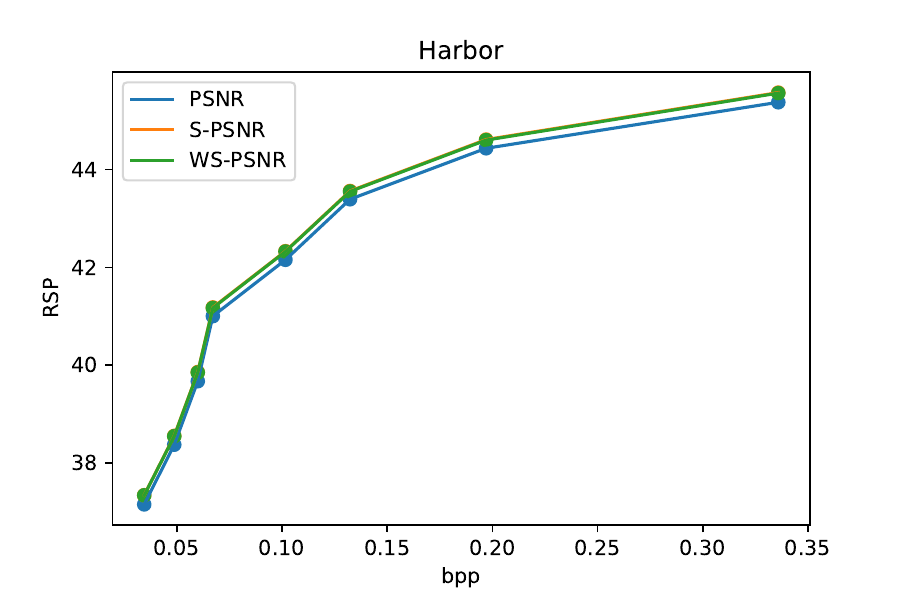}
        \caption{RSP}
    \end{subfigure}
    
    \caption{Harbor end-to-end (PSNR, S-PSNR, WS-PSNR) rate-distortion curves.}
    \label{fig:Harbor-psnr}
\end{figure}

\subsection{Projection compression}

At this stage, the compressed projected formats can be evaluated using evaluation metrics such as PSNR, S-PSNR, and WS-PSNR, which can be calculated using 360Lib software. To compute these metrics, 360Lib requires a reference file (projected frames before compression) and a source file (projected frames after compression). By utilizing these metrics and the bits-per-pixel (bpp) calculated from the compression model, rate-distortion curves for each projection can be generated.

The generated rate-distortion curves serve as the main metric for comparing projection performance. To create these curves for each metric, the average of PSNR, S-PSNR, and WS-PSNR is calculated for all frames of each projection at each compression level. The average PSNR and the calculated bits per pixel at each compression level correspond to individual points on the rate-distortion curves.

\begin{figure}
    \centering
    \begin{subfigure}[b]{0.3\textwidth}
        \includegraphics[width=\textwidth]{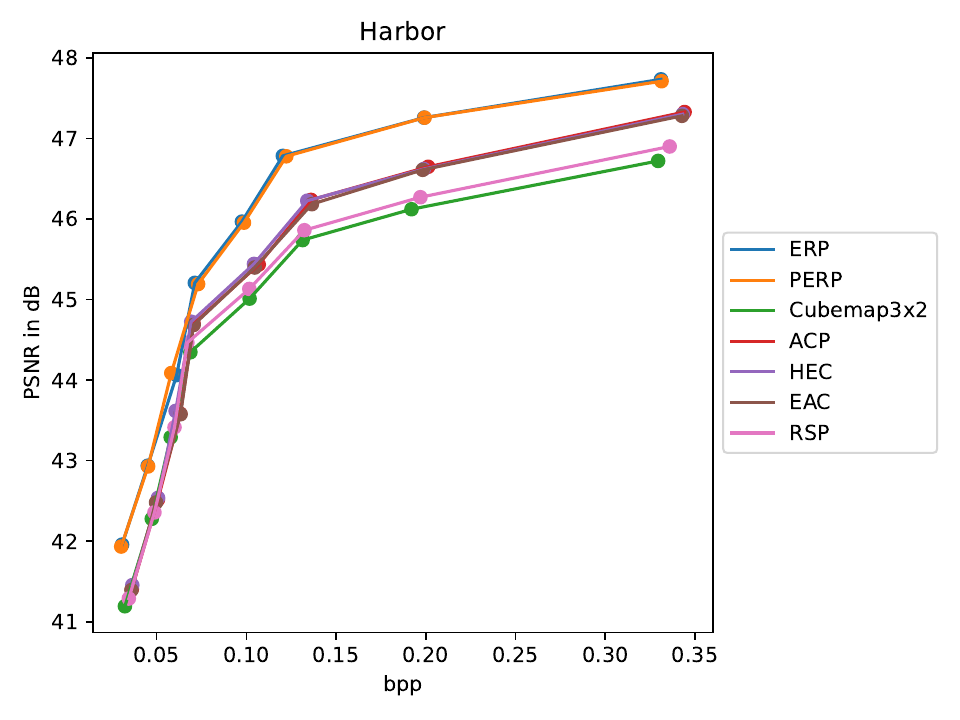}
        \caption{PSNR}
    \end{subfigure}
    \begin{subfigure}[b]{0.3\textwidth}
        \includegraphics[width=\textwidth]{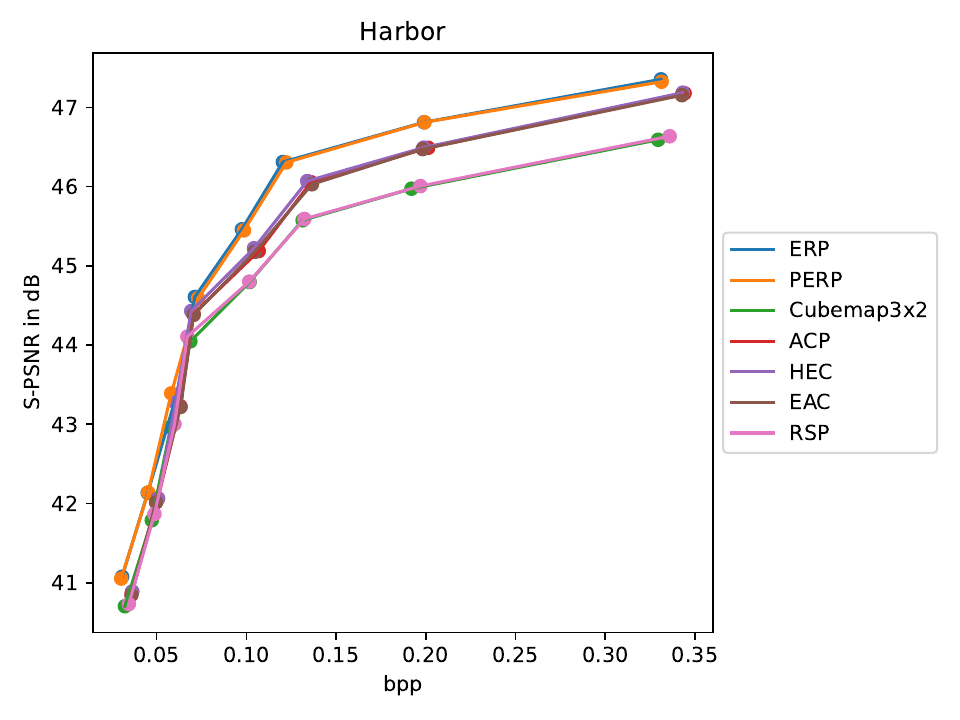}
        \caption{S-PSNR}
    \end{subfigure}
    \begin{subfigure}[b]{0.3\textwidth}
        \includegraphics[width=\textwidth]{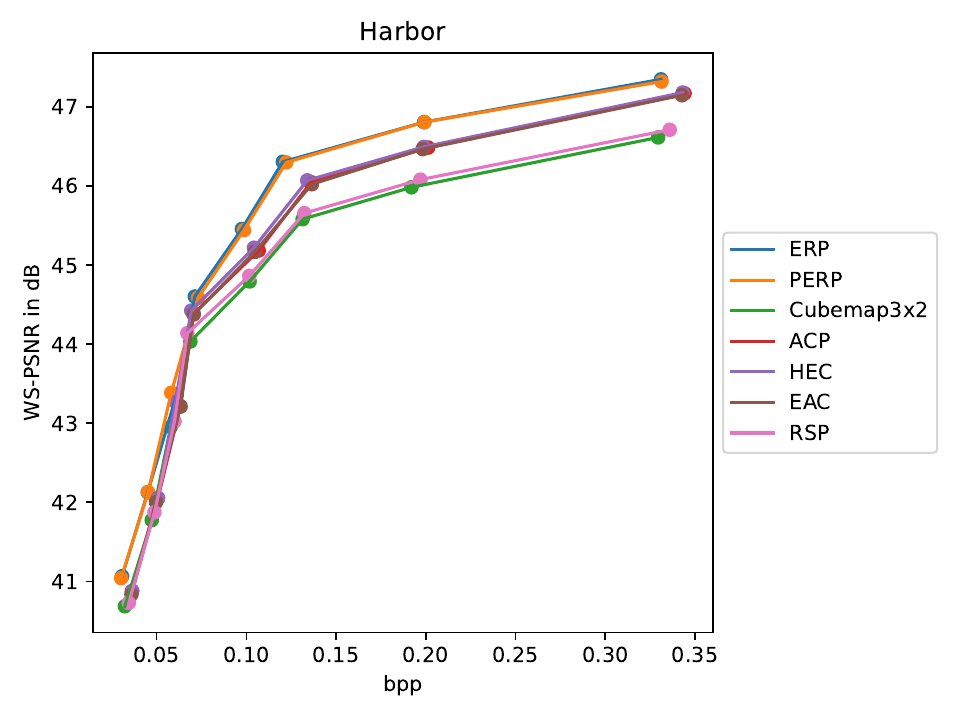}
        \caption{WS-PSNR}
    \end{subfigure}  
\caption{Harbor projections comparison end-to-end}
\label{fig:Harbor3}
\end{figure}

Figure \ref{fig:Harbor-psnr} showcases the rate-distortion curves for the Harbor sequence with each projection present. The PSNR, S-PSNR, and WS-PSNR calculated here are between the reference projection format and the compressed frame in the reference projection format, indicating an end-to-end measurement. As seen, all rate-distortion curves increase with higher bpp, demonstrating that the pipeline is functioning as intended. These rate-distortion curves are collectively displayed in Figure \ref{fig:Harbor3}. It is noted that the cube map-based projections (CMP, ACP, EAC, HEC) exhibit similar performance, while ERP-based projections (ERP and PERP) outperform the others. These figures show that ERP and Padded ERP surpass the other projections, while RSP and CMP demonstrate the lowest rate-distortion values, with EAC, HEC, and ACP positioned closely in between.

\subsection{Projection comparison}

To determine which projection offers the best performance, the differences between the rate-distortion curves of each projection need to be compared. The Bjøntegaard-Delta interpolation method was employed for this purpose, and the results are presented in Table \ref{table:projection_comparison}. The ERP projection is used as a reference in this table, and the other projections are represented by their differences relative to ERP. Each row in Table \ref{table:projection_comparison} shows the difference between the ERP and the other listed projections for each sequence.

Positive bd-rate values in this table indicate that ERP outperforms the corresponding projection. The larger the bd-rate value, the greater the difference between that projection and ERP. As observed, all bd-rate values for cube map-based and RSP-based projections are positive, suggesting that the ERP projection outperforms all of these projections. The bd-rate value quantifies the magnitude of this performance gap. \\

It is also worth mentioning that the PERP projection, a variation of ERP, exhibits performance closest to ERP. In all sequences except for Harbor and PoleVault, the PERP projection outperforms ERP. To confirm the results indicating ERP's superior performance over other projections, experiments were conducted using alternative projections including adjusted equal-area projection (AEP), generalized cubemap projection format (GCMP), cube map 3x2 SEI, and cubemap 4x3. In all cases, the same results were obtained, demonstrating that ERP and PERP consistently outperform all other projections.

\begin{table}[t]
\caption{Comparison of projections compressed using scale space flow model}
\centering
\begin{tabular}{|c|c|c|c|c|c|c|}
\hline
\textbf{Sequance} &\textbf{PERP} &\textbf{CMP} &\textbf{EAC} & \textbf{ACP}  &\textbf{HEC} &\textbf{RSP} \\
\hline
Harbor               & 0.204   & 48.629 & 36.303  & 35.246 & 34.548 & 35.155                \\
\hline
SkateboardInLot      & -0.543  & 52.856 & 40.941 & 40.777 & 39.167 & 52.688               \\
\hline
ChairLift            & -0.430  & 59.634 & 47.645 & 46.683 & 46.020 & 54.958               \\
\hline
KiteFlite            & -0.150  & 39.792 & 27.942 & 27.783 & 27.308 & 28.243              \\
\hline
GasLamp              & -0.467  & 33.156 & 20.83  & 19.720 & 17.316 & 17.056               \\
\hline
Balboa               & -1.397  & 43.245 & 38.428 & 39.376 & 37.019 & 45.708               \\
\hline
Broadway             & -1.536  & 39.224 & 31.077 & 31.015 & 31.270 & 39.327                \\
\hline
Landing2             & -0.660  & 67.419 & 51.638 & 51.115 & 50.313 & 51.065               \\
\hline
BranCastle2          & -0.689  & 68.186 & 52.960 & 52.498 & 52.601 & 54.698               \\
\hline
AerialCity           & -0.849  & 58.326 & 43.569 & 48.173 & 47.539 & 53.977              \\
\hline
PoleVault            & 0.107   & 63.895 & 47.418 & 46.668 & 47.228 & 47.089             \\
\hline
\textbf{Average}     & \textbf{-0.4629} & \textbf{52.4733} & \textbf{41.9074} & \textbf{39.8417} & \textbf{39.8753} & \textbf{43.6056} \\
\hline
\end{tabular}
\label{table:projection_comparison}
\end{table}

Now that the influence of different projections on the compression performance of 360-degree videos using a learning-based compression model has been demonstrated, it is crucial to compare these results with other pipelines using the same method and traditional compression techniques such as HM-16.16. Table \ref{tab:HM-16.16} presents the coding performance of different projections based on the HM-16.16 standard. In this table, the ERP projection serves as a reference, and other projections are represented by their differences relative to ERP. Each row in Table \ref{table:projection_comparison} displays the difference between PERP and the other listed projections for each type of sequence. The results demonstrate that all listed projections (CMP, ACP, RSP, EAC) outperform PERP in this pipeline, as their bd\_rates are negative. EAC and ACP exhibit similar performance and are superior to the other projections. In contrast, CMP has the lowest performance compared to the others, while RSP falls within these two sets. These results indicate that in traditional video codecs, which are primarily hybrid video codecs, cubemap-based projections outperform ERP-based projections.
While the results stated in Table \ref{table:projection_comparison} indicate that ERP-based projections have better performance than CMP-based projections when utilizing the scale-space flow model for compression. This highlights that learning-based models have a fundamentally different approach towards projection compared to traditional video codecs. In hybrid video codecs, PERP has the worst performance, which contrasts with its performance in learning-based compression. Putting this aside, the order in which CMP-based projections perform is similar. In both methods, EAC and ACP outperform CMP and RSP, and RSP shows better results compared to CMP. The discrepancy in the performance of the projections across different codecs suggests that learning-based models may offer unique advantages when it comes to video compression. This could warrant further investigation into their potential applications and optimizations, especially as technology continues to advance and new codecs and compression techniques are developed.

The differing performance of hybrid video codecs and learning-based video codecs for various projections can be attributed to their distinct architectures.  Hybrid video codecs perform better with CMP-based projections due to their block-based approach, which is compatible with the multiple faces of CMP. This structure allows hybrid codecs to naturally partition spherical content into blocks and utilize different motion vectors for each block to predict and compensate for changes between frames. Consequently, the compatibility between the CMP geometry and hybrid codecs' block-based coding techniques leads to improved compression performance and overall better coding efficiency, making CMP a more suitable choice for applications where high-quality spherical content representation is vital. While in learning-based compression models like scale-space flow, compression is pixel-based, and optical flow is computed for each pixel. When working with single-face projection formats like Equirectangular Projection (ERP), this results in smoother motion vectors and increased compression efficiency. However, when dealing with CubeMap Projection (CMP) based formats, which have multiple faces, discontinuities arise in motion vector calculations, as each vector travels in different directions. \\

This leads to a decline in the performance of the compression model. Indeed, training the learning-based compression models with multi-face projection data can be helpful in improving their performance when dealing with such formats. By incorporating a diverse set of multi-face projection data, such as CubeMap Projection (CMP), into the training dataset, the model can learn to recognize and adapt to the unique challenges associated with these formats. This approach can help the model better handle discontinuities in motion vector calculations and develop more robust compression strategies for multi-face projections.

\begin{table}[t]
\caption{Comparison of projections compressed using HM-16.16 codec \cite{he2018jvet}.}
\centering
\begin{tabular}{|c|c|c|c|c|}
\hline
Sequences & \textbf{CMP} & \textbf{ACP} & \textbf{RSP} & \textbf{EAC} \\
\hline
\textbf{All sequences } & -1.1 & -6.9 & -5.7 & -6.9 \\
\hline
\textbf{8K sequences } & -3.9 & -8.4 & -8.2 & -8.5 \\
\hline
\textbf{6K sequences } & 2.9 & -4.9 & -2.6 & -5.1 \\
\hline
\textbf{4K sequences } & -0.7 & -6.4 & -4.9 & -6.1 \\
\hline
\textbf{Average} & -0.7 & -6.65 & -5.35 & -6.65 \\
\hline
\end{tabular}
\label{tab:HM-16.16}
\end{table}

\section{Evaluation of the projection layer}

Now that the best-performing projection has been identified, the performance of a fully end-to-end model for applying projection compression and converting the projection back to the reference will be demonstrated. All steps are performed using a single model implemented in Python, consisting of three main parts, which were explained in detail in the previous chapter. In this section, the performance of this model will be evaluated and compared to the projection conversion performed by the 360lib pipeline.

\begin{figure} [t]
    \centering
    \begin{subfigure}[b]{0.3\textwidth}
        \includegraphics[width=\textwidth]{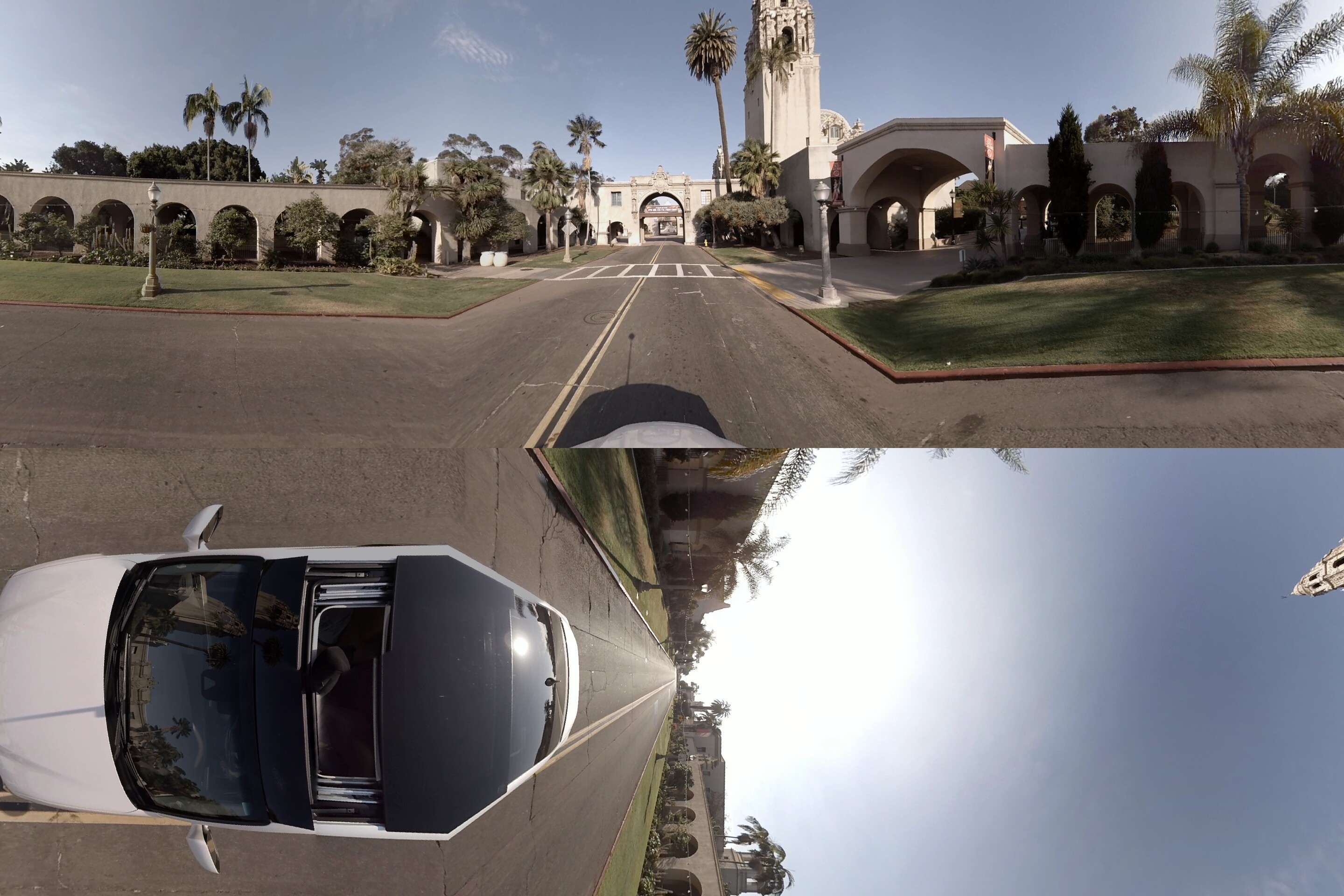}
        \caption{CMP frame before compression}
    \end{subfigure}
    \begin{subfigure}[b]{0.3\textwidth}
        \includegraphics[width=\textwidth]{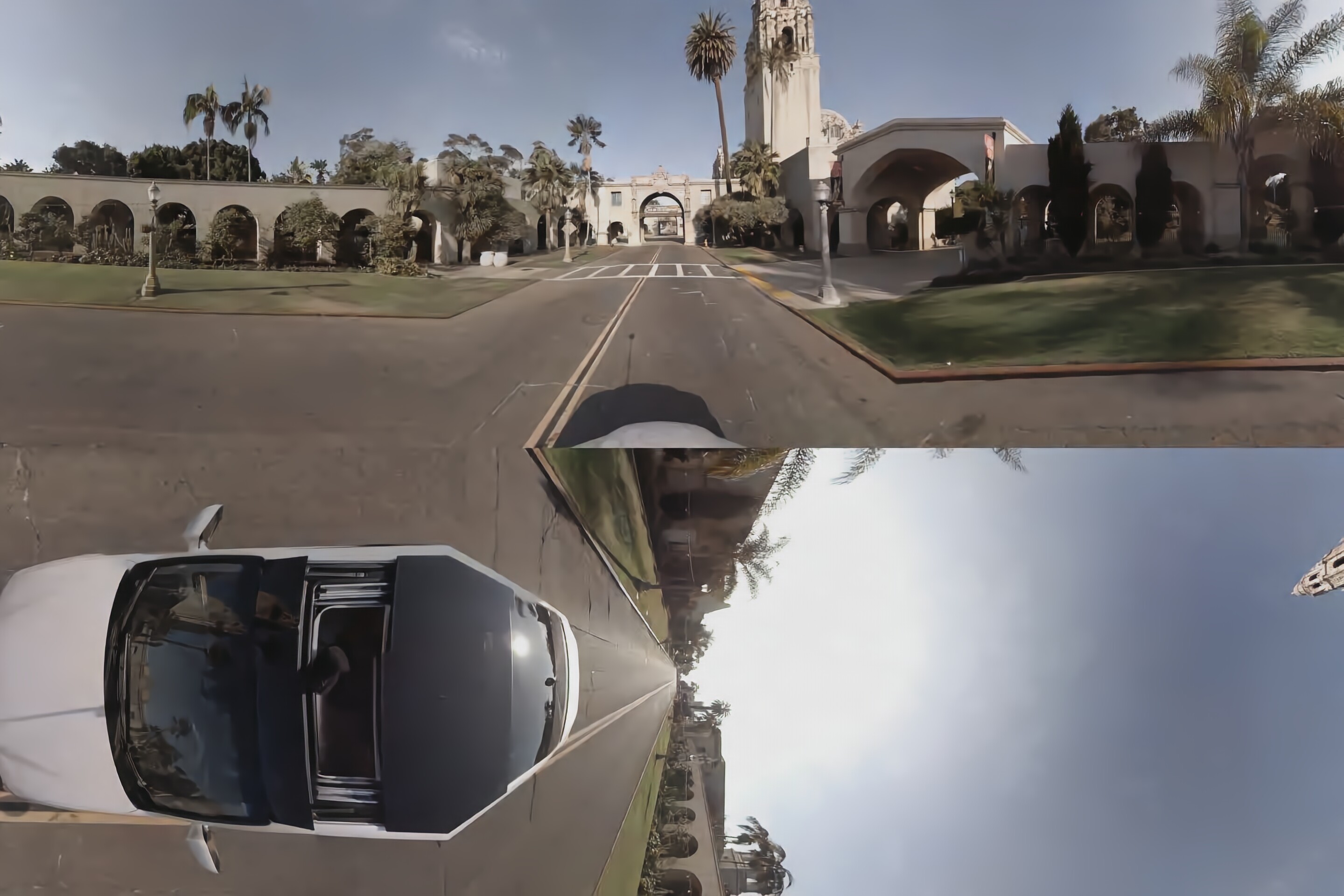}
        \caption{CMP frame after compression}
    \end{subfigure}
  
    \begin{subfigure}[b]{0.3\textwidth}
        \includegraphics[width=\textwidth]{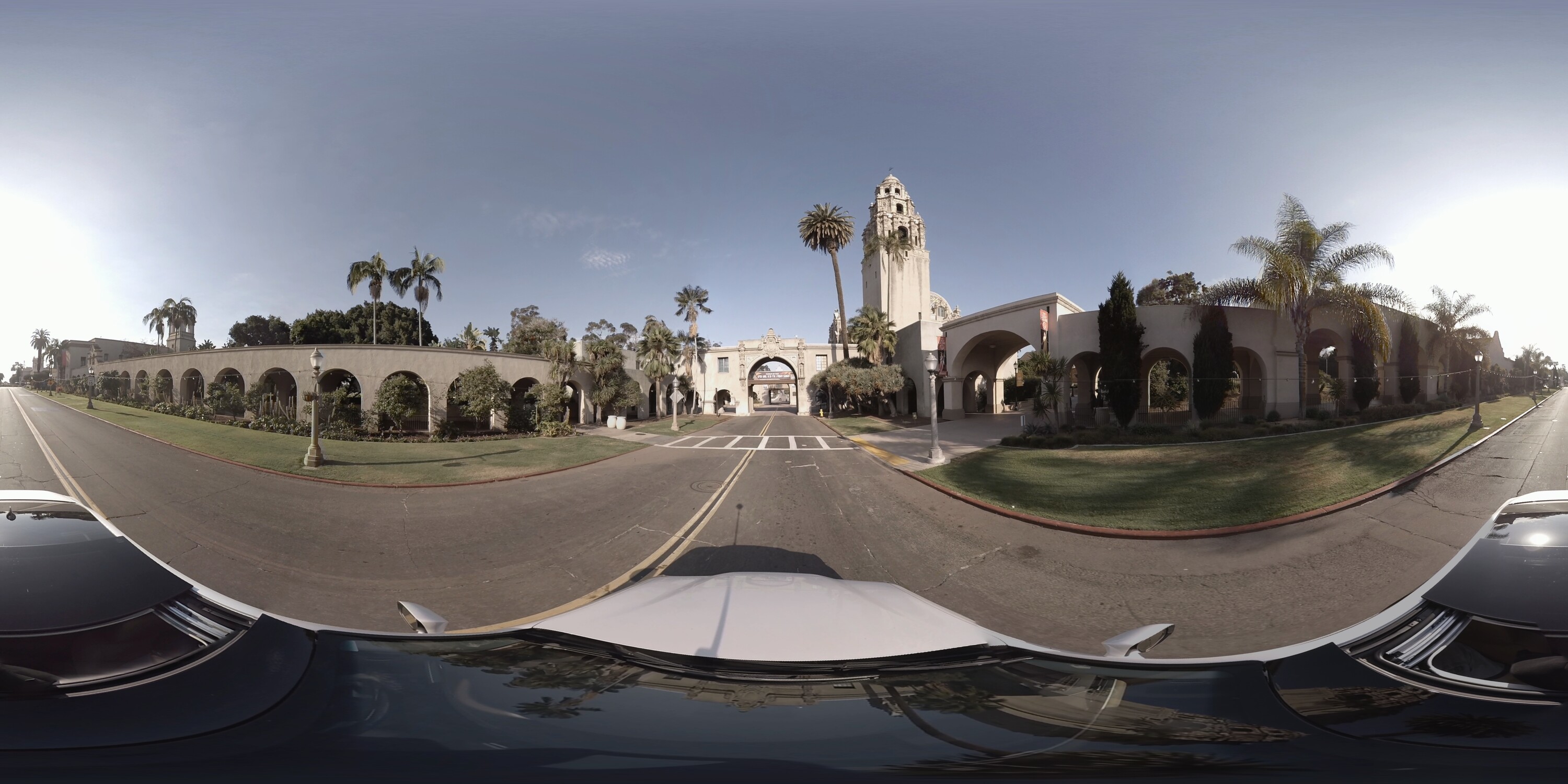}
        \caption{ERP frame before compression}
    \end{subfigure}
    \begin{subfigure}[b]{0.3\textwidth}
        \includegraphics[width=\textwidth]{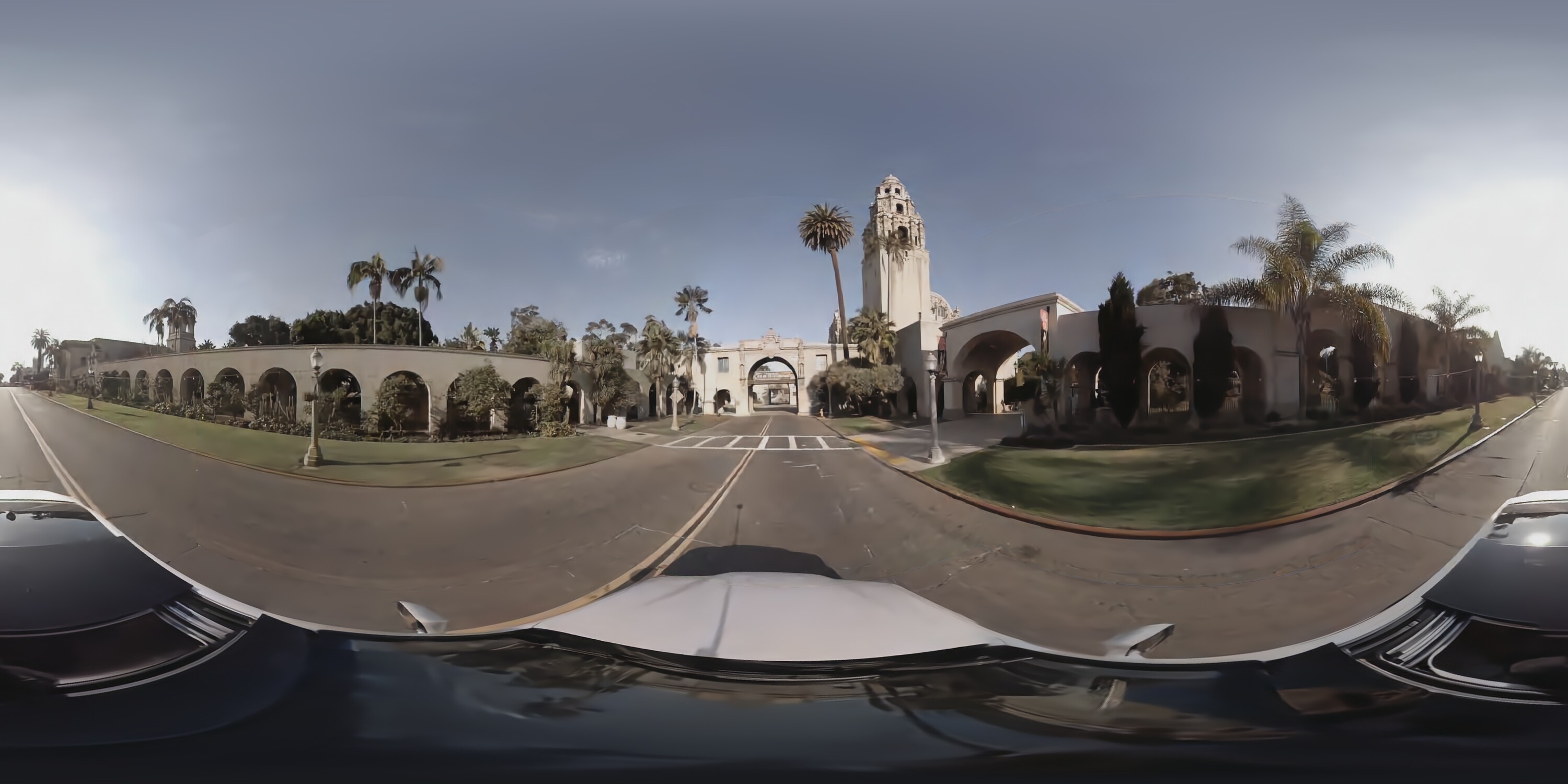}
        \caption{ERP frame after compression}
    \end{subfigure}

\caption{Projection layer output}
\label{fig:layerout}
\end{figure}

This model takes as input each of the ERP projected sequences represented in Table \ref{tab:sequences}. The reference ERP frame \ref{fig:layerout}(c) is then converted to CMP \ref{fig:layerout}(a) by applying the forward pass of the first layer of the model. For this conversion, the model changes the reference ERP frame resolution to the chosen $(2880\times 1920)$ CMP points. Next, the CMP frames are used as input for the second layer of the model, which is the compression layer \ref{fig:layerout}(b). To complete this end-to-end process, the model converts the compressed CMP frames back to the reference ERP projection \ref{fig:layerout}(d).

The output of each of these layers is shown in Figure \ref{fig:layerout}. It can be seen that the model effectively applies both projections (CMP-ERP) and compression.

The layers in the model convert projections using the nearest-neighbor interpolation method. In other words, the model finds the position of the coding projection in the reference projection by rounding to the nearest neighboring pixels. As this is the primary operation performed in each projection layer, there are no trainable parameters involved in this step. To enable learning and influence the weight update during backpropagation, the input and output of the projection layer have active gradients. The nearest neighbor interpolation in the forward pass of the projection layer is implemented using PyTorch, with active autograd. This means that the autograd functionality only considers this function when calculating gradients during the backward pass. As a result, the model can automatically backpropagate through the entire architecture without the need to write a separate backpropagation function. This approach ensures that the model learns effectively and can handle the projection conversion and compression processes efficiently.

To validate the layer backpropagation, each layer was evaluated separately to ensure that autograd is functioning as proven in the previous chapter. A 2D $3\times 3$ matrix was used as input to the projection layer

\begin{equation}
\text{layer input (x)} = \begin{pmatrix}
0 & 1 & 2 \\
3 & 4 & 5 \\
6 & 7 & 8 \\
\end{pmatrix}.
\end{equation}

The projection was specified to have the size of a 2D $2\times2$ matrix as follows:

\begin{equation}
\text{layer output (y)} = \begin{pmatrix}
8 & 5 \\
3 & 5 \\
\end{pmatrix}.
\end{equation}

The ground truth value of the layer was defined as an all-ones matrix with the same size as the coding projection $(2\times 2)$. In the first step, the matrices $x$ and $y$ are compared, and at positions where the coding projection is rounded or equal to the reference projection, the gradient is calculated. The first step involves calculating the loss between the projection output and the ground truth. The gradient matrix has the same size as the input and is all zero, except for the positions where the comparison is positive. The gradient of the loss with respect to the input matrix should be as follows

\begin{equation}
\text{Gradient of loss with respect to input }=
\begin{pmatrix}
0 & 0 & 0 \\
G_1 & 0 & G_2 \\
0 & 0 & G_3 \\
\end{pmatrix},
\end{equation}

where these $G$ values depend on the gradient of the loss with respect to the output, which is calculated by PyTorch automatically using the loss.backward() function.

The gradient of the loss with respect to the output for this projection is 

\begin{equation}
\frac{\partial L}{\partial y}=\begin{pmatrix}
\frac{\partial L}{\partial y_1} \\
\frac{\partial L}{\partial y_2} \\
\frac{\partial L}{\partial y_3} \\
\frac{\partial L}{\partial y_4} \\
\end{pmatrix} =
\begin{pmatrix}
3.4441 \\
3.2480 \\
1.1205 \\
1.6531 \\
\end{pmatrix}
\label{equ:gly}
\end{equation}
for the provided example.
The $G_1$, $G_2$, and $G_3$ values are calculated using dot products of the gradient of the loss with respect to the output and the weight vector which shows which values should influence the gradient this vector should be one if the output is round of the input and zero otherwise, as 

\begin{align}
G_1 &= [0, 0, 1, 0] \cdot \frac{\partial L}{\partial y}, \nonumber \\
G_2 &= [0, 1, 0, 1] \cdot \frac{\partial L}{\partial y}, \nonumber \\
G_3 &= [0, 0, 1, 0] \cdot \frac{\partial L}{\partial y}.
\end{align}
\clearpage

The resulting gradient of the loss with respect to the input calculated by pytorch is 

\begin{equation}
\begin{pmatrix}
0.0000 & 0.0000 & 0.0000 \\
1.1205 & 0.0000 & 4.9011 \\
0.0000 & 0.0000 & 3.4441 \\
\end{pmatrix}
\end{equation}
This output confirms that the autograd is working as mathematically stated in Chapter 4 for the model's background function.

\begin{figure}
    \centering
    \begin{subfigure}[b]{0.3\textwidth}
        \includegraphics[width=\textwidth]{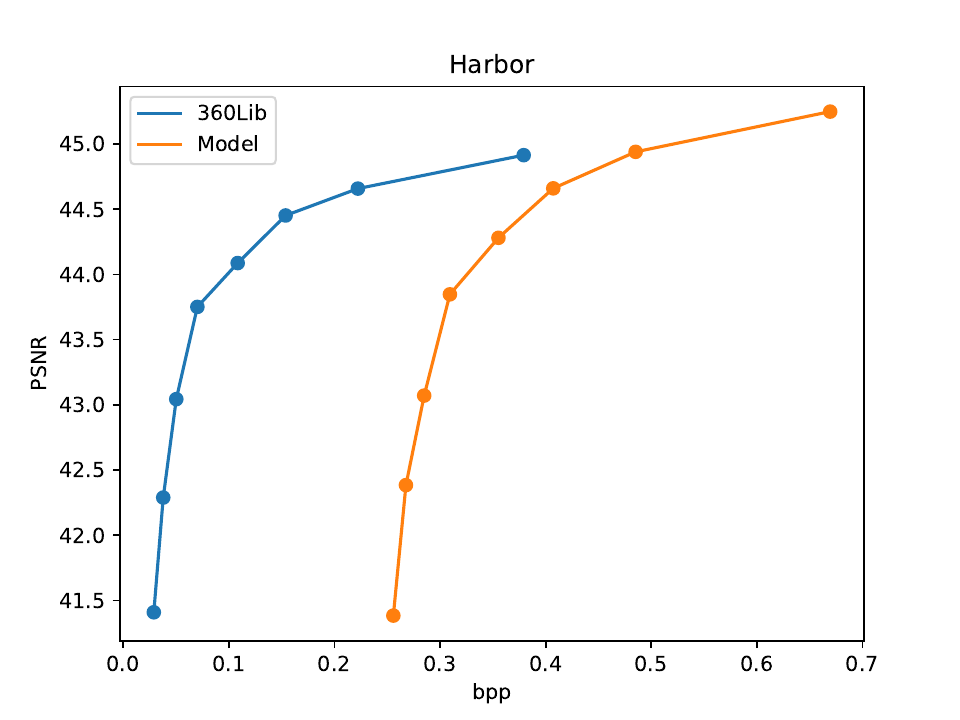}
        \caption{Harbor}
    \end{subfigure}
    \begin{subfigure}[b]{0.3\textwidth}
        \includegraphics[width=\textwidth]{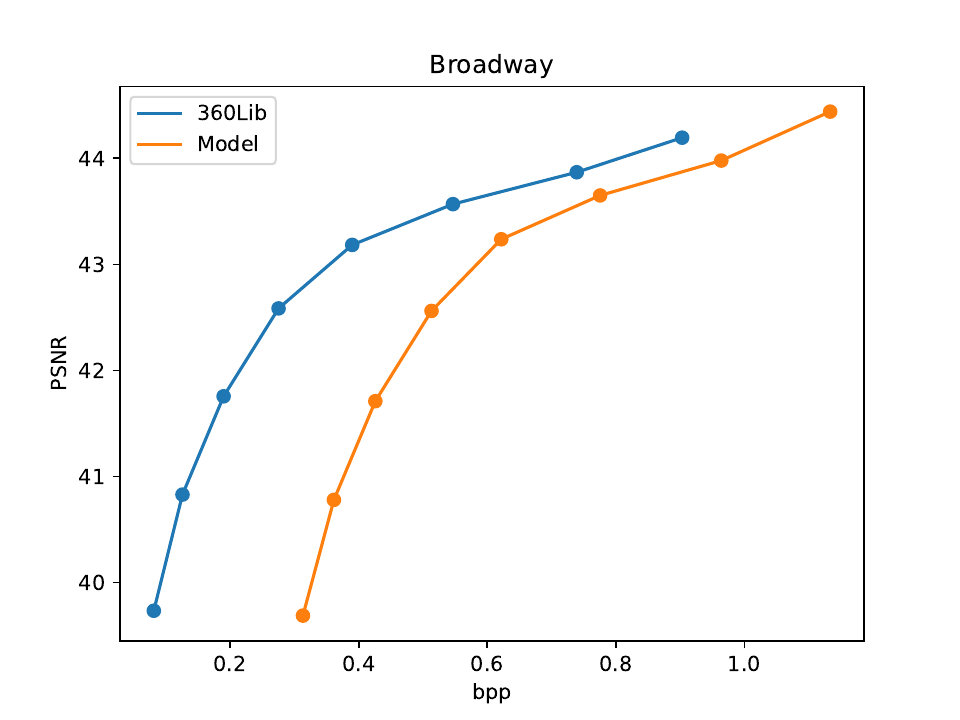}
        \caption{Broadway}
    \end{subfigure}
    \begin{subfigure}[b]{0.3\textwidth}
        \includegraphics[width=\textwidth]{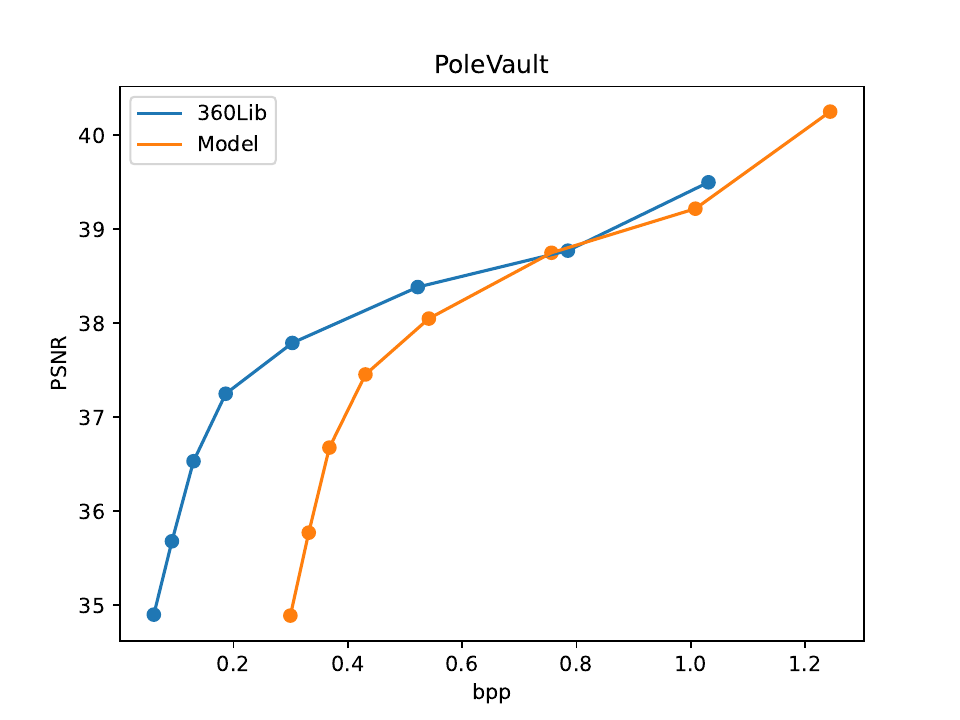}
        \caption{PoleVault}
    \end{subfigure}
    
    \begin{subfigure}[b]{0.3\textwidth}
        \includegraphics[width=\textwidth]{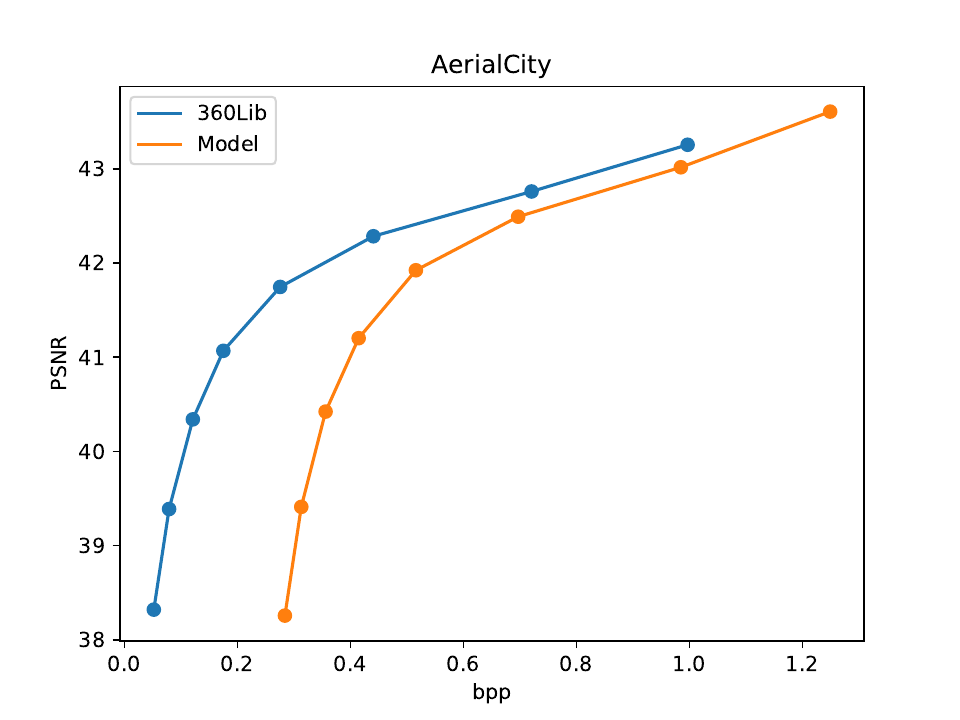}
        \caption{AerialCity}
    \end{subfigure}
    \begin{subfigure}[b]{0.3\textwidth}
        \includegraphics[width=\textwidth]{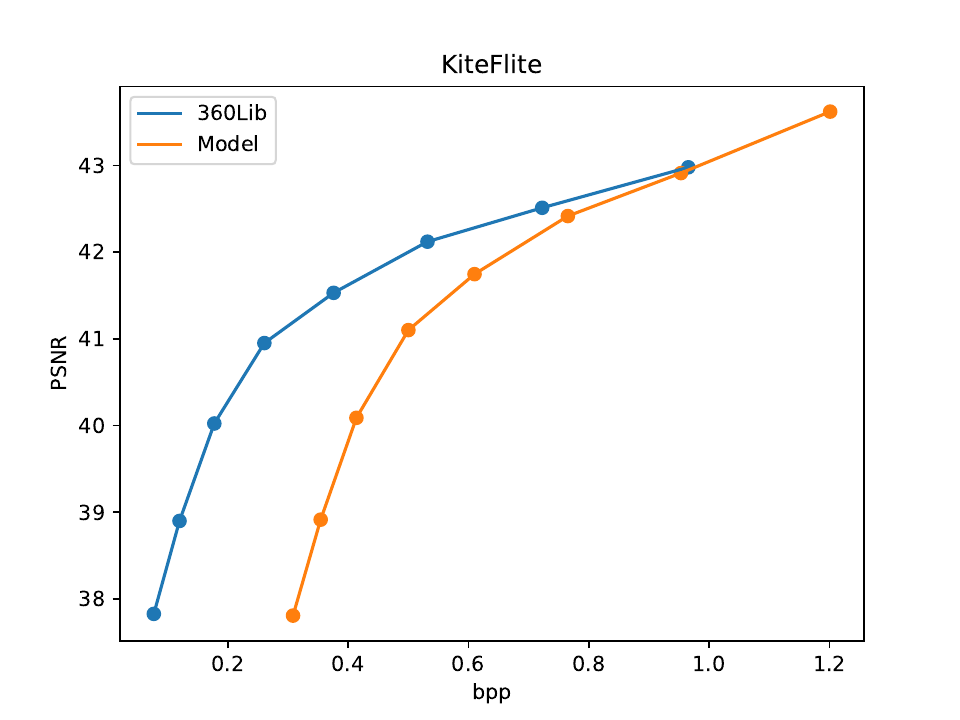}
        \caption{KiteFlite}
    \end{subfigure}
    \begin{subfigure}[b]{0.3\textwidth}
        \includegraphics[width=\textwidth]{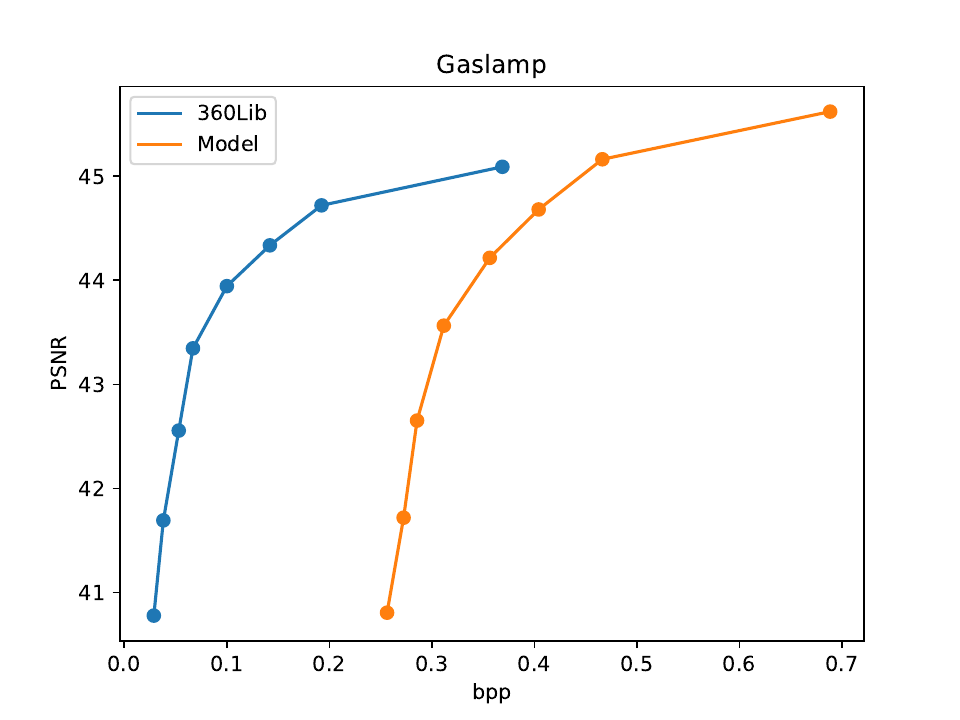}
        \caption{Gaslamp}
    \end{subfigure}
    
    \begin{subfigure}[b]{0.3\textwidth}
        \includegraphics[width=\textwidth]{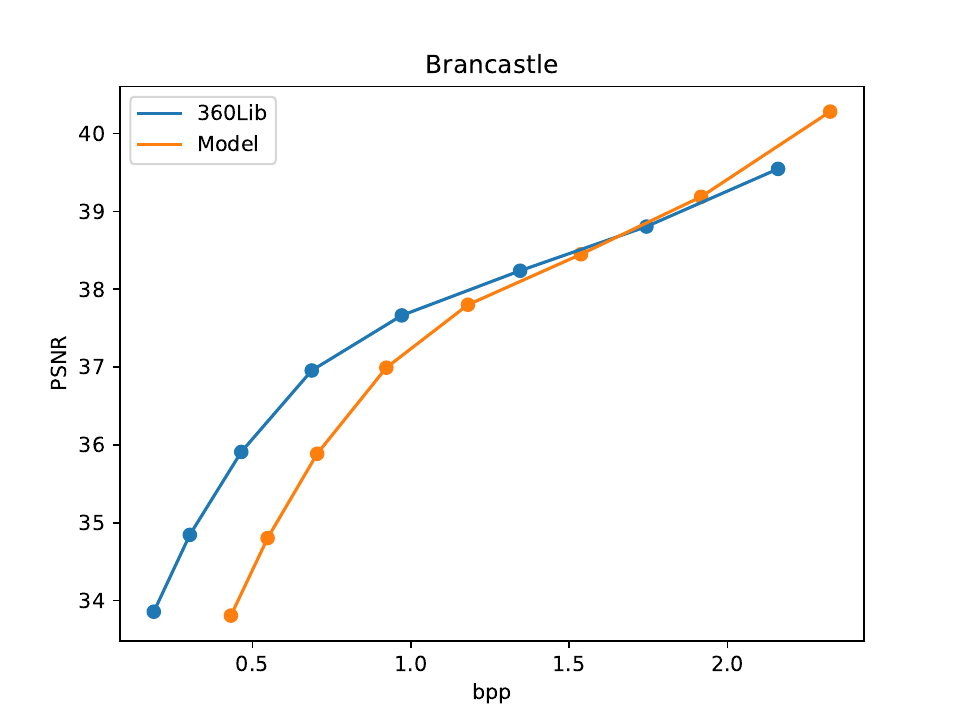}
        \caption{Brancastle}
    \end{subfigure}
    
    \caption{Method comparison using rate-distortion (PSNR) curves.}
    \label{fig:methods-compare}
\end{figure}

\clearpage

\section{Methods comparison}
\begin{figure}
    \centering
    \begin{subfigure}[b]{0.3\textwidth}
        \includegraphics[width=\textwidth]{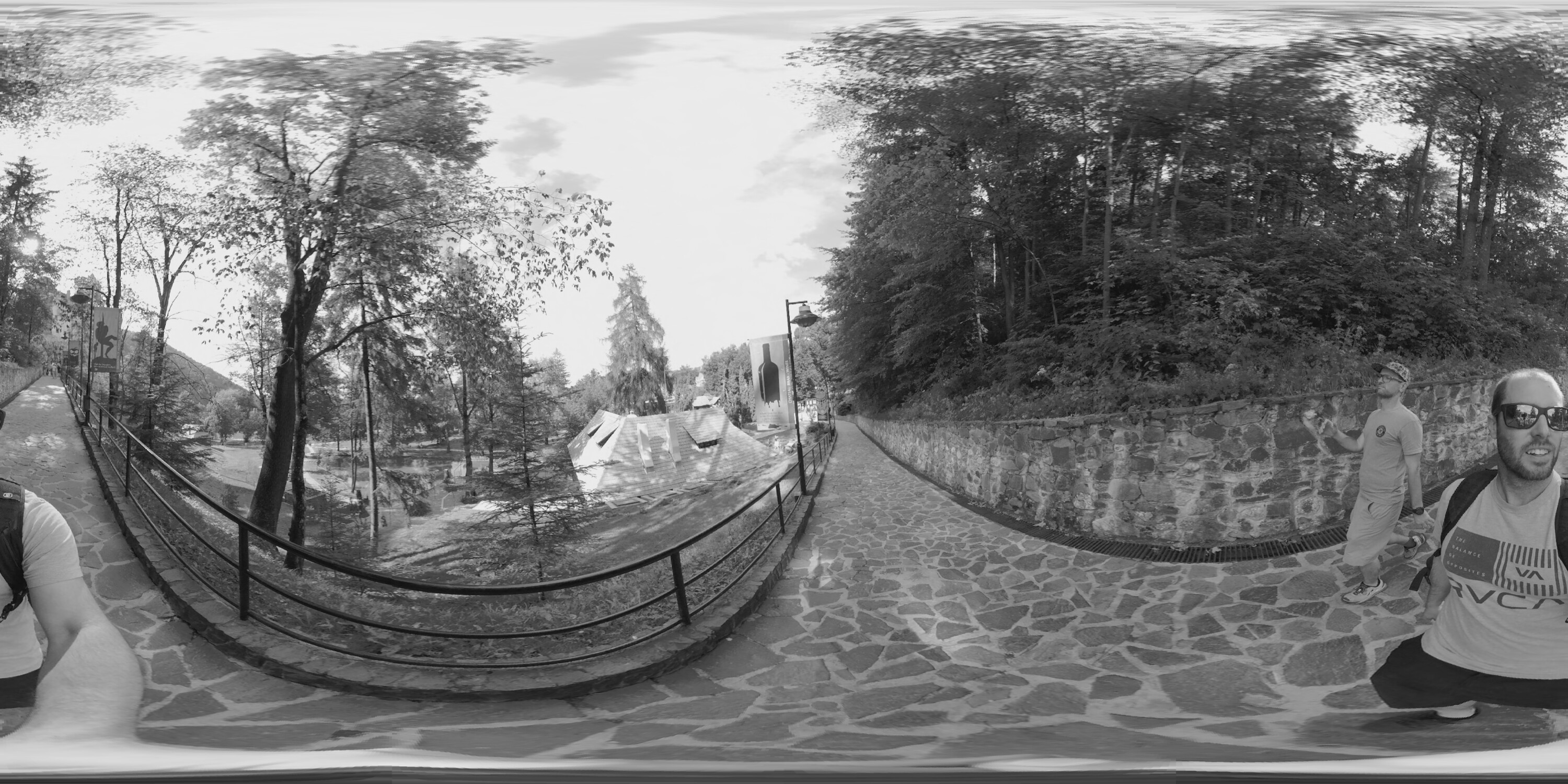}
        \caption{360-lib ERP projection output}
    \end{subfigure}
    \begin{subfigure}[b]{0.3\textwidth}
        \includegraphics[width=\textwidth]{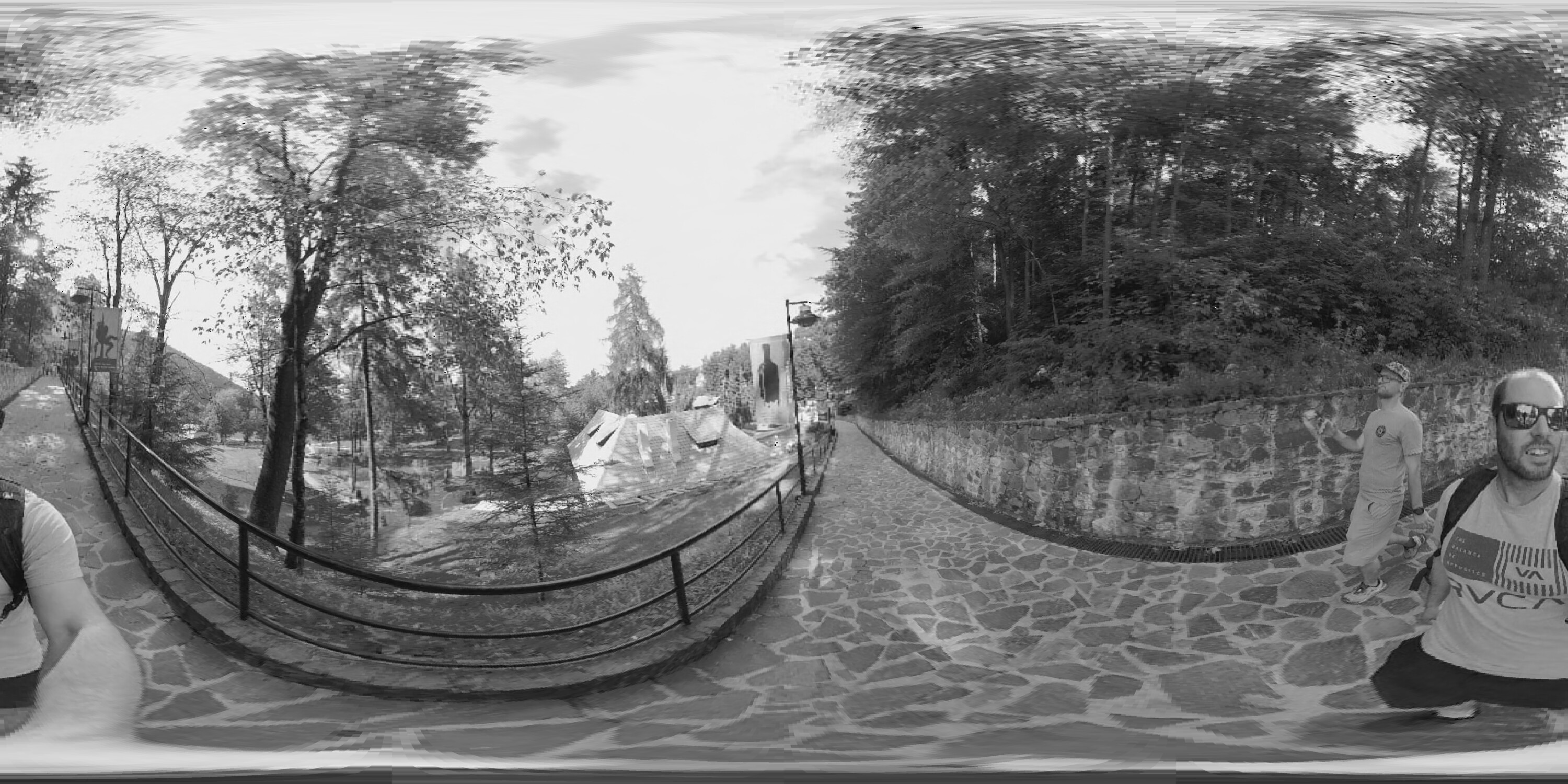}
        \caption{ERP projection layer output}
    \end{subfigure}
    \begin{subfigure}[b]{0.3\textwidth}
        \includegraphics[width=\textwidth]{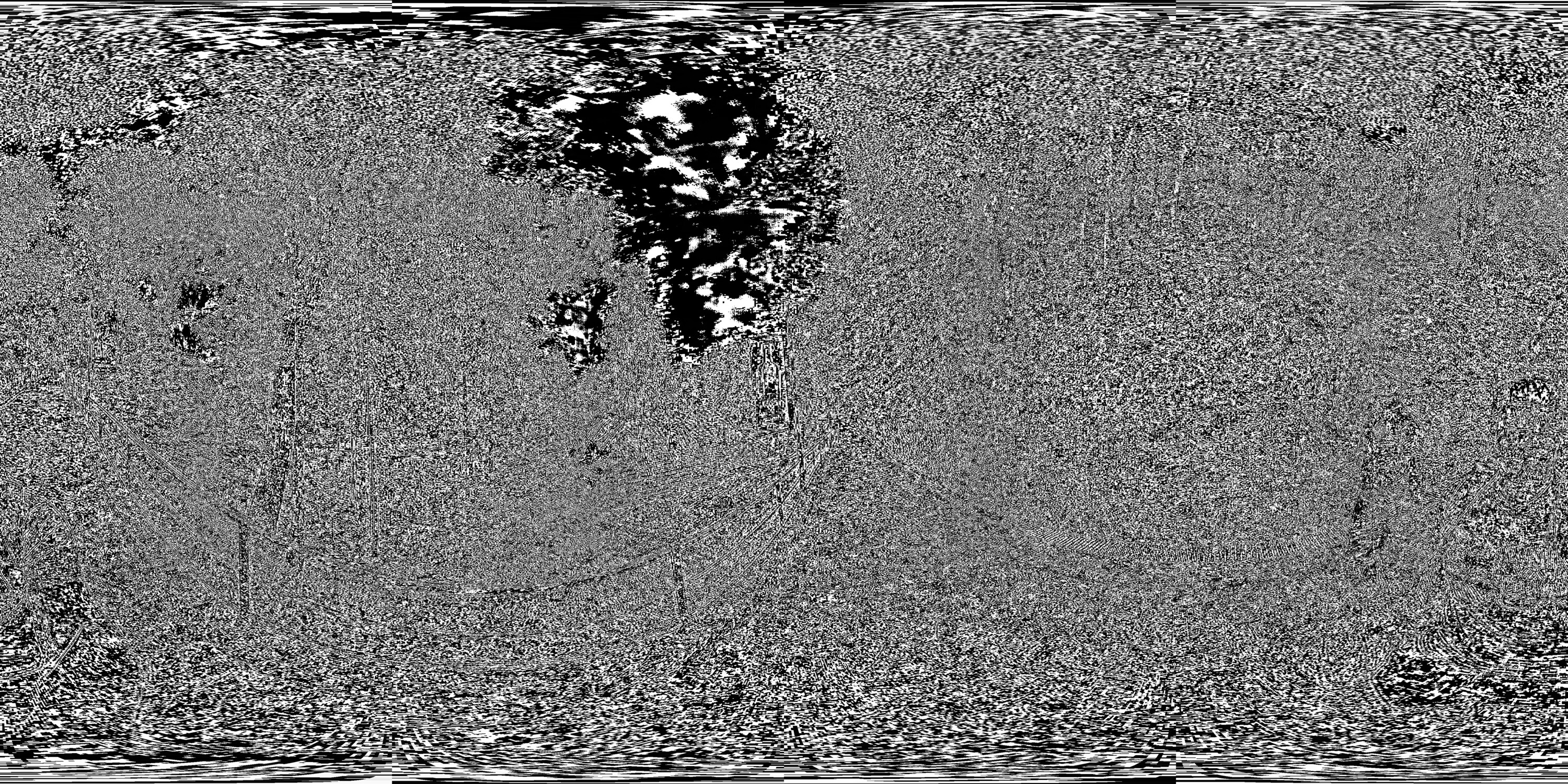}
        \caption{Difference between two methods}
    \end{subfigure}
    \caption{ERP projection comparison for both methods }
    \label{fig:methods-diff}
\end{figure}

So far, two separate methods for projecting and compressing a frame have been conducted. It is now time to compare the results of these two methods. However, it should be kept in mind that, for the model with projection layers only, only CMP projection is currently supported. Therefore, the comparison between the two methods will focus on this projection for different sequences. For each sequence, the rate-distortion curve from the 360lib method, based on nearest neighbor and the model output for 16 frames, is compared and depicted in Figure \ref{fig:methods-compare}. As can be seen, the rate-distortion curves of the 360lib-based projection method perform better for all sequences. It is worth noting that the curves have a similar shape, but the end-to-end PSNR value calculated is lower for the model-based projection method. To better understand the difference between these two curves, the Bjøntegaard-Delta was used to calculate the bd\_rate between each pair of curves, and the difference is shown in Table \ref{table:projection_comparison}. To investigate the cause of the gap between projection layer outputs and 360lib projection outputs, tests were conducted. The output of both methods was compared and their differences are shown in Figure \ref{fig:methods-diff}. Figure \ref{fig:methods-diff} (a, b) shows that the outputs of both methods are similar in appearance with no visible artifacts. However, Figure \ref{fig:methods-diff} (c) illustrates that the cause of the error is uniformly distributed across the projections. The main reason for this difference is due to a discrepancy in the methods used for converting projections, both from ERP to CMP and vice versa. Additionally, when calculating the PSNR, the function developed for the 360lib software is used, which employs the reference frame in ERP format originally generated by 360lib. These factors may also contribute to the observed gap between the two methods.

\begin{table}[t]
\centering
\caption{Method comparison based on bd\_rate}
\begin{tabular}{|c|c|c|c|c|c|c|}
\hline
Harbor  & Broadway   & PoleVault  & AerialCity  & KiteFlite  & Gaslamp & Brancastle \\
\hline
369.74  & 114.10   & 105.15  & 142.86  & 103.49  & 349.59 & 37.65 \\
\hline
\end{tabular}
\label{tab:method-compare}
\end{table}

    \chapter{Conclusion}
\label{chap: conclusion}

In conclusion, this thesis has studied deep-learning-based end-to-end optimized video compression networks applied to encoding and decoding projected 360-degree video frames. The research aimed to assess the performance of these networks in the context of projection functions for 360-degree videos by employing the scale-space flow network by following the JVET common test conditions and evaluation procedures.

A pipeline was established to evaluate different projections utilizing 360-degree testing sequences provided by JVET. Objective metrics calculated by the 360Lib software were used to measure the effectiveness of the proposed compression technique. Additionally, an innovative three-layer model with trainable reprojection layers in Python was introduced to address the unique challenges of 360-degree video compression, which was developed alongside the existing 360Lib software. This method consisted of a first layer transforming the reference projection into the coding projection, a middle layer compressing the coding projection, and a final layer converting the coding projection back to the reference projection. Notably, the projection conversion layers were solely responsible for pre- and post-processing of the input and output of the compression layer, without containing learning parameters.

The study also explored the influence of different projections on the compression performance of 360-degree videos using both learning-based compression models and traditional compression techniques, such as HM-16.16. Results showed that, in traditional video codecs, CMP outperformed ERP-based projections, whereas in learning-based compression model method, the performance of ERP-based projections surpassed that of CMP-based projections. This finding underscores the fundamental differences between the approaches to projection in traditional and learning-based video codecs.

The varying performance of different projections across codecs can be attributed to their distinct architectures. Hybrid video codecs are better suited for CMP-based projections due to their block-based approach, which is compatible with the multiple faces of CMP, leading to improved compression performance and overall better coding efficiency. Conversely, learning-based compression models like scale-space flow exhibit superior performance with single-face projection formats, such as ERP, resulting in smoother motion vectors and increased compression efficiency.

These results highlight that learning-based models possess unique advantages in video compression, necessitating further research into their potential uses and enhancements. As technology improves and new compression methods are developed, understanding the functioning of different techniques in various codecs becomes increasingly crucial. Moreover, training learning-based models with data from multiple perspectives can enhance their performance when handling formats like CMP, enabling them to tackle changes in motion calculations and devise more effective compression strategies for multi-angle projections

Furthermore, a comparison between two separate methods for projecting and compressing a frame was conducted, focusing on the CMP projection for different sequences. The results showed that the 360lib-based projection method outperformed the model-based projection method, with a performance gap observed in the end-to-end PSNR value calculated. The discrepancy between the methods used for converting projections, both from ERP to CMP and vice versa, was identified as the main reason for this difference.

In summary, this study has provided valuable insights into the performance of different projections in traditional and learning-based video codecs, paving the way for future research and development in the field of 360-degree video compression.

    \appendix

    \backmatter
    \listoffigures
    \listoftables
    \bibliographystyle{ieeetr}
    \bibliography{bibliography/bibliography}

@inproceedings{sreedhar2016viewport,
  title={Viewport-adaptive encoding and streaming of 360-degree video for virtual reality applications},
  author={Sreedhar, Kashyap Kammachi and Aminlou, Alireza and Hannuksela, Miska M and Gabbouj, Moncef},
  booktitle={Proc. IEEE International Symposium on Multimedia (ISM)},
  pages={583--586},
  year={2016},
  organization={IEEE}
}

@inproceedings{chang2019argoverse,
  title={Argoverse: 3d tracking and forecasting with rich maps},
  author={Chang, Ming-Fang and Lambert, John and Sangkloy, Patsorn and Singh, Jagjeet and Bak, Slawomir and Hartnett, Andrew and Wang, De and Carr, Peter and Lucey, Simon and Ramanan, Deva and others},
  booktitle={Proc. IEEE/CVF Conference on Computer Vision and Pattern Recognition},
  pages={8748--8757},
  year={2019}
}

@inproceedings{feurstein2018towards,
  title={Towards an integration of 360-degree video in higher education},
  author={Feurstein, Michael Sebastian},
  booktitle={Proc. DeLFI Workshops 2018 co-located with 16th e-Learning Conference of the German Computer Society (DeLFI 2018)},
  pages={1--12},
  year={2018},
  organization={CEUR WS}
}

@article{huang2019utility,
  title={Utility-oriented resource allocation for 360-degree video transmission over heterogeneous networks},
  author={Huang, Wei and Ding, Lianghui and Zhai, Guangtao and Min, Xiongkuo and Hwang, Jenq-Neng and Xu, Yiling and Zhang, Wenjun},
  journal={Digital Signal Processing},
  volume={84},
  pages={1--14},
  year={2019},
  publisher={Elsevier}
}

@article{he2017360lib,
  title={360Lib software manual},
  author={He, Yuwen and Xiu, Xiaoyu and Ye, Yan and Zakharchenko, Vladyslav and Alshina, Elena and Dsouza, Amith and Lin, Jian-Liang and Chang, Shen-Kai and Huang, Chao-Chih and Sun, Yule and others},
  journal={Joint Video Exploration Team (JVET) of ITU-T SG},
  volume={16},
  year={2017}
}

@article{boyce2017jvet,
  title={JVET common test conditions and evaluation procedures for 360 video},
  author={Boyce, Jill and Alshina, Elena and Abbas, Adeel and Ye, Yan},
  journal={Joint Video Exploration Team of ITU-T SG},
  volume={16},
  pages={1--8},
  year={2017}
}

@inproceedings{agustsson2020scale,
  title={Scale-space flow for end-to-end optimized video compression},
  author={Agustsson, Eirikur and Minnen, David and Johnston, Nick and Balle, Johannes and Hwang, Sung Jin and Toderici, George},
  booktitle={Proc. IEEE/CVF Conference on Computer Vision and Pattern Recognition},
  pages={8503--8512},
  year={2020}
}

@inproceedings{pessoa2020end,
  title={End-to-end learning of video compression using spatio-temporal autoencoders},
  author={Pessoa, Jorge and Aidos, Helena and Tom{\'a}s, Pedro and Figueiredo, M{\'a}rio AT},
  booktitle={Proc. IEEE Workshop on Signal Processing Systems (SiPS)},
  pages={1--6},
  year={2020},
  organization={IEEE}
}

@inproceedings{xiu2017evaluation,
  title={An evaluation framework for 360-degree video compression},
  author={Xiu, Xiaoyu and He, Yuwen and Ye, Yan and Vishwanath, Bharath},
  booktitle={Proc. IEEE Visual Communications and Image Processing (VCIP)},
  pages={1--4},
  year={2017},
  organization={IEEE}
}

@article{hussain2021evaluation,
  title={Evaluation of 360° Image Projection Formats; Comparing Format Conversion Distortion Using Objective Quality Metrics},
  author={Hussain, Ikram and Kwon, Oh-Jin},
  journal={Journal of Imaging},
  volume={7},
  number={8},
  pages={137},
  year={2021},
  publisher={MDPI}
}

@article{theis2017lossy,
  title={Lossy image compression with compressive autoencoders},
  author={Theis, Lucas and Shi, Wenzhe and Cunningham, Andrew and Husz{\'a}r, Ferenc},
  year={2017}
}

@article{hinton2006reducing,
  title={Reducing the dimensionality of data with neural networks},
  author={Hinton, Geoffrey E and Salakhutdinov, Ruslan R},
  journal={science},
  volume={313},
  number={5786},
  pages={504--507},
  year={2006},
  publisher={American Association for the Advancement of Science}
}

@inproceedings{dumas2018autoencoder,
  title={Autoencoder based image compression: can the learning be quantization independent?},
  author={Dumas, Thierry and Roumy, Aline and Guillemot, Christine},
  booktitle={Proc. IEEE International Conference on Acoustics, Speech and Signal Processing (ICASSP)},
  pages={1188--1192},
  year={2018},
  organization={IEEE}
}

@inproceedings{
balle2016end,
title={End-to-end Optimized Image Compression},
author={Johannes Ball{\'e} and Valero Laparra and Eero P. Simoncelli},
booktitle={Proc. International Conference on Learning Representations},
year={2017},
}

@inproceedings{wu2018video,
  title={Video compression through image interpolation},
  author={Wu, Chao-Yuan and Singhal, Nayan and Krahenbuhl, Philipp},
  booktitle={Proc. European Conference on Computer Vision (ECCV)},
  pages={416--431},
  year={2018}
}

@inproceedings{niklaus2017video,
  title={Video frame interpolation via adaptive separable convolution},
  author={Niklaus, Simon and Mai, Long and Liu, Feng},
  booktitle={Proc. IEEE International Conference on Computer Vision},
  pages={261--270},
  year={2017}
}

@article{baker2011database,
  title={A database and evaluation methodology for optical flow},
  author={Baker, Simon and Scharstein, Daniel and Lewis, JP and Roth, Stefan and Black, Michael J and Szeliski, Richard},
  journal={International journal of computer vision},
  volume={92},
  pages={1--31},
  year={2011},
  publisher={Springer}
}

@inproceedings{werlberger2011optical,
  title={Optical flow guided TV-L 1 video interpolation and restoration},
  author={Werlberger, Manuel and Pock, Thomas and Unger, Markus and Bischof, Horst},
  booktitle={Proc. Energy Minimization Methods in Computer Vision and Pattern Recognition},
  pages={273--286},
  year={2011},
  organization={Springer}
}

@article{wadhwa2013phase,
  title={Phase-based video motion processing},
  author={Wadhwa, Neal and Rubinstein, Michael and Durand, Fr{\'e}do and Freeman, William T},
  journal={ACM Transactions on Graphics (TOG)},
  volume={32},
  number={4},
  pages={1--10},
  year={2013},
  publisher={ACM New York, NY, USA}
}

@article{didyk2013joint,
  title={Joint view expansion and filtering for automultiscopic 3D displays},
  author={Didyk, Piotr and Sitthi-Amorn, Pitchaya and Freeman, William and Durand, Fr{\'e}do and Matusik, Wojciech},
  journal={ACM Transactions on Graphics (TOG)},
  volume={32},
  number={6},
  pages={1--8},
  year={2013},
  publisher={ACM New York, NY, USA}
}

@inproceedings{meyer2015phase,
  title={Phase-based frame interpolation for video},
  author={Meyer, Simone and Wang, Oliver and Zimmer, Henning and Grosse, Max and Sorkine-Hornung, Alexander},
  booktitle={Proc. IEEE Conference on Computer Vision and Pattern Recognition},
  pages={1410--1418},
  year={2015}
}

@inproceedings{lu2019dvc,
  title={Dvc: An end-to-end deep video compression framework},
  author={Lu, Guo and Ouyang, Wanli and Xu, Dong and Zhang, Xiaoyun and Cai, Chunlei and Gao, Zhiyong},
  booktitle={Proc. IEEE/CVF Conference on Computer Vision and Pattern Recognition},
  pages={11006--11015},
  year={2019}
}

@article{barron1994performance,
  title={Performance of optical flow techniques},
  author={Barron, John L and Fleet, David J and Beauchemin, Steven S},
  journal={International Journal of Computer Vision},
  volume={12},
  pages={43--77},
  year={1994},
  publisher={Kluwer Academic Publishers}
}

@inproceedings{gallego2018unifying,
  title={A unifying contrast maximization framework for event cameras, with applications to motion, depth, and optical flow estimation},
  author={Gallego, Guillermo and Rebecq, Henri and Scaramuzza, Davide},
  booktitle={Proc. IEEE Conference on Computer Vision and Pattern Recognition},
  pages={3867--3876},
  year={2018}
}

@inproceedings{agustsson2019generative,
  title={Generative adversarial networks for extreme learned image compression},
  author={Agustsson, Eirikur and Tschannen, Michael and Mentzer, Fabian and Timofte, Radu and Gool, Luc Van},
  booktitle={Proc. IEEE/CVF International Conference on Computer Vision},
  pages={221--231},
  year={2019}
}

@inproceedings{dosovitskiy2015flownet,
  title={Flownet: Learning optical flow with convolutional networks},
  author={Dosovitskiy, Alexey and Fischer, Philipp and Ilg, Eddy and Hausser, Philip and Hazirbas, Caner and Golkov, Vladimir and Van Der Smagt, Patrick and Cremers, Daniel and Brox, Thomas},
  booktitle={Proc. IEEE International Conference on Computer Vision},
  pages={2758--2766},
  year={2015}
}

@article{ye2017jvet,
  title={JVET-E1003: Algorithm descriptions of projection format conversion and video quality metrics in 360Lib},
  author={Ye, Y and Alshima, E and Boyce, J},
  journal={Joint Video Exploration Team (JVET) of ITU-T SG},
  volume={16},
  year={2017}
}

@inproceedings{podborski2017virtual,
  title={Virtual reality and DASH},
  author={Podborski, D and Thomas, E and Hannuksela, MM and Oh, S and Stockhammer, T and Pham, S},
  booktitle={International broadcasting convention, Ibc},
  year={2017}
}

@inproceedings{djelouah2019neural,
  title={Neural inter-frame compression for video coding},
  author={Djelouah, Abdelaziz and Campos, Joaquim and Schaub-Meyer, Simone and Schroers, Christopher},
  booktitle={Proc. IEEE/CVF International Conference on Computer Vision},
  pages={6421--6429},
  year={2019}
}

@inproceedings{
balle2018variational,
title={Variational image compression with a scale hyperprior},
author={Johannes Ballé and David Minnen and Saurabh Singh and Sung Jin Hwang and Nick Johnston},
booktitle={Proc. International Conference on Learning Representations},
year={2018},
}

@article{minnen2018joint,
  title={Joint autoregressive and hierarchical priors for learned image compression},
  author={Minnen, David and Ball{\'e}, Johannes and Toderici, George D},
  journal={Advances in Neural Information Processing Systems},
  volume={31},
  year={2018}
}

@inproceedings{habibian2019video,
  title={Video compression with rate-distortion autoencoders},
  author={Habibian, Amirhossein and Rozendaal, Ties van and Tomczak, Jakub M and Cohen, Taco S},
  booktitle={Proc. IEEE/CVF International Conference on Computer Vision},
  pages={7033--7042},
  year={2019}
}

@article{wiegand2003overview,
  title={Overview of the H. 264/AVC video coding standard},
  author={Wiegand, Thomas and Sullivan, Gary J and Bjontegaard, Gisle and Luthra, Ajay},
  journal={IEEE Transactions on Circuits and Systems for Video Technology},
  volume={13},
  number={7},
  pages={560--576},
  year={2003},
  publisher={IEEE}
}

@article{sullivan2012overview,
  title={Overview of the high efficiency video coding (HEVC) standard},
  author={Sullivan, Gary J and Ohm, Jens-Rainer and Han, Woo-Jin and Wiegand, Thomas},
  journal={IEEE Transactions on Circuits and Systems for Video Technology},
  volume={22},
  number={12},
  pages={1649--1668},
  year={2012},
  publisher={IEEE}
}

@article{alkhayrat2020comparative,
  title={A comparative dimensionality reduction study in telecom customer segmentation using deep learning and PCA},
  author={Alkhayrat, Maha and Aljnidi, Mohamad and Aljoumaa, Kadan},
  journal={Journal of Big Data},
  volume={7},
  pages={1--23},
  year={2020},
  publisher={Springer}
}

@book{goodfellow2016deep,
  title={Deep learning},
  author={Goodfellow, Ian and Bengio, Yoshua and Courville, Aaron},
  year={2016},
  publisher={MIT press}
}

@article{li2020anomaly,
  title={Anomaly detection of time series with smoothness-inducing sequential variational auto-encoder},
  author={Li, Longyuan and Yan, Junchi and Wang, Haiyang and Jin, Yaohui},
  journal={IEEE Transactions on Neural Networks and Learning Systems},
  volume={32},
  number={3},
  pages={1177--1191},
  year={2020},
  publisher={IEEE}
}

@inproceedings{sun2021image,
  title={Image compression algorithm based on variational autoencoder},
  author={Sun, Ying and Li, Lang and Ding, Yang and Bai, Jiabao and Xin, Xiangning},
  booktitle={Journal of Physics: Conference Series},
  volume={2066},
  number={1},
  pages={012008},
  year={2021},
  organization={IOP Publishing}
}

@article{kingma2013auto,
  title={Auto-encoding variational bayes},
  author={Kingma, Diederik P and Welling, Max},
  journal={arXiv preprint arXiv:1312.6114},
  year={2013}
}

@inproceedings{rezende2014stochastic,
  title={Stochastic backpropagation and approximate inference in deep generative models},
  author={Rezende, Danilo Jimenez and Mohamed, Shakir and Wierstra, Daan},
  booktitle={International Conference on Machine Learning},
  pages={1278--1286},
  year={2014},
  organization={PMLR}
}

@inproceedings{liu2017video,
  title={Video frame synthesis using deep voxel flow},
  author={Liu, Ziwei and Yeh, Raymond A and Tang, Xiaoou and Liu, Yiming and Agarwala, Aseem},
  booktitle={Proc. IEEE International Conference on Computer Vision},
  pages={4463--4471},
  year={2017}
}

@inproceedings{long2016learning,
author = {Long, Gucan and Kneip, Laurent and Alvarez, Jose M. and li, Hongdong and Zhang, Xiaohu and Yu, Qifeng},
pages = {434-450},
title = {Proc. Learning Image Matching by Simply Watching Video},
volume = {9910},
year = {2016}
}

@article{geiger2013vision,
  title={Vision meets robotics: The kitti dataset},
  author={Geiger, Andreas and Lenz, Philip and Stiller, Christoph and Urtasun, Raquel},
  journal={The International Journal of Robotics Research},
  volume={32},
  number={11},
  pages={1231--1237},
  year={2013},
  publisher={Sage Publications Sage UK: London, England}
}

@article{sun2014quantitative,
  title={A quantitative analysis of current practices in optical flow estimation and the principles behind them},
  author={Sun, Deqing and Roth, Stefan and Black, Michael J},
  journal={International Journal of Computer Vision},
  volume={106},
  pages={115--137},
  year={2014},
  publisher={Springer}
}

@article{ye2019omnidirectional,
  title={Omnidirectional 360° video coding technology in responses to the joint call for proposals on video compression with capability beyond HEVC},
  author={Ye, Yan and Boyce, Jill M and Hanhart, Philippe},
  journal={IEEE Transactions on Circuits and Systems for Video Technology},
  volume={30},
  number={5},
  pages={1241--1252},
  year={2019},
  publisher={IEEE}
}

@techreport{boyce2017ee4,
  author = {Boyce, J. and Deng, Z.},
  title = {{EE4: Padded ERP (PERP) Projection Format}},
  institution = {Joint Video Exploration Team (JVET)},
  number = {JVET-G0098},
  month = {Jul},
  year = {2017}
}

@article{zhou2017ahg8,
  title={AHG8: A study on Equi-Angular Cubemap projection (EAC)},
  author={Zhou, Minhua},
  journal={Proc. Joint Video Exploration Team (JVET) of ITU-T SG},
  volume={16},
  year={2017}
}

@article{coban2017ahg8,
  title={AHG8: Adjusted cubemap projection for 360-degree video},
  author={Coban, M and Van der Auwera, G and Karczewicz, M},
  journal={Joint Video Exploration Team of ITU-T SG16 WP3 and ISO/IEC JTC1/SC29/WG11, JVET-F0025},
  year={2017}
}

@inproceedings{duanmu2018hybrid,
  title={Hybrid cubemap projection format for 360-degree video coding},
  author={Duanmu, Fanyi and He, Yuwen and Xiu, Xiaoyu and Hanhart, Philippe and Ye, Yan and Wang, Yao},
  booktitle={Proc. Data Compression Conference},
  pages={404--404},
  year={2018},
  organization={IEEE}
}

@article{lin2019efficient,
  title={Efficient projection and coding tools for 360° video},
  author={Lin, Jian-Liang and Lee, Ya-Hsuan and Shih, Cheng-Hsuan and Lin, Sheng-Yen and Lin, Hung-Chih and Chang, Shen-Kai and Wang, Peng and Liu, Lin and Ju, Chi-Cheng},
  journal={Proc. IEEE Journal on Emerging and Selected Topics in Circuits and Systems},
  volume={9},
  number={1},
  pages={84--97},
  year={2019},
  publisher={IEEE}
}

@article{ma2019image,
  title={Image and video compression with neural networks: A review},
  author={Ma, Siwei and Zhang, Xinfeng and Jia, Chuanmin and Zhao, Zhenghui and Wang, Shiqi and Wang, Shanshe},
  journal={IEEE Transactions on Circuits and Systems for Video Technology},
  volume={30},
  number={6},
  pages={1683--1698},
  year={2019},
  publisher={IEEE}
}

@article{sadeeq2021image,
  title={Image compression using neural networks: a review},
  author={Sadeeq, Haval T and Hameed, Thamer H and Abdi, Abdo S and Abdulfatah, Ayman N},
  journal={International Journal of Online and Biomedical Engineering (iJOE)},
  volume={17},
  number={14},
  pages={135--153},
  year={2021}
}

@article{hanhart2018jvet,
  title={JVET common test conditions and evaluation procedures for 360 video},
  author={Hanhart, Philippe and Boyce, Jill and Choi, Kiho and Lin, Jian-Liang},
  journal={Joint Video Exploration Team (JVET) of ITU-T SG},
  volume={16},
  year={2018}
}

@article{sun2017weighted,
  title={Weighted-to-spherically-uniform quality evaluation for omnidirectional video},
  author={Sun, Yule and Lu, Ang and Yu, Lu},
  journal={IEEE Signal Processing Letters},
  volume={24},
  number={9},
  pages={1408--1412},
  year={2017},
  publisher={IEEE}
}

@inproceedings{herglotz2022beyond,
  title={Beyond Bj{\o}ntegaard: Limits of Video Compression Performance Comparisons},
  author={Herglotz, Christian and Kr{\"a}nzler, Matthias and Mons, Ruben and Kaup, Andr{\'e}},
  booktitle={Proc. IEEE International Conference on Image Processing (ICIP)},
  pages={46--50},
  year={2022},
  organization={IEEE}
}

@article{huynh2008scope,
  title={Scope of validity of PSNR in image/video quality assessment},
  author={Huynh-Thu, Quan and Ghanbari, Mohammed},
  journal={Electronics letters},
  volume={44},
  number={13},
  pages={800--801},
  year={2008},
  publisher={IET}
}

@article{yejoint,
  title={Joint Video Exploration Team (JVET) of ITU-T SG 16 WP 3 and ISO/IEC JTC 1/SC 29/WG 11},
  author={Ye, Yan and Alshina, Elena and Boyce, Jill}
}

@article{kingma2014adam,
  title={Adam: A method for stochastic optimization},
  author={Kingma, Diederik P and Ba, Jimmy},
  journal = {International Conference on Learning Representations},
  year={2014}
}

@article{wallace1991jpeg,
  title={The JPEG still picture compression standard},
  author={Wallace, Gregory K},
  journal={Communications of the ACM},
  volume={34},
  number={4},
  pages={30--44},
  year={1991},
  publisher={AcM New York, NY, USA}
}

@article{zhu2000new,
  title={A new diamond search algorithm for fast block-matching motion estimation},
  author={Zhu, Shan and Ma, Kai-Kuang},
  journal={IEEE Transactions on Image Processing},
  volume={9},
  number={2},
  pages={287--290},
  year={2000},
  publisher={IEEE}
}

@article{watson1994image,
  title={Image compression using the discrete cosine transform},
  author={Watson, Andrew B and others},
  journal={Mathematica journal},
  volume={4},
  number={1},
  pages={81},
  year={1994},
  publisher={Citeseer}
}

@inproceedings{chen2017deepcoder,
  title={Deepcoder: A deep neural network based video compression},
  author={Chen, Tong and Liu, Haojie and Shen, Qiu and Yue, Tao and Cao, Xun and Ma, Zhan},
  booktitle={Proc. IEEE Visual Communications and Image Processing (VCIP)},
  pages={1--4},
  year={2017},
  organization={IEEE}
}

@article{liu2016cu,
  title={CU partition mode decision for HEVC hardwired intra encoder using convolution neural network},
  author={Liu, Zhenyu and Yu, Xianyu and Gao, Yuan and Chen, Shaolin and Ji, Xiangyang and Wang, Dongsheng},
  journal={IEEE Transactions on Image Processing},
  volume={25},
  number={11},
  pages={5088--5103},
  year={2016},
  publisher={IEEE}
}

@inproceedings{song2017neural,
  title={Neural network-based arithmetic coding of intra prediction modes in HEVC},
  author={Song, Rui and Liu, Dong and Li, Houqiang and Wu, Feng},
  booktitle={Proc. IEEE Visual Communications and Image Processing (VCIP)},
  pages={1--4},
  year={2017},
  organization={IEEE}
}

@inproceedings{lu2018deep,
  title={Deep kalman filtering network for video compression artifact reduction},
  author={Lu, Guo and Ouyang, Wanli and Xu, Dong and Zhang, Xiaoyun and Gao, Zhiyong and Sun, Ming-Ting},
  booktitle={Proc. the European Conference on Computer Vision (ECCV)},
  pages={568--584},
  year={2018}
}

@article{toderici2015variable,
  title={Variable rate image compression with recurrent neural networks},
  author={Toderici, George and O'Malley, Sean M and Hwang, Sung Jin and Vincent, Damien and Minnen, David and Baluja, Shumeet and Covell, Michele and Sukthankar, Rahul},
  journal={arXiv preprint arXiv:1511.06085},
  year={2015}
}

@inproceedings{rippel2017real,
  title={Real-time adaptive image compression},
  author={Rippel, Oren and Bourdev, Lubomir},
  booktitle={International Conference on Machine Learning},
  pages={2922--2930},
  year={2017},
  organization={PMLR}
}

@article{begaint2020compressai,
  title={Compressai: a pytorch library and evaluation platform for end-to-end compression research},
  author={B{\'e}gaint, Jean and Racap{\'e}, Fabien and Feltman, Simon and Pushparaja, Akshay},
  journal={arXiv preprint arXiv:2011.03029},
  year={2020}
}

@inproceedings{mun2012dpcm,
  title={DPCM for quantized block-based compressed sensing of images},
  author={Mun, Sungkwang and Fowler, James E},
  booktitle={Proc. 20th European Signal Processing Conference (EUSIPCO)},
  pages={1424--1428},
  year={2012},
  organization={IEEE}
}

@article{o2015introduction,
  title={An introduction to convolutional neural networks},
  author={O'Shea, Keiron and Nash, Ryan},
  journal={arXiv preprint arXiv:1511.08458},
  year={2015}
}

@article{lecun2015deep,
  title={Deep learning. nature, 521, 436-444},
  author={LeCun, Yann and Bengio, Yoshua and Hinton, Geoffrey and others},
  pages={25},
  year={2015}
}

@inproceedings{hosseini2016adaptive,
  title={Adaptive 360 VR video streaming: Divide and conquer},
  author={Hosseini, Mohammad and Swaminathan, Viswanathan},
  booktitle={Proc. IEEE International Symposium on Multimedia (ISM)},
  pages={107--110},
  year={2016},
  organization={IEEE}
}

@inproceedings{herglotz2019efficient,
  title={Efficient coding of 360° videos exploiting inactive regions in projection formats},
  author={Herglotz, Christian and Jamali, Mohammadreza and Coulombe, St{\'e}phane and Vazquez, Carlos and Vakili, Ahmad},
  booktitle={Proc. IEEE International Conference on Image Processing (ICIP)},
  pages={1104--1108},
  year={2019},
  organization={IEEE}
}

@inproceedings{jamali2019comparison,
  title={Comparison of 3D 360-degree video compression performance using different projections},
  author={Jamali, Mohammadreza and Golaghazadeh, Firouzeh and Coulombe, St{\'e}phane and Vakili, Ahmad and Vazquez, Carlos},
  booktitle={Proc. IEEE Canadian Conference of Electrical and Computer Engineering (CCECE)},
  pages={1--6},
  year={2019},
  organization={IEEE}
}

@inproceedings{argyriou2016engaging,
  title={Engaging immersive video consumers: Challenges regarding 360-degree gamified video applications},
  author={Argyriou, Lemonia and Economou, Daphne and Bouki, Vassiliki and Doumanis, Ioannis},
  booktitle={Proc. 15th international conference on ubiquitous computing and communications and 2016 international symposium on cyberspace and security (IUCC-CSS)},
  pages={145--152},
  year={2016},
  organization={IEEE}
}

@article{horn1981determining,
  title={Determining optical flow},
  author={Horn, Berthold KP and Schunck, Brian G},
  journal={Artificial intelligence},
  volume={17},
  number={1-3},
  pages={185--203},
  year={1981},
  publisher={Elsevier}
}

@article{bruhn2005lucas,
  title={Lucas/Kanade meets Horn/Schunck: Combining local and global optic flow methods},
  author={Bruhn, Andr{\'e}s and Weickert, Joachim and Schn{\"o}rr, Christoph},
  journal={International Journal of Computer Vision},
  volume={61},
  pages={211--231},
  year={2005},
  publisher={Springer}
}

@inproceedings{he2018jvet,
  title={JVET AHG Report: 360 Video Conversion Software Development, document JVET-I0006, Joint Video Exploration Team (JVET) of ITU-T SG 16 WP3 and ISO/IEC JTC1 SC29/WG11, Gwangju, South Korea},
  author={He, Y and Choi, K and Zakharchenko, V},
  booktitle={9th Meeting, Jan},
  year={2018}
}

@article{christopoulos2000jpeg2000,
  title={The JPEG2000 still image coding system: an overview},
  author={Christopoulos, Charilaos and Skodras, Athanassios and Ebrahimi, Touradj},
  journal={IEEE Transactions on Consumer Electronics},
  volume={46},
  number={4},
  pages={1103--1127},
  year={2000},
  publisher={IEEE}
}

@article{taubman2000high,
  title={High performance scalable image compression with EBCOT},
  author={Taubman, David},
  journal={IEEE Transactions on Image Processing},
  volume={9},
  number={7},
  pages={1158--1170},
  year={2000},
  publisher={IEEE}
}

@inproceedings{hershey2007approximating,
  title={Approximating the Kullback Leibler divergence between Gaussian mixture models},
  author={Hershey, John R and Olsen, Peder A},
  booktitle={Proc. IEEE International Conference on Acoustics, Speech and Signal Processing-ICASSP'07},
  volume={4},
  pages={IV--317},
  year={2007},
  organization={IEEE}
}

@inproceedings{ilg2017flownet,
  title={Flownet 2.0: Evolution of optical flow estimation with deep networks},
  author={Ilg, Eddy and Mayer, Nikolaus and Saikia, Tonmoy and Keuper, Margret and Dosovitskiy, Alexey and Brox, Thomas},
  booktitle={Proc. IEEE Conference on Computer Vision and Pattern Recognition},
  pages={2462--2470},
  year={2017}
}

@inproceedings{ranjan2017optical,
  title={Optical flow estimation using a spatial pyramid network},
  author={Ranjan, Anurag and Black, Michael J},
  booktitle={Proc. IEEE Conference on Computer Vision and Pattern Recognition},
  pages={4161--4170},
  year={2017}
}

@article{shah2021traditional,
  title={Traditional and modern strategies for optical flow: an investigation},
  author={Shah, Syed Tafseer Haider and Xuezhi, Xiang},
  journal={SN Applied Sciences},
  volume={3},
  pages={1--14},
  year={2021},
  publisher={Springer}
}

@article{beauchemin1995computation,
  title={The computation of optical flow},
  author={Beauchemin, Steven S. and Barron, John L.},
  journal={ACM computing surveys (CSUR)},
  volume={27},
  number={3},
  pages={433--466},
  year={1995},
  publisher={ACM New York, NY, USA}
}

@inproceedings{wang2003multiscale,
  title={Multiscale structural similarity for image quality assessment},
  author={Wang, Zhou and Simoncelli, Eero P and Bovik, Alan C},
  booktitle={The Thrity-Seventh Asilomar Conference on Signals, Systems \& Computers, 2003},
  volume={2},
  pages={1398--1402},
  year={2003},
  organization={Ieee}
}

\end{document}